\documentclass{article}

\PassOptionsToPackage{numbers,sort&compress}{natbib}
\usepackage[preprint]{neurips/neurips_2026}
\usepackage[utf8]{inputenc}
\usepackage[T1]{fontenc}
\usepackage{amsmath, amssymb, amsthm, amsfonts}
\usepackage{graphicx}
\usepackage{booktabs}
\usepackage{hyperref}
\usepackage{url}
\usepackage{microtype}
\usepackage{placeins}
\usepackage{float}
\usepackage{cleveref}
\usepackage{algorithm}
\usepackage{algorithmic}
\usepackage{xcolor}
\usepackage{multirow}
\usepackage{array}

\newcolumntype{L}[1]{>{\raggedright\arraybackslash}p{#1}}

\title{Sign-Aware Gated Sparse Autoencoders:\\ Modeling Anticorrelated Features with Bi-Jump-ReLU Activations}
\author{
    \textbf{Bartosz Wieciech} \quad
    \textbf{Zmnako Awrahman} \quad
    \textbf{Marcin Czelej} \\
    \textbf{Victor Hugo Jaramillo Velasquez} \quad
    \textbf{Wioletta Stobieniecka} \\
    Amazon Web Services \\
    \texttt{bartwie@amazon.com}
}
\date{}

\newcommand{\R}{\mathbb{R}}
\newcommand{\E}{\mathbb{E}}

\begin{document}

\maketitle

\begin{abstract}
Sparse Autoencoders (SAEs) extract interpretable features from Large Language Model activations, but standard
variants enforce non-negative latents, so a bidirectional semantic axis (e.g., ``pressure too high'' vs.\
``pressure too low'') must be split across two latents, wasting dictionary capacity on anticorrelated features.
We propose the \textit{Sign-Aware Gated SAE} (SA-GSAE), which combines two-sided gated sparsity, signed
shrinkage-free magnitudes, and auxiliary gate supervision in a new \textit{Bi-Jump-ReLU} activation, so that a
single latent carries both polarities of one decoder direction; parameter accounting shows sign-awareness stays
parameter-efficient even when anticorrelated pairs are rare. Across three mid-depth hookpoints on Pythia-1B and
SmolLM3-3B (six cells, three seeds), a half-width SA-GSAE empirically dominates the aggregate mean frontier of a
full-width Gated SAE on three of six cells, matches its $R^2$ within $0.025$ on the remaining three, and cuts
dead fraction by $0.35$--$0.82$ absolute at matched $L_0 = 64$ on all six. Ablations show the two-sided gate and
the auxiliary loss are essential whereas per-polarity asymmetry is not; we recommend the fully tied symmetric
variant as the default. A blinded semantic audit finds nameable opposition between a latent's two sides is rare
for SA-GSAE and all tested baselines, while sign-conditioned interventions show a single signed latent acts as a
bidirectional causal dial where a pair of ``opposite'' non-negative latents does not; we scope interpretability
claims accordingly. At full width, SA-GSAE is over-parameterized and its reported configuration exhibits a
reproducible reconstruction collapse at the SmolLM3-3B residual-stream site; the recommended configuration
(small threshold initialization with dead-latent threshold resets) prevents it.
\end{abstract}

\section{Introduction}

Mechanistic interpretability seeks human-usable descriptions of the internal computations of trained neural networks.
In large language models, the most useful units of analysis often look less like individual neurons and more like
feature directions or subspaces that participate in larger circuits \cite{olah2022mechanistic, olah2020circuits, vaswani2017attention}.
A common working picture is that a token activation $x$ is a superposition of a small number of latent features,
even if this picture is only approximate in practice \cite{bolukbasi2021interpretability, elhage2022toy, park2023linear, engels2024notalllinear}.

Classical sparse coding provides a natural formalism for this goal: given data vectors $x \in \R^{d_{\text{in}}}$, learn a
dictionary $D \in \R^{d_{\text{in}}\times H}$ and sparse coefficients $z$ via
\begin{equation}
\label{eq:dict_learning}
\min_{D,\{z^{(n)}\}} \sum_{n} \tfrac12\|x^{(n)} - D z^{(n)}\|_2^2 + \lambda \|z^{(n)}\|_1,
\qquad \|d_i\|_2 = 1,
\end{equation}
where $d_i$ denotes the $i$th column of $D$ \cite{olshausen1996emergence, tibshirani1996lasso, aharon2006ksvd, mairal2009online}.
In this formulation the coefficients are signed: positive and negative values correspond to opposite directions along the
same atom, unlike explicitly non-negative factorizations that enforce parts-based structure \cite{lee1999nmf}. Sparse
autoencoders amortize the per-example inference problem into a single forward pass, making sparse dictionary learning
practical for large activation caches and highly overcomplete regimes \cite{gregor2010lista}.

Applied to LLM activations, SAEs are now a standard way to learn overcomplete dictionaries of residual-stream, MLP, or attention-output
features \cite{bricken2023towards, cunningham2023sparse, gao2024scaling, lieberum2024gemmascope, he2024llamascope}.
Empirically, many learned latents align with
human-interpretable concepts and support causal interventions, suggesting that SAEs can partially ``resolve''
superposition in practice \cite{bricken2023towards, cunningham2023sparse, gao2024scaling}. At the same time, SAEs do not
necessarily recover a unique or canonical basis of features, and theoretical analyses have begun to characterize their
intrinsic amortization gaps and representational limits \cite{oneill2024amortisation, cui2025limits}. Larger SAEs can
reveal novel features, and meta-SAEs can
decompose some latents into combinations of others \cite{leask2025canonical}. Independently initialized SAEs trained on
the same data can also recover substantially different feature sets \cite{paulo2025different}. We treat SAE
dictionaries as useful decompositions rather than unique ground-truth bases.

A structural limitation of many popular SAE variants is non-negativity. ReLU-based and many gated models can only express
$f_i(x) \ge 0$, so a bidirectional semantic axis is often represented by two latents, one for positive evidence and one for
negative evidence. For axes such as ``pressure too high'' versus ``pressure too low'', this sign splitting duplicates
decoder capacity without adding geometric expressivity. Classical sparse coding does not require this, because coefficients
are signed, and recent SAE work has started to make the same point explicitly, for example via AbsTopK \cite{zhu2025abstopk}.

Building on Gated SAEs, which separate support selection from magnitude estimation and thereby avoid classical $L_1$
shrinkage \cite{rajamanoharan2024gated}, we introduce the \textit{Sign-Aware Gated SAE}. A single latent can now represent
positive or negative evidence along one decoder direction via a
new \textit{Bi-Jump-ReLU} activation. \S\ref{sec:parameter_efficiency} gives a simple accounting argument showing that if a
fraction $p$ of features occurs as anticorrelated pairs, the required width drops to $H_\pm = H(1-p/2)$ and the construction
becomes parameter-efficient once $p \gtrsim 6/(d_{\text{in}}+3)$. We treat this as a structural sanity check, not an
identifiability theorem.

\section{Contributions}

\begin{enumerate}
    \item \textbf{Two-sided gated sparsity with signed magnitude and auxiliary supervision.} We introduce a
    polarity-sensitive Bi-Jump-ReLU activation combining a two-sided learnable dead zone with a signed-magnitude
    path, and train it with Gated-style auxiliary reconstruction. Ablations show the two-sided gate and the
    auxiliary loss are required on real LLM activations (removing the auxiliary collapses LR to
    $0.27$ and pushes dead fraction to $98\%$); bipolar sharing along a single decoder direction is a
    consequence of this architecture rather than an independent mechanism.
    \item \textbf{A parameter-efficiency argument and a simplified tied default.} We give a simple
    width/parameter accounting argument for when sign awareness helps, and we test its most direct implication via a
    half-width versus full-width comparison on real LLM activations. Tying $r_i^+ = r_i^-$ (symmetric-magnitude
    variant) is practically indistinguishable from independent per-polarity scaling on real activations
    ($|\Delta R^2| = 0.0015$, $|\Delta\text{MSE}| = 0.0002$ on Pythia-1B \texttt{mlp\_out}; $\gamma_+$/$\gamma_-$
    medians agree within $\le 0.05$ on every cell), and separate thresholds are \emph{exactly} reparameterizable as a
    single symmetric threshold plus a gate-bias shift (\cref{app:theory}); we therefore recommend the fully tied
    symmetric variant as the default (validated at matched $L_0$: tied $R^2 = 0.7193 \pm 0.0005$ vs.\ asymmetric
    reference $0.7169 \pm 0.0004$; \cref{app:rebuttal_studies}).
    \item \textbf{Controlled and real-model evidence.} On a controlled signed-axis benchmark, SA-GSAE matches a $2\times$ width
    non-negative Gated SAE with half as many latents. On real activations, across three mid-depth hookpoints and two
    backbones, a half-width SA-GSAE cuts dead fraction by $0.35$--$0.82$ absolute at matched $L_0 = 64$
    relative to a full-width Gated SAE (per-cell sweep-geomean reduction factors of ${\sim}1.5\times$ to
    ${\sim}56\times$ under a disclosed $0.01$ denominator floor; ratios are secondary to the absolute reductions), while
    matching or exceeding its reconstruction fidelity and empirically dominating its aggregate mean frontier over the full swept $L_0$ overlap on
    3 of 6 cells (listed in \S\ref{sec:results_llm_benchmark}); on the remaining 3 cells
    $\Delta R^2 \in [-0.008,\,-0.001]$ at matched $L_0 = 64$. The MLP-output
    capacity wins come from most latents carrying signal on both polarities; on attention, bipolar structure is
    instead concentrated in a small set of top-activation latents. At full width, SA-GSAE is over-parameterized and its reported
    configuration exhibits a reproducible
    collapse on the SmolLM3-3B residual stream that the reported half-width configuration -- which also enables
    dead-latent threshold resets -- avoids (diagnosis in \cref{app:collapse_diagnosis}); the halved-width operating
    point remains motivated primarily by the capacity argument rather than by this stability contrast.
\end{enumerate}

To our knowledge, this is the first two-sided, gated, no-shrinkage SAE for signed feature sharing.

\section{Method}
\label{sec:method}

We generalize the Gated SAE to signed activations while preserving its no-shrinkage separation between detection and magnitude
\cite{rajamanoharan2024gated}. Let $D \in \mathbb{R}^{d_{\text{in}} \times H}$ denote the decoder dictionary, where $H$ is the SAE
width (the number of latents), with columns $D_{:,i}$, and let $b_{\text{dec}} \in \mathbb{R}^{d_{\text{in}}}$ denote the decoder bias.
As in standard SAE implementations, we center inputs by $b_{\text{dec}}$ when forming encoder projections.

\subsection{The Bi-Jump-ReLU Activation}
\label{sec:bijumprelu}

For each latent we use a decoder-aligned projection, and the gate pre-activation is an affine transformation of it,
\begin{equation}
    t_i(x) = D_{:,i}^\top (x - b_{\text{dec}}), \qquad
    \pi_i(x) = \alpha_i\, t_i(x) + \beta_i, \qquad i = 1,\dots,H,
\end{equation}
with trainable per-latent parameters $\alpha_i, \beta_i \in \mathbb{R}$ (in the implementation
$\alpha_i = \exp(\log \alpha_i) > 0$; see Appendix~\ref{app:theory}).

We introduce two threshold parameters $\delta_i^+, \delta_i^- \ge 0$, defining a learnable dead zone, and two unconstrained
log-scale parameters $r_i^+, r_i^- \in \R$, which parameterize non-negative magnitude scales $g_i^+ = \exp(r_i^+)$ and
$g_i^- = \exp(r_i^-)$. Define the polarity variable
\begin{equation}
    s_i(x) =
    \begin{cases}
        +1 & \text{if } \pi_i(x) > \delta_i^+ \\
        -1 & \text{if } \pi_i(x) < -\delta_i^- \\
        0 & \text{otherwise.}
    \end{cases}
\end{equation}
The signed latent activation is then
\begin{equation}
    \label{eq:bi_jump_relu}
    a_i(x) =
    \begin{cases}
        \text{ReLU}\big(g_i^+ t_i(x) + b_{\text{mag}, i}\big) & \text{if } s_i(x) = +1 \\
        -\text{ReLU}\big(-g_i^- t_i(x) + b_{\text{mag}, i}\big) & \text{if } s_i(x) = -1 \\
        0 & \text{if } s_i(x) = 0,
    \end{cases}
\end{equation}
where $b_{\text{mag},i} \in \mathbb{R}$ is a shared magnitude offset and $g_i^\pm = \exp(r_i^\pm) \ge 0$ are separate gains for
positive and negative activations, with the exponent used to guarantee positive values.

Bi-Jump-ReLU is a signed dead-zone unit: it is identically zero inside $[-\delta_i^-, \delta_i^+]$, uses gain $g_i^+$ on the
positive side, and gain $g_i^-$ on the negative side, while both polarities share the same decoder column. Unlike soft-thresholding
or one-sided JumpReLU, thresholds decide support but are not subtracted from the active magnitude. Related two-sided thresholded
units have appeared in other settings, including symmetric threshold-linear networks and symmetric-threshold ReLU constructions
\cite{hahnloser2003thresholdlinear, han2023strelu}. Our novelty is the integration of this signed dead-zone behavior into a
decoder-aligned gated SAE with shared signed decoder directions and the no-shrinkage support/magnitude separation of Gated SAEs.

\begin{figure}[t]
    \centering
    \includegraphics[width=0.72\linewidth]{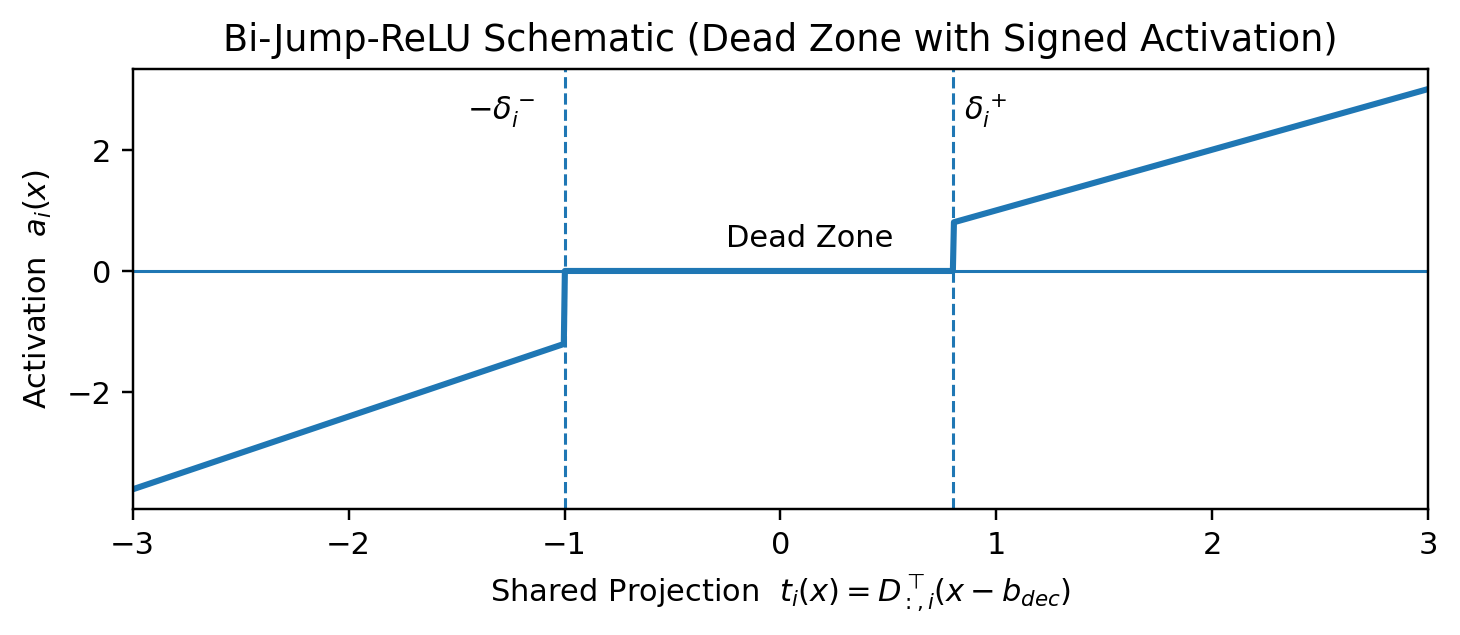}
    \caption{Bi-Jump-ReLU is zero inside a learnable dead zone $[-\delta_i^-, \delta_i^+]$ and emits signed magnitude outside it, using one decoder direction for both polarities.}
    \label{fig:bijump_schematic}
\end{figure}

\subsection{Training Objective}

The decoder reconstructs as $\hat{x} = D a(x) + b_{\text{dec}}$,
where $a(x) \in \mathbb{R}^H$ collects the Bi-Jump-ReLU activations. We adapt the Gated SAE loss by placing sparsity pressure on the
gate and using an auxiliary reconstruction to train detection separately from the magnitude path:
\begin{equation}
    \mathcal{L}(x) = \underbrace{\|x - D a(x) - b_{\text{dec}}\|_2^2}_{\text{Main Reconstruction}} + \lambda \sum_i \Omega_{\text{gate}}(\pi_i(x)) + \lambda_{\text{aux}}\,\mathcal{L}_{\text{aux}}(x).
\end{equation}
The gate penalty is a two-sided hinge,
\begin{equation}
    \Omega_{\text{gate}}(\pi_i(x)) = \text{ReLU}\big(\pi_i(x) - \delta_i^+\big) + \text{ReLU}\big(-\pi_i(x) - \delta_i^-\big),
\end{equation}
which is zero inside the dead zone and grows linearly outside it.

To supervise the gate directly, we use an auxiliary reconstruction with a stop-gradient decoder path,
\begin{equation}
    \mathcal{L}_{\text{aux}}(x) = \big\|x - D^{\text{sg}} \big[ \text{ReLU}(\pi(x) - \delta^+) - \text{ReLU}(-\pi(x) - \delta^-) \big] - b_{\text{dec}}^{\text{sg}}\big\|_2^2,
\end{equation}
where $D^{\text{sg}} := \mathrm{stopgrad}(D)$ and $b_{\text{dec}}^{\text{sg}} := \mathrm{stopgrad}(b_{\text{dec}})$ are treated as constants for this term.
This preserves the intended separation: the auxiliary objective teaches the gate which side of the dead zone to cross, while the
main reconstruction learns the signed magnitude once a latent is active. Implementation details and effective-threshold
analysis are deferred to Appendix~\ref{app:theory}. LLM baseline definitions are summarized in
\cref{sec:llm_baseline_defs}.

\subsection{Parameter Efficiency \& Width Reduction}
\label{sec:parameter_efficiency}

Suppose a baseline SAE has width $H$, and that a fraction $p \in [0, 1]$ of the ground-truth features in the data
representation appear as perfectly anticorrelated pairs aligned to directions $\pm d$. If there are $N$ total ground-truth
features, then $pN$ of them participate in pairs, yielding $(pN)/2$ pairs and $(1-p)N$ unpaired features. A baseline SAE
requires two latents for each pair, so it uses $H = N$ latents in total.

The Sign-Aware model instead requires only one latent per anticorrelated pair, plus one latent for each unpaired
feature. The effective width required for the Sign-Aware model is therefore
\begin{equation}
    H_{\pm} = (1-p)N + \frac{pN}{2} = N \left( 1 - \frac{p}{2} \right) = H \left( 1 - \frac{p}{2} \right).
\end{equation}
If $p=1$ (all features are paired), the dictionary size halves.

We now analyze the parameter overhead. Let $d_{\text{in}}$ be the input dimensionality. In a standard Gated SAE, each
latent carries roughly $d_{\text{in}}$ parameters for the decoder direction plus $O(1)$ scalar parameters for gate biases
and scales; we summarize this as $d_{\text{in}} + c$ parameters per latent for some small constant $c$. Relative to this
baseline, the Sign-Aware architecture adds 3 scalars per latent: the two thresholds $\delta_i^\pm$ and one
\emph{additional} magnitude scale ($r_i^-$; the gate parameters $\alpha_i, \beta_i$ and the first magnitude scale are
already present in the Gated baseline). Per-latent parameters therefore become $d_{\text{in}} + c + 3$. This is a decoder-only
count for the decoder-aligned tied implementation used throughout this paper, which has no independent encoder matrix (the
encoder projection reuses the normalized decoder, \cref{sec:bijumprelu}).

Comparing total parameter counts $P_{\text{base}} \approx H (d_{\text{in}} + c)$ and
$P_{\text{sign}} \approx H_{\pm} (d_{\text{in}} + c + 3) = H \left(1 - \frac{p}{2}\right) (d_{\text{in}} + c + 3)$,
the condition for net parameter savings $P_{\text{sign}} < P_{\text{base}}$ reduces to
$p > 6/(d_{\text{in}} + c + 3)$ (full derivation in \cref{app:param_savings_derivation}).
Taking $c \approx 0$ for simplicity yields the approximate threshold
\begin{equation}
    p \gtrsim \frac{6}{d_{\text{in}} + 3}.
\end{equation}
At all six hookpoint $\times$ backbone cells of our LLM benchmark the hookpoint activation
dimension is $d_{\text{in}} = 2048$ (both Pythia-1B and SmolLM3-3B have hidden size $2048$),
so the threshold becomes $p \gtrsim 6/2051 \approx 0.29\%$. Under a simple
parameter-counting model, the Sign-Aware SAE is therefore strictly parameter-efficient as
soon as anticorrelated pairs constitute a small fraction of the feature set, on the order
of $10^{-3}$ for typical transformer-scale widths. More generally, a variant that adds $q$
scalars per latent breaks even at $p > 2q/(d_{\text{in}} + c + q)$: the tied-gain variant
($q = 2$) at $p \gtrsim 4/(d_{\text{in}}+2) \approx 0.20\%$ and the fully tied variant
(single symmetric threshold and tied gains, $q = 1$) at
$p \gtrsim 2/(d_{\text{in}}+1) \approx 0.10\%$.

\paragraph{Symmetric-magnitude and tied variants.}
A symmetric-magnitude variant that ties $r_i^+=r_i^-$, reducing the per-latent scalar overhead by one, and the exact
reparameterization of separate thresholds into a single symmetric threshold, are discussed in
Appendix~\ref{app:theory}.

\paragraph{Both-sign calibration coverage at transformer hookpoints.}
A direct test of the parameter-efficiency premise requires an observation of the
fraction $p$ of feature axes that admit sign-sharing in real activations. Since the
ground-truth feature basis is not known, we measure
$\hat{p} = \texttt{both\_fraction\_valid}$ -- the fraction of trained SA-GSAE latents
whose $\gamma_+$ and $\gamma_-$ are each individually identifiable from a held-out
activation cache (each regime has $\ge 32$ firing tokens with non-degenerate
squared-activation mass). We call this the \emph{both-sign calibration coverage}. It is a
capacity-usage statistic, not a semantic one, and it is neither a lower nor an upper bound
on the true $p$: unrelated sign-packing can inflate it, while consolidation failures and
strongly sign-imbalanced firing deflate it (see the blinded semantic audit in
\cref{app:rebuttal_studies}). For the width/parameter accounting above, however, two-sided
\emph{usage} is exactly the relevant premise. At matched $L_0 = 64$ and $d_{\text{in}} = 2048$ the theoretical threshold is
$p \gtrsim 0.29\%$. The observed $\hat{p}$ ranges from $0.039$ (\texttt{resid-mid/SmolLM3-3B} full-width) to $0.767$
(\texttt{mlp\_out-mid/SmolLM3-3B} half-width), clearing the theoretical threshold by
${\sim}13\times$ at the worst observed cell and by roughly $260\times$ at the best
(\cref{tab:bipolar_census_full}). Sign-awareness is therefore parameter-efficient in
practice, not merely in principle, across every benchmark cell we tested.

\section{Results}
\label{sec:results_llm_benchmark}

All experiments, including failed runs, closed in approximately 4000 compute hours of AWS g5.2xlarge instances
(NVIDIA A10G Tensor Core GPU) and 200 compute hours of AWS g4dn.xlarge instances (NVIDIA T4 GPU). Controlled synthetic
and toy experiments are reported in Appendix~\ref{app:protocols} and Appendix~\ref{app:protocol_results}; the licenses
and date of verification of models and datasets are listed in Appendix~\ref{app:asset_licenses}.

\paragraph{Benchmark setup.}
We cache token-level activations at three mid-depth hookpoints in each of two models: MLP output (\texttt{mlp\_out}),
attention output (\texttt{attn}), and post-block residual stream (\texttt{resid}). The backbones are Pythia-1B at layer
$8/16$ \cite{biderman2023pythia} (pretrained on The Pile \cite{gao2020pile}) and SmolLM3-3B at layer $18/36$ \cite{bakouch2025smollm3}. Activations are collected on
OpenWebText \cite{Gokaslan2019OpenWeb}, using length-128 sequences with a fixed 90/5/5 train/val/test split. The cache
uses 250k sequences per backbone with float16 precision.

We train width-$32{,}768$ SAEs (full width) and width-$16{,}384$ SAEs (half width) on the cached activations for each
variant, holding the LLM frozen and using the same cache across all variants.
Training runs in mixed precision at learning rate $3\cdot 10^{-4}$; Pythia uses batch size 256 for 225{,}000 steps and
SmolLM3 batch size 128 for 450{,}000 steps, with dead-latent resampling every 12{,}500 steps on both backbones.
SA-GSAE thresholds are frozen for the first 40{,}000 (Pythia) / 80{,}000 (SmolLM3) steps -- $17.8\%$ of training -- and
train for the remaining $82\%$; per-cell threshold initializations and the dead-latent threshold-reset setting are
listed in \cref{tab:sa_config} (\cref{app:collapse_diagnosis}).
For threshold/gated variants (Gated SAE, SA-GSAE) we sweep the sparsity coefficient $\lambda$ over 8 log-spaced values;
for AbsTopK we train with $k \in \{16, 32, 64, 128\}$. All results report mean $\pm$ SE over 3
random seeds. We report reconstruction MSE and $R^2$, mean $L_0$, dead-feature fraction, and Loss Recovered (LR), defined
as the fraction of the clean-to-zero-ablation language-model loss gap recovered when the SAE reconstruction is patched back
into the same hookpoint.

LR saturates at residual-stream hookpoints and has a noise-dominated ceiling on \texttt{attn-mid/SmolLM3-3B}; we therefore
treat $R^2$ and MSE as primary discriminators there and do not compare LR across hookpoints.

\paragraph{Comparison protocol and scope.}
All LLM variants are trained in the same shared harness with matched width, activation cache, data split, seed count,
and overall training budget. For continuous sweeps (Gated SAE, SA-GSAE) the exact per-method $\lambda$ grid differs
slightly, so matched-point quantities are linearly interpolated
in $\log L_0$ between adjacent sweep points on each seed's curve before aggregating; for AbsTopK we report the nearest
fixed-$k$ anchor. The benchmark covers three mid-depth hookpoints per backbone and focuses on reconstruction/capacity
metrics; it does not by itself establish downstream interpretability wins across layers or model families.

\paragraph{Baselines.}
The main-text comparison uses Gated SAE and AbsTopK baselines implemented in the shared harness; exact
implementation details, and a fairness note on encoder parameterization (SA-GSAE and Gated SAE use the shared
decoder-aligned projection, whereas AbsTopK retains its native encoder, so the LLM tables are end-to-end recipe
comparisons rather than a pure isolation of encoder tying), are given in \cref{sec:llm_baseline_defs}. The hybrid
AbsTopK+GatedMag ablation in Appendix~\ref{sec:ablations} partially disentangles the encoder question by pairing
our magnitude path with a fixed-$k$ signed selector.

\begin{figure*}[!h]
    \centering
    \IfFileExists{llm_benchmarking_plots/pythia1b_mlp_out_mid/dead_fraction_vs_log_l0.png}{
        \includegraphics[width=0.32\textwidth]{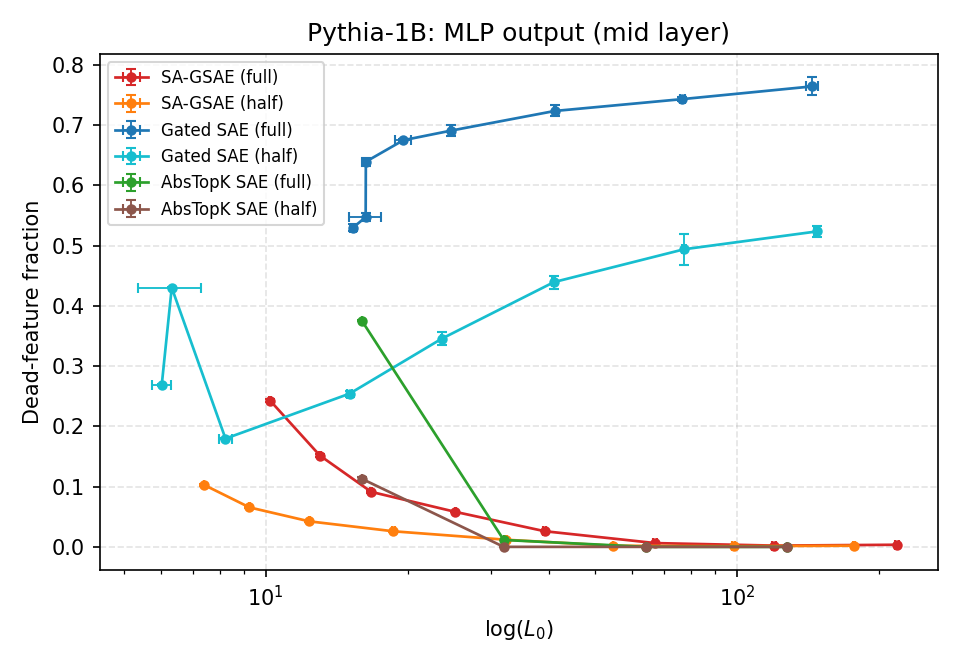}
    }{
        \fbox{\parbox[c][0.18\textwidth][c]{0.30\textwidth}{\centering Missing: pythia mlp\_out dead}}
    }
    \hfill
    \IfFileExists{llm_benchmarking_plots/pythia1b_attn_mid/dead_fraction_vs_log_l0.png}{
        \includegraphics[width=0.32\textwidth]{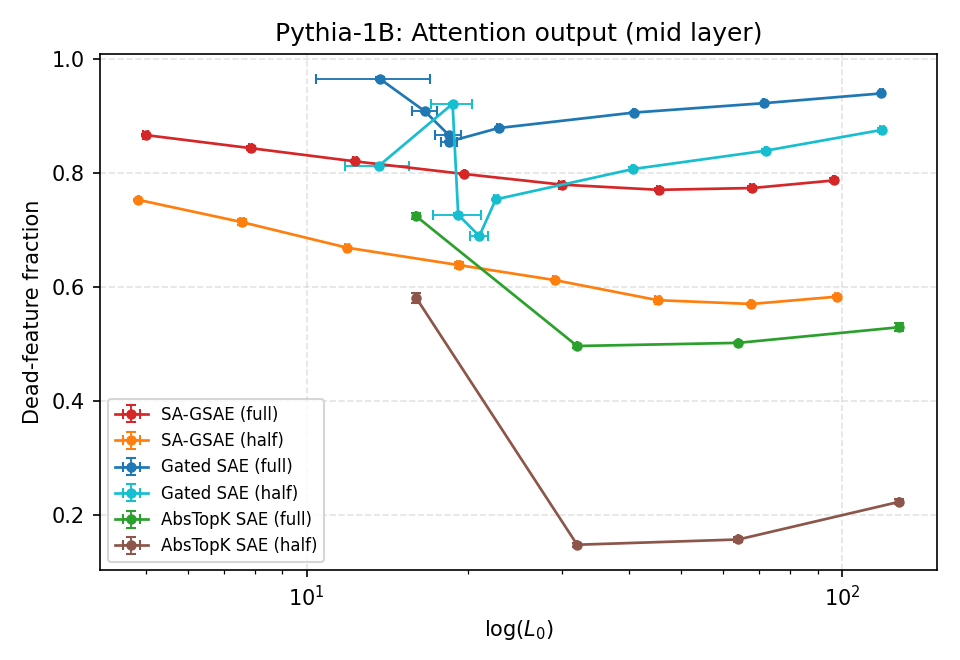}
    }{
        \fbox{\parbox[c][0.18\textwidth][c]{0.30\textwidth}{\centering Missing: pythia attn dead}}
    }
    \hfill
    \IfFileExists{llm_benchmarking_plots/pythia1b_resid_mid/dead_fraction_vs_log_l0.png}{
        \includegraphics[width=0.32\textwidth]{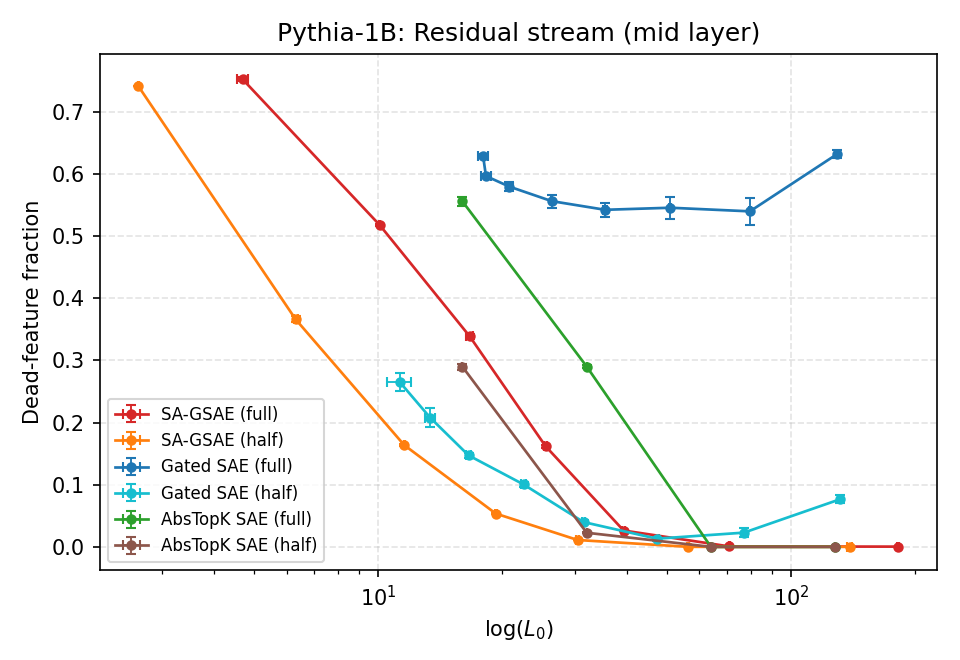}
    }{
        \fbox{\parbox[c][0.18\textwidth][c]{0.30\textwidth}{\centering Missing: pythia resid dead}}
    }

    \vspace{0.5em}

    \IfFileExists{llm_benchmarking_plots/smollm3_3b_mlp_out_mid/dead_fraction_vs_log_l0.png}{
        \includegraphics[width=0.32\textwidth]{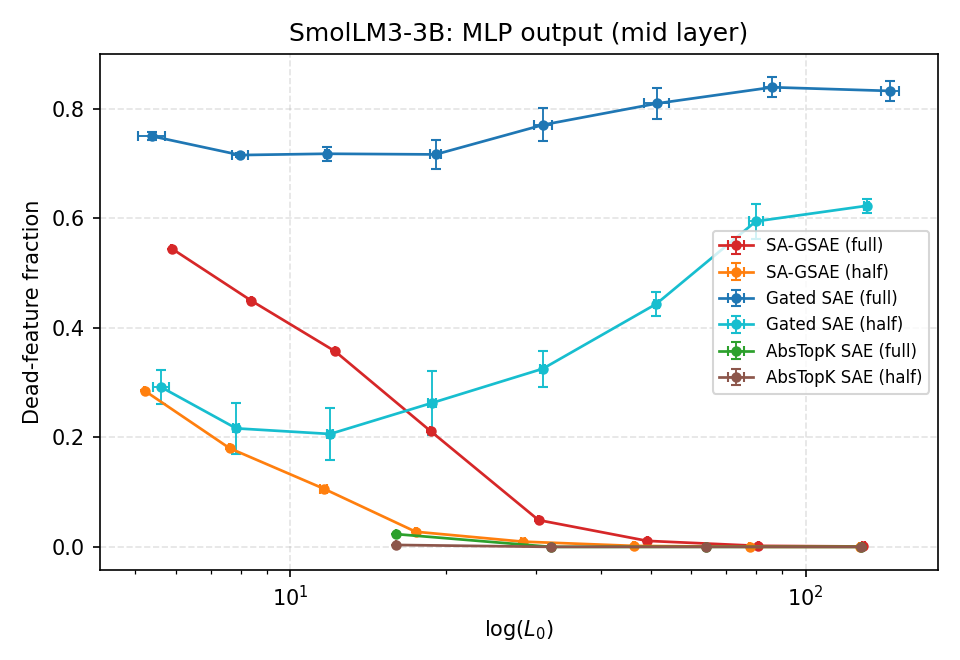}
    }{
        \fbox{\parbox[c][0.18\textwidth][c]{0.30\textwidth}{\centering Missing: SmolLM3 mlp\_out dead}}
    }
    \hfill
    \IfFileExists{llm_benchmarking_plots/smollm3_3b_attn_mid/dead_fraction_vs_log_l0.png}{
        \includegraphics[width=0.32\textwidth]{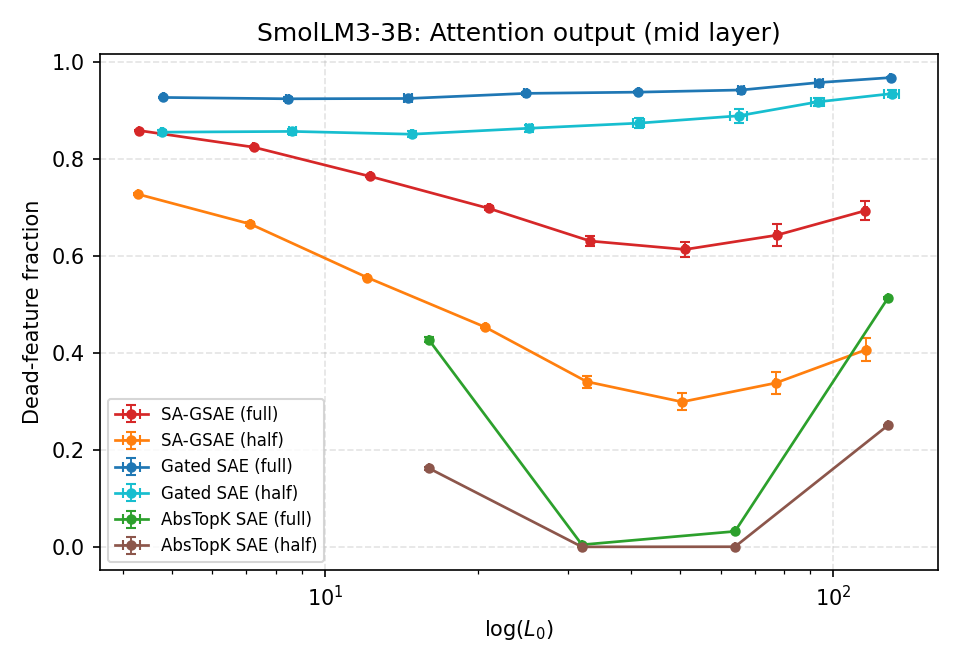}
    }{
        \fbox{\parbox[c][0.18\textwidth][c]{0.30\textwidth}{\centering Missing: SmolLM3 attn dead}}
    }
    \hfill
    \IfFileExists{llm_benchmarking_plots/smollm3_3b_resid_mid/dead_fraction_vs_log_l0.png}{
        \includegraphics[width=0.32\textwidth]{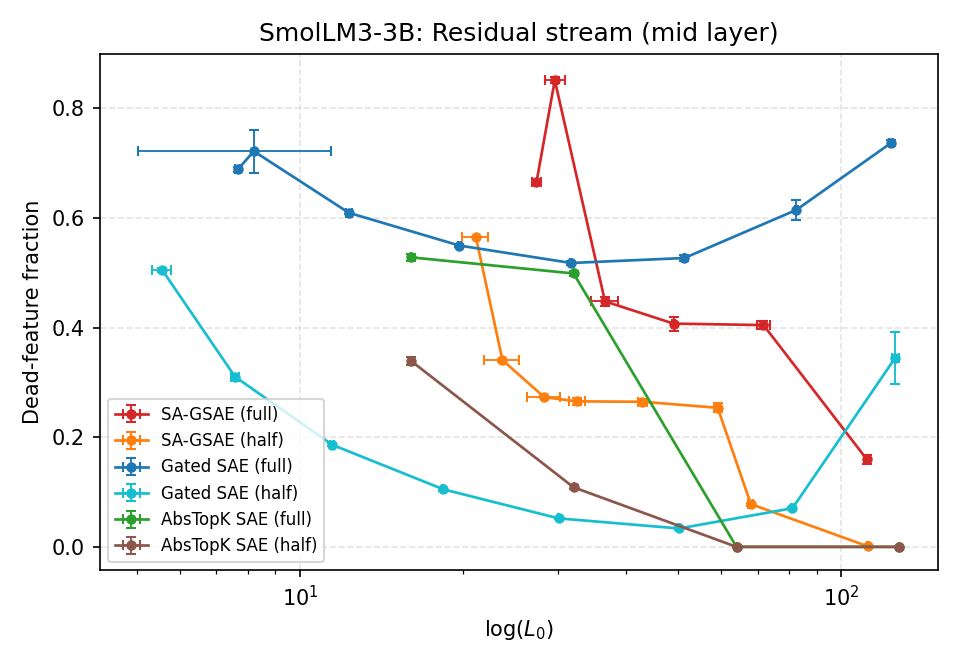}
    }{
        \fbox{\parbox[c][0.18\textwidth][c]{0.30\textwidth}{\centering Missing: SmolLM3 resid dead}}
    }
    \caption{Dead-feature fraction vs $\log(L_0)$ across the three mid-depth hookpoints (columns: \texttt{mlp\_out}, \texttt{attn}, \texttt{resid}). Top row: Pythia-1B (layer $8/16$). Bottom row: SmolLM3-3B (layer $18/36$). Error bars show $\pm$SE over 3 seeds on both axes.}
    \label{fig:llm_dead_fraction_tradeoffs}
\end{figure*}

\begin{figure*}[!h]
    \centering
    \IfFileExists{llm_benchmarking_plots/pythia1b_mlp_out_mid/loss_recovered_vs_log_l0.png}{
        \includegraphics[width=0.32\textwidth]{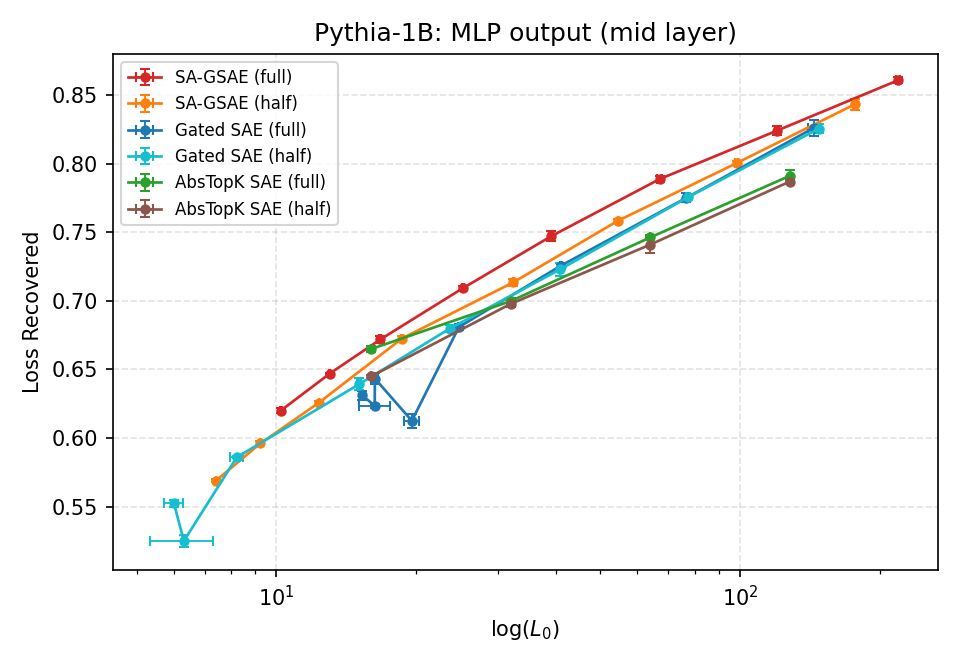}
    }{
        \fbox{\parbox[c][0.18\textwidth][c]{0.30\textwidth}{\centering Missing: pythia mlp\_out LR}}
    }
    \hfill
    \IfFileExists{llm_benchmarking_plots/pythia1b_attn_mid/loss_recovered_vs_log_l0.png}{
        \includegraphics[width=0.32\textwidth]{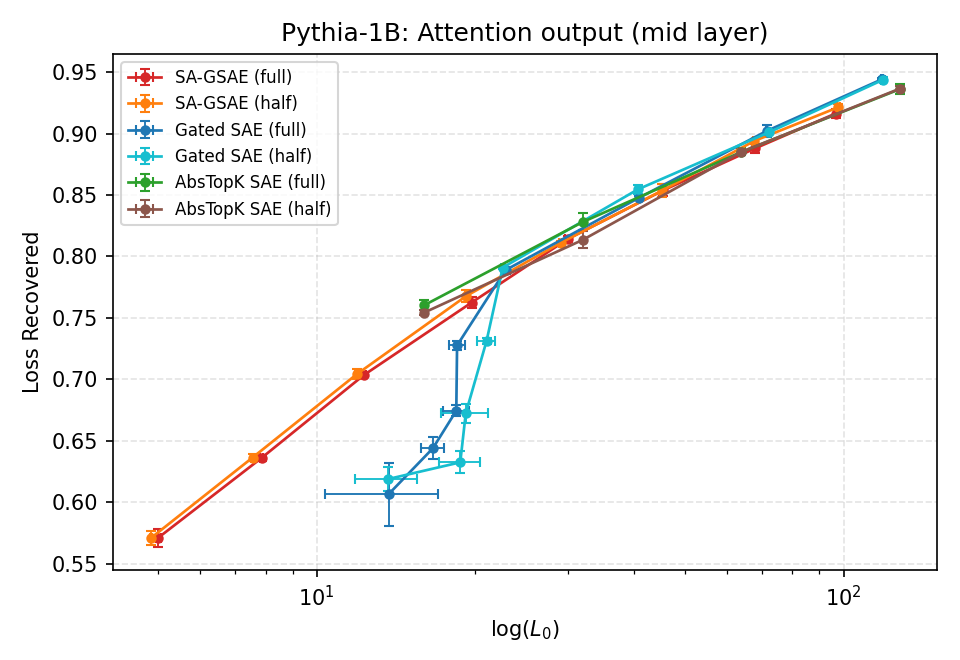}
    }{
        \fbox{\parbox[c][0.18\textwidth][c]{0.30\textwidth}{\centering Missing: pythia attn LR}}
    }
    \hfill
    \IfFileExists{llm_benchmarking_plots/pythia1b_resid_mid/loss_recovered_vs_log_l0.png}{
        \includegraphics[width=0.32\textwidth]{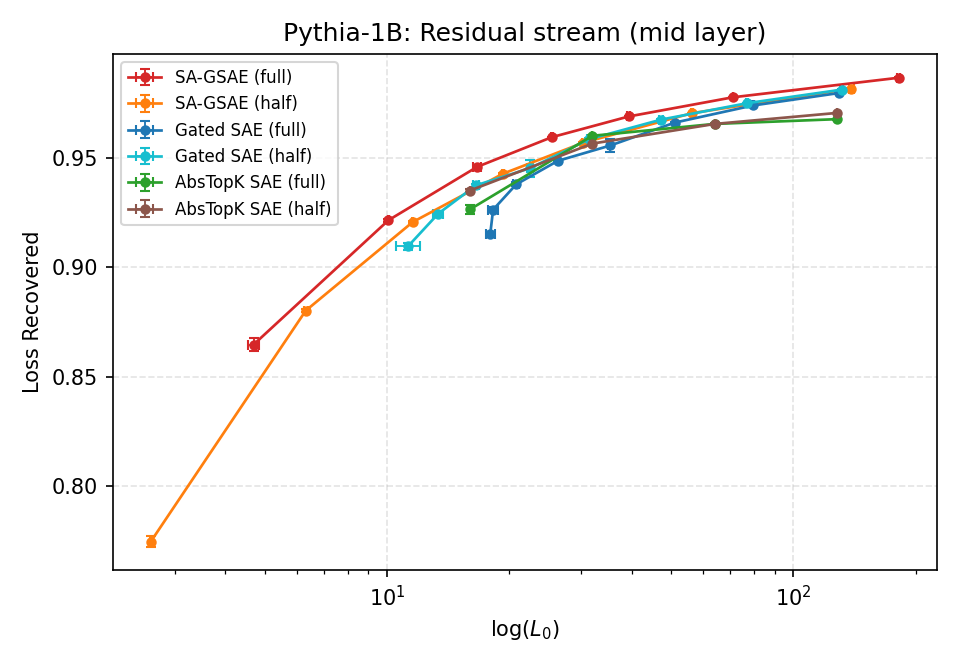}
    }{
        \fbox{\parbox[c][0.18\textwidth][c]{0.30\textwidth}{\centering Missing: pythia resid LR}}
    }

    \vspace{0.5em}

    \IfFileExists{llm_benchmarking_plots/smollm3_3b_mlp_out_mid/loss_recovered_vs_log_l0.png}{
        \includegraphics[width=0.32\textwidth]{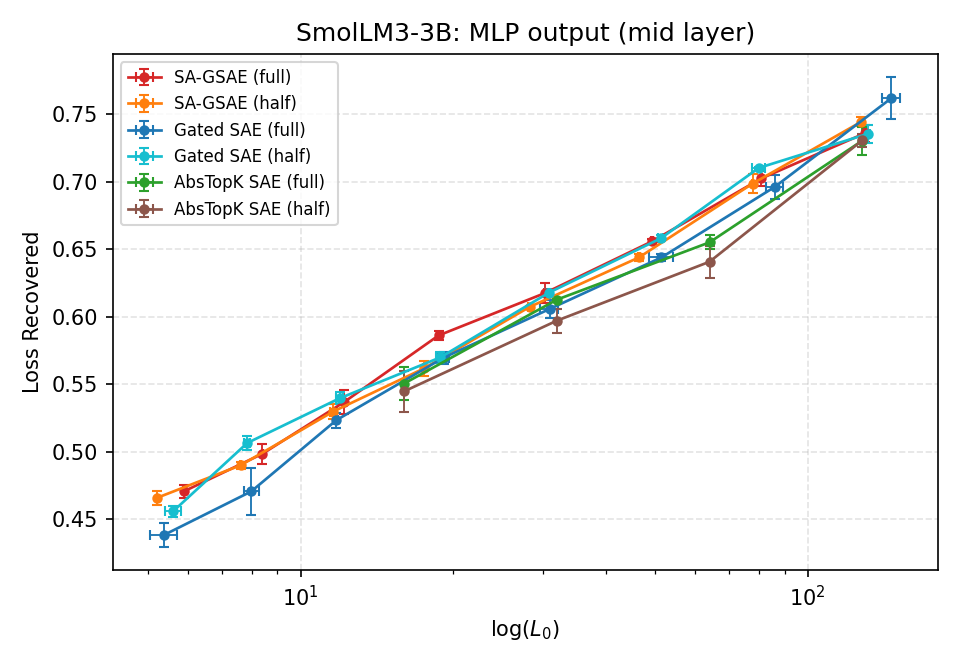}
    }{
        \fbox{\parbox[c][0.18\textwidth][c]{0.30\textwidth}{\centering Missing: SmolLM3 mlp\_out LR}}
    }
    \hfill
    \IfFileExists{llm_benchmarking_plots/smollm3_3b_attn_mid/loss_recovered_vs_log_l0.png}{
        \includegraphics[width=0.32\textwidth]{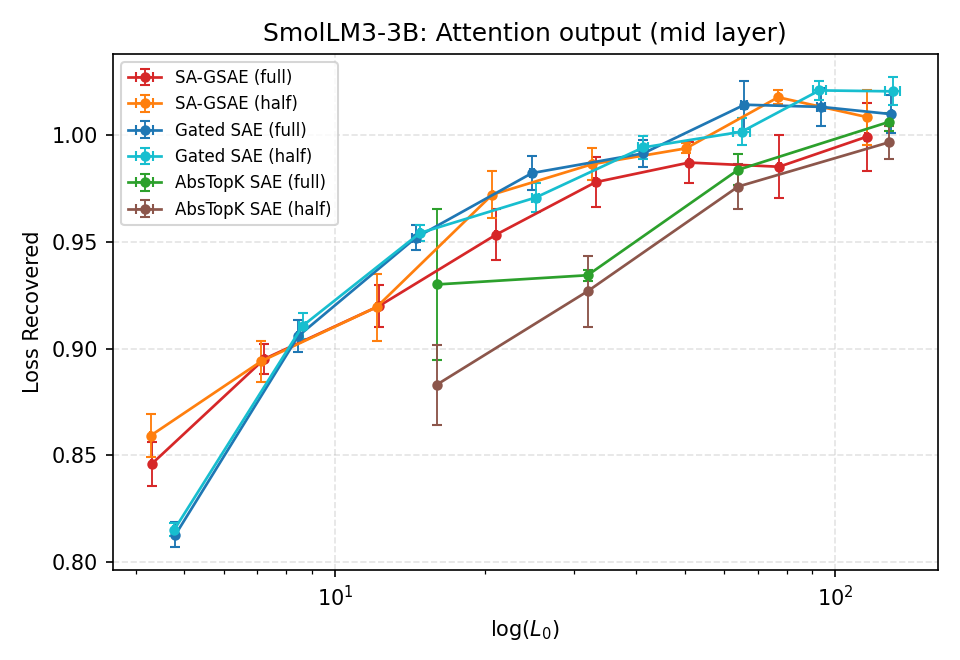}
    }{
        \fbox{\parbox[c][0.18\textwidth][c]{0.30\textwidth}{\centering Missing: SmolLM3 attn LR}}
    }
    \hfill
    \IfFileExists{llm_benchmarking_plots/smollm3_3b_resid_mid/loss_recovered_vs_log_l0.png}{
        \includegraphics[width=0.32\textwidth]{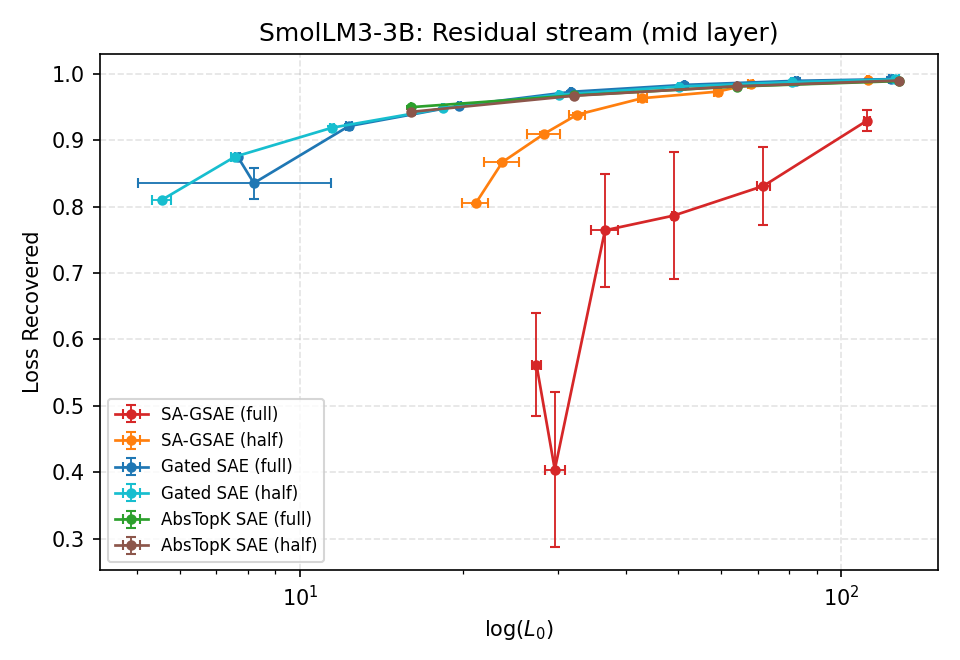}
    }{
        \fbox{\parbox[c][0.18\textwidth][c]{0.30\textwidth}{\centering Missing: SmolLM3 resid LR}}
    }
    \caption{Loss Recovered (LR) vs $\log(L_0)$ across the three mid-depth hookpoints (columns: \texttt{mlp\_out}, \texttt{attn}, \texttt{resid}). Top row: Pythia-1B (layer $8/16$). Bottom row: SmolLM3-3B (layer $18/36$). Error bars show $\pm$SE over 3 seeds on both axes. LR saturates near 1 at residual-stream hookpoints and has a noise-dominated ceiling on \texttt{attn}/SmolLM3-3B (baseline gap $0.037$ nats); see benchmark setup for the LR-not-cross-hookpoint-comparable caveat.}
    \label{fig:llm_loss_recovered_tradeoffs}
\end{figure*}

\paragraph{Dead-fraction/$L_0$ frontiers and mean-frontier dominance.}
We plot the full $\lambda$-sweep dead-fraction frontier on each of the six hookpoint $\times$ backbone cells in
\cref{fig:llm_dead_fraction_tradeoffs}, with the Loss-Recovered companion in
\cref{fig:llm_loss_recovered_tradeoffs} and the per-cell sweep statistics (overlap ranges, dominance fractions,
reduction ratios) tabulated in \cref{tab:pareto_sweep_summary}. Our top-level claim is curve-level rather than
point-level, and we phrase it as \emph{empirical mean-frontier dominance}: SA-half's aggregate mean curve is at
or better than Gated-full's on both metrics over the swept overlap. This is a statement about seed-averaged
curves, not a statistically certified per-point claim.

On every cell, the half-width SA-GSAE dead-fraction frontier lies at or below the full-width Gated SAE frontier
over essentially the entire swept sparsity range ($100\%$ of the overlap grid on five cells; $99.4\%$ on
\texttt{resid-mid/SmolLM3-3B}). On three of the six cells --
\texttt{mlp\_out-mid/Pythia-1B}, \texttt{mlp\_out-mid/SmolLM3-3B}, and \texttt{resid-mid/Pythia-1B} -- SA-half
simultaneously matches or exceeds Gated-full $R^2$ at every $L_0$ in the overlap. On those cells the sweep-median
dead-fraction reduction is $54\times$ to $73\times$; all ratios use a disclosed $0.01$ denominator floor, because
unfloored ratios blow up as the SA-half dead fraction approaches zero, and we treat absolute differences as
primary. The largest in-overlap absolute drops are $0.626 \to 4.3\!\cdot\!10^{-5}$ on
\texttt{resid-mid/Pythia-1B} ($L_0 \approx 125$) and $0.835 \to 8.4\!\cdot\!10^{-5}$ on
\texttt{mlp\_out-mid/SmolLM3-3B} ($L_0 \approx 121$, $\Delta R^2 = +0.015$).

On the remaining three cells the $R^2$ gap reverses at high $L_0$ -- Gated-full edges ahead by up to $0.011$ on
\texttt{attn-mid/Pythia-1B} and up to $0.025$ on \texttt{attn-mid/SmolLM3-3B} -- and the dead-fraction reductions
are more modest (sweep-median $1.5\times$ to $2.2\times$). SA-half still mean-frontier-dominates the low-$L_0$
region of both attention cells (up to $L_0 \approx 23$ on Pythia-1B and $L_0 \approx 39$ on SmolLM3-3B; slightly
further under a noise-tolerant criterion, \cref{tab:pareto_sweep_summary}). Across all six cells the seed-mean gap
stays within $\Delta R^2 \in [-0.025,\,+0.023]$ over the entire overlap (the $-0.025$ extremum being the high-$L_0$
\texttt{attn-mid/SmolLM3-3B} point above) and within $[-0.008,\,+0.022]$ at matched $L_0 = 64$ (\cref{tab:half_width}).

Two frontier-visible pathologies are worth flagging. First, on \texttt{mlp\_out-mid/Pythia-1B} the full-width
Gated SAE $L_0$ curve is non-monotone in $\lambda$ on $3/3$ seeds (drops to ${\sim}14$ at
$\lambda \approx 3.1\cdot 10^{-3}$, rebounds to ${\sim}19$--$21$), while full-width SA-GSAE is strictly
decreasing in $\lambda$ on $3/3$ seeds: the $L_0$-reversal pathology belongs to Gated SAE, not SA-GSAE. Second,
on \texttt{resid-mid/SmolLM3-3B} the full-width SA-GSAE frontier exhibits a reproducible reconstruction collapse
(MSE up to $6.0$, $R^2$ to $-4.3$, $3/24$ sweep points with LR${<}0.5$) that the reported half-width
configuration entirely avoids ($R^2 \ge 0.985$, MSE $\le 0.018$); per-seed audit in
\cref{app:full_width_results,tab:full_width_resid_smollm_per_seed}, training-dynamics diagnosis in
\cref{app:collapse_diagnosis}. Finally, AbsTopK reaches effectively zero dead on MLP and residual hookpoints but
is not a universally zero-dead baseline: on \texttt{attn-mid/Pythia-1B} it records $50.2\%$ dead at $k=64$,
rising to $72.5\%$ at $k=16$.

\paragraph{Matched-LR crossings on MLP outputs.}
LR is only discriminating on MLP-output cells (elsewhere it saturates; see benchmark setup). Reading
\cref{fig:llm_loss_recovered_tradeoffs} as a family of frontiers rather than at a fixed $L_0$: on
\texttt{mlp\_out-mid/Pythia-1B} the SA-half curve crosses LR $=0.75$ at $L_0 = 50.4 \pm 0.6$, MSE
$0.0418 \pm 0.0001$, $0.4\% \pm 0.0\%$ dead, whereas the Gated-full ($2H$) curve crosses the same LR target at
$L_0 = 58.8 \pm 1.7$, MSE $0.0431 \pm 0.0002$, $73.2\% \pm 0.6\%$ dead. On \texttt{mlp\_out-mid/SmolLM3-3B} the
SA-GSAE family tops out near LR $\approx 0.68$; at that common target, SA-half crosses at $L_0 = 67.6 \pm 3.1$,
MSE $0.00238 \pm 0.00003$, $0.1\% \pm 0.0\%$ dead, versus Gated-full at $L_0 = 77.0 \pm 3.4$, MSE
$0.00243 \pm 0.00003$, $83.0\% \pm 2.4\%$ dead.

\paragraph{Calibration symmetry and bipolar-latent usage.}
For each signed latent we fit split-regime slopes $\gamma_+,\gamma_-$ and report
$\hat p := \texttt{both\_fraction\_valid}$, the \emph{both-sign calibration coverage} of
\S\ref{sec:parameter_efficiency} (a capacity-usage statistic; neither a lower nor an upper bound on the semantic
anticorrelation rate $p$); full detail in
\cref{tab:gamma_symmetry,tab:bipolar_census_full,app:gamma_bipolar_full,app:gamma_symmetry_table,app:qualitative_latents}.
(i)~On every cell, SA-GSAE medians of $\gamma_+$ and $\gamma_-$ agree within $\le 0.05$ and $[p_{10},p_{90}]$
intervals overlap almost exactly -- direct real-activation evidence for the symmetric-magnitude default
$r_i^+ = r_i^-$ (ablation: $|\Delta R^2| = 0.0015$, $|\Delta\text{MSE}| = 0.0002$, \cref{tab:ablations}).
(ii)~AbsTopK attains median $\gamma \approx 1.00$ with $[p_{10},p_{90}] \approx [0.97,1.03]$ on every cell; this
follows mechanically from its lack of a learnable dead zone (the support-selection scale mismatch in
\cref{app:theory} cannot arise) and is not a calibration-training win. SA-GSAE's $\gamma_\pm$ medians of
$\approx 2$--$3$ reflect support-selection overshoot, not $L_1$ shrinkage. (iii)~Two bipolar statistics,
two hookpoint patterns: dictionary-wide coverage $\hat p$ concentrates on MLP outputs and
\texttt{resid-mid/Pythia-1B} -- SA-half $\hat p = 0.767$ on \texttt{mlp\_out-mid/SmolLM3-3B}, $0.481$ on
\texttt{resid-mid/Pythia-1B}, $0.283$ on \texttt{mlp\_out-mid/Pythia-1B}, versus $0.082$--$0.322$ on attention and
\texttt{resid-mid/SmolLM3-3B} (\cref{tab:bipolar_census_full}) -- exactly the cells where half-width SA-GSAE
mean-frontier dominates or ties Gated SAE (\cref{tab:half_width}); broad coverage drives the MLP-output capacity wins.
Top-$64$ qualitative bipolar latents, by contrast, concentrate at attention: $12/64$ ($19\%$) on
\texttt{attn-mid/Pythia-1B}, $3/64$ on \texttt{attn-mid/SmolLM3-3B}, $5/64$ on \texttt{resid-mid/Pythia-1B}, and
$0$--$1$ per file elsewhere (see \cref{app:qualitative_latents,tab:llm_latent_examples}, including the
Pythia \texttt{attn} latent $22641$ contrasting ``Parent Revolution'' with Murdoch-owned-media tokens). The
quantitative payoff is visible on the $\gamma_+$-valid/alive-fraction axis: SA-GSAE-full on
\texttt{mlp\_out-mid/Pythia-1B} reaches $0.651 \pm 0.002$, $2.5\times$ Gated SAE's
$0.264 \pm 0.004$, and is preserved at $0.642 \pm 0.001$ by SA-half.

\subsection{Half-Width Validation: SA-GSAE at $H$ Matches Gated SAE at $2H$}
\label{sec:half_width}

The halved-width operating point is the natural regime for SA-GSAE: if each sign-aware latent carries double the
information, the dictionary needs only half as many entries. We train SA-GSAE at $H = 16{,}384$ and compare it to a
Gated SAE at $2H = 32{,}768$ on identical cached activations, with the same optimizer, training schedule, and
evaluation protocol, on all six hookpoint $\times$ backbone cells. \Cref{tab:half_width} reports the resulting $\Delta R^2$,
$\Delta$MSE, and $\Delta$Dead (half-width SA-GSAE minus full-width Gated SAE) at matched $L_0 = 64$.

We separate two effects. \textit{(i) Sign-awareness at matched width.} Against Gated SAE at the same width $H$ (Gated-half), SA-half improves both $R^2$ and dead fraction on both \texttt{mlp\_out-mid} cells ($\Delta R^2 = +0.017, +0.013$; $\Delta$Dead $= -0.476, -0.518$) and effectively ties on \texttt{resid-mid/Pythia-1B} ($\Delta R^2 = +0.0001$, $\Delta$Dead $= -0.019$); it trades small $R^2$ for substantial dead-fraction reductions on both attention cells ($\Delta R^2 \in [-0.008, -0.006]$, $\Delta$Dead $\in [-0.568, -0.262]$); and it is dominated by Gated-half only on \texttt{resid-mid/SmolLM3-3B} ($\Delta R^2 = -0.002$, $\Delta$Dead $= +0.102$). Thus sign-awareness -- not width -- drives the $R^2$ and dead-fraction gains on the MLP-output cells. \textit{(ii) Operational-budget comparison.} Against the nominally $2\times$-wider Gated-full, SA-half matches reconstruction with $\Delta R^2 \in [-0.008, +0.023]$ (ties or exceeds on 3/6 cells; max negative gap $-0.008$ on \texttt{attn-mid/Pythia-1B}) while cutting dead fraction by $0.35$--$0.82$ absolute on all 6 cells. The halved-width operating-point recommendation is thus supported by both comparators jointly.

\begin{table}[!t]
    \centering
    \caption{Half-width validation at matched $L_0 = 64$ (mean over 3 seeds; $\pm$ SE shown on the Gated-full $\Delta R^2$ column; Gated-half comparator SEs are comparable, $\le 0.002$). We compare half-width SA-GSAE ($H = 16{,}384$) against both full-width Gated SAE ($2H = 32{,}768$, operational-budget comparison) and matched-width Gated SAE ($H$, which isolates sign-awareness) on every hookpoint $\times$ backbone cell.}
    \label{tab:half_width}
    \footnotesize
    \setlength{\tabcolsep}{3pt}
    \begin{tabular}{llcccccc}
        \toprule
        & & \multicolumn{3}{c}{SA-half $-$ Gated-full ($2H$)} & \multicolumn{3}{c}{SA-half $-$ Gated-half ($H$)} \\
        \cmidrule(lr){3-5}\cmidrule(lr){6-8}
        Backbone & Hookpoint & $\Delta R^2$ & $\Delta$MSE & $\Delta$Dead & $\Delta R^2$ & $\Delta$MSE & $\Delta$Dead \\
        \midrule
        \multirow{3}{*}{Pythia-1B}
        & \texttt{mlp\_out} & $+0.016 \pm 0.001$   & $-0.0024$  & $-0.735$ & $+0.017$  & $-0.0025$  & $-0.476$ \\
        & \texttt{attn}     & $-0.008 \pm 0.001$   & $+0.0003$  & $-0.349$ & $-0.008$  & $+0.0004$  & $-0.262$ \\
        & \texttt{resid}    & $+0.0005 \pm 0.0002$ & $-0.014$   & $-0.543$ & $+0.0001$ & $-0.0023$  & $-0.019$ \\
        \midrule
        \multirow{3}{*}{SmolLM3-3B}
        & \texttt{mlp\_out} & $+0.022 \pm 0.005$   & $-0.00010$ & $-0.820$ & $+0.013$  & $-0.00007$ & $-0.518$ \\
        & \texttt{attn}     & $-0.007 \pm 0.001$   & $+0.00002$ & $-0.622$ & $-0.006$  & $+0.00002$ & $-0.568$ \\
        & \texttt{resid}    & $-0.001 \pm 0.001$   & $+0.0008$  & $-0.413$ & $-0.002$  & $+0.0025$  & $+0.102$ \\
        \bottomrule
    \end{tabular}
\end{table}

\paragraph{Ablation summary.}
Appendix~\ref{sec:ablations} (\cref{tab:ablations}) isolates the design choices on Pythia-1B \texttt{mlp\_out} at
matched $L_0 = 64$. Removing the auxiliary loss is catastrophic (LR $= 0.27$, dead $= 98\%$, $L_0$ clamps at
${\sim}2$). Tying $r_i^+ = r_i^-$ is practically indistinguishable from the full model
($|\Delta R^2| = 0.0015$, $|\Delta\text{MSE}| = 0.0002$); corroborated by cell-wise $\gamma_\pm$ symmetry
(\cref{tab:gamma_symmetry}), by the exact threshold-tying reparameterization (\cref{app:theory}), and by the
half-width tied-unit ablations of \cref{app:rebuttal_studies} (tied $R^2 = 0.7193 \pm 0.0005$ vs.\ asymmetric
$0.7169 \pm 0.0004$ at matched $L_0$), we recommend the \emph{fully tied symmetric variant}
(\S\ref{sec:parameter_efficiency}) as the SA-GSAE default. The $\delta_0$ sweep is U-shaped
($\{10^{-3},10^{-2}\}$ harmful; $\{0.5,1.0\}$ neutral at this cell), and the Hybrid AbsTopK$+$gated-magnitude variant is
dominated by full SA-GSAE (MSE $0.0400$ vs.\ $0.0375$, $40\%$ vs.\ $0.8\%$ dead).

\section{Limitations}
\label{sec:limitations}

\textbf{Empirical scope.} Three hookpoints (\texttt{mlp\_out-mid}, \texttt{attn-mid}, \texttt{resid-mid}) $\times$
two backbones (Pythia-1B, SmolLM3-3B) $\times$ three seeds, at two widths. Layer sweeps, longer-context activations,
RLHF'd / instruction-tuned backbones, and $>$3B-parameter models are out of scope. A width sweep down to $H/4$
and below, which would test the parameter-efficiency prediction of \S\ref{sec:parameter_efficiency} at tighter
budgets, is a natural follow-up.

\textbf{Residual-stream LR saturation and cross-hookpoint incomparability.} On residual-stream cells the Gated-SAE
Loss-Recovered baseline gap is large ($6.6$/$8.6$ nats vs.\ $0.037$/$0.123$ nats on \texttt{mlp\_out} and
\texttt{attn}), making LR arithmetically compressible above $\sim 0.95$ and near-artifactual LR $> 1$ values
possible; we anchor residual-stream comparisons on dead fraction and $R^2$ rather than LR.

\textbf{Residual-stream collapse at full width on SmolLM3-3B.} MSE up to $6.0$, $R^2$ to $-4.3$, $3 / 24$ sweep
points with LR $< 0.5$ (\cref{tab:full_width_resid_smollm_per_seed}); the reported half-width SA-GSAE run avoids this
pathology on the same cell. A training-dynamics diagnosis (\cref{app:collapse_diagnosis}) locates the trigger at the
threshold-unfreeze step and rules out the warmup schedule and the auxiliary coefficient; it also shows that width alone
does not isolate the effect -- earlier half-width pilots without dead-latent threshold resets destabilized as well,
whereas every run with threshold resets enabled (either width) shows no collapse. The reported full- and half-width
configurations differ in this reset setting (\cref{tab:sa_config}), so the collapse contrast between them should not be
read as a pure width effect; small threshold initialization with dead-latent threshold resets is the robust
configuration we recommend.

\textbf{Semantic coherence of two-sided latents.} A post-submission blinded audit (\cref{app:rebuttal_studies}) finds
that judge-nameable semantic opposition between the two sides of a latent is rare at this scale for SA-GSAE \emph{and}
for all tested baselines (including near-antipodal Gated-SAE latent pairs); two-sided usage is primarily a capacity
mechanism, causally bidirectional under intervention, but not usually a nameable semantic axis. Interpretability
claims in this paper are scoped accordingly.

\textbf{Missing evaluation axes and baseline coverage.} The main benchmark reports reconstruction-capacity
metrics; the post-submission studies of \cref{app:rebuttal_studies} add a blinded semantic audit and
sign-conditioned causal interventions, both with controls. SAEBench-style probing evaluations (e.g., sae-probes,
RAVEL) remain absent and are planned as follow-up; we deliberately prioritized the two direct evaluations above
over targeted-probe-perturbation (TPP) and spurious-correlation-removal (SCR) metrics, whose reliability was
called into question by a 2026 audit of those protocols.
Baselines exclude Switch and Matryoshka
SAEs \cite{mudide2024switch,bussmann2025matryoshka}; sign-awareness targets sign splitting, not absorption or
hedging \cite{chanin2024absorption,chanin2025hedging}. Paired-duplicate and feature-ablation diagnostics are deferred.

\section{Conclusion}
SA-GSAE is \emph{two-sided gated sparsity with signed magnitude and auxiliary supervision}: ablations show the
gate and auxiliary reconstruction are the components the method cannot train without. Across 3 hookpoints $\times$ 2 backbones $\times$ 3 seeds, a
half-width SA-GSAE at $H$ cuts dead fraction by $0.35$--$0.82$ absolute at matched $L_0 = 64$ vs.\ a
full-width Gated SAE at $2H$, matches $R^2$ within $0.025$ on every cell, and empirically dominates the aggregate
mean frontier over the full swept $L_0$ overlap on 3 of 6 cells (\S\ref{sec:results_llm_benchmark}). We recommend the \emph{fully tied symmetric}
variant (single threshold, $r_i^+ = r_i^-$) at half width with small threshold initialization and dead-latent
threshold resets, a configuration that also avoids the reproducible \texttt{resid-mid/SmolLM3-3B} full-width
collapse (\cref{app:collapse_diagnosis}).

\clearpage
\FloatBarrier

\setlength{\bibsep}{2pt}
\renewcommand{\bibfont}{\small}
\bibliographystyle{unsrtnat}
\bibliography{references}

\clearpage

\appendix

\raggedbottom

\section{Background and Related Work}
\label{app:background}

\subsection{Sparse autoencoders as amortized sparse inference}
Solving \cref{eq:dict_learning} exactly requires an inner optimization to infer $z^{(n)}$ for each sample,
often via proximal methods (for example, ISTA/FISTA \cite{daubechies2004ista,beck2009fista}) that implement repeated
shrinkage steps.
Autoencoder-based approaches amortize this inference: an encoder network $f_\theta(x)$ predicts $z$ in a
single forward pass, and a decoder reconstructs $\hat{x} = D z + b$.
This ``learned inference'' view has a long history in sparse coding \cite{gregor2010lista}, and it becomes
particularly attractive when the dataset is large and the dictionary is highly overcomplete ($H\gg d_{\text{in}}$).

In the SAE setting used for mechanistic interpretability, the decoder is typically linear and the encoder is
a linear map followed by a sparsifying nonlinearity.
Popular choices include:
\begin{itemize}
    \item ReLU with an $L_1$ penalty on activations.
    \item Hard TopK masking, which keeps exactly $k$ active latents per sample \cite{makhzani2013ksparse}.
\end{itemize}
Recent work emphasizes that the training objective, the sparsity mechanism, and the normalization of
decoder columns jointly determine the learned feature geometry \cite{gao2024scaling}.

\subsection{Sparsity mechanisms and the shrinkage problem}
A core tension in SAE training is the sparsity-fidelity trade-off: increasing sparsity typically harms
reconstruction.
For $L_1$-regularized SAEs, the sparsity penalty biases the magnitudes of active latents downward,
an effect often called shrinkage.
This is not just a cosmetic issue: underestimating feature magnitudes can systematically distort the
reconstructed activation and degrade downstream ``loss recovered'' metrics \cite{gao2024scaling}.
Practical SAE training therefore involves broader trade-offs around shrinkage, dead latents, and how
sparsity is enforced. Recent work has accordingly explored gated objectives, thresholded activations,
fixed-$k$ selection, and broader evaluation suites
\cite{rajamanoharan2024gated, rajamanoharan2024jumprelu, bussmann2024batchtopk, karvonen2025saebench, paulo2025evaluating}.

Among these directions, Gated SAEs address shrinkage by decoupling deciding which latents are active from estimating
their magnitudes \cite{rajamanoharan2024gated}.
Concretely, a gating path computes pre-activations $\pi(x)$ that are penalized for sparsity, while a
separate magnitude path produces activations without paying an $L_1$ cost.
This separation yields Pareto improvements in reconstruction at a fixed average sparsity and has become a
common baseline in recent work \cite{rajamanoharan2024gated, karvonen2025saebench}.
Other approaches modify the sparsity nonlinearity directly, for example JumpReLU to improve
reconstruction fidelity at similar sparsity \cite{rajamanoharan2024jumprelu}, or BatchTopK to relax fixed
per-sample sparsity constraints \cite{bussmann2024batchtopk}. Evaluation work likewise argues that these
mechanisms should be compared across broader suites of reconstruction, interpretability, and utility metrics
\cite{karvonen2025saebench, paulo2025evaluating}. In principle these ideas could be combined:
Gated SAEs separate support from magnitude, threshold-learning methods sharpen support selection, BatchTopK
changes population sparsity control, whereas our focus is whether opposite polarities must occupy separate
latents at all.

\subsection{Non-negativity, bidirectional features, and sign-awareness}
Most widely used SAE variants in interpretability impose non-negativity on activations, either explicitly
(ReLU) or implicitly through their sparsity mechanism.
This interacts poorly with concepts that naturally live on a signed axis, where both positive and negative
evidence are meaningful.
A common workaround is sign splitting: represent a signed coefficient with two non-negative latents
corresponding to the positive and negative parts.
However, sign splitting doubles dictionary usage for every bidirectional axis, and in the SAE setting it
can fragment what would otherwise be a single coherent feature direction into two mutually exclusive units.

The literature has recently begun to address this limitation directly.
\cite{zhu2025abstopk} derive common SAE nonlinearities from proximal updates for sparse coding and argue
that non-negativity is an unnecessary constraint for recovering bidirectional features; they propose AbsTopK,
which selects latents by absolute magnitude to retain signed activations.
Our work is complementary: rather than modifying the selection rule alone, we incorporate sign-awareness
into the gated, no-shrinkage objective by designing a single latent that can gate and scale positive and
negative evidence along a shared decoder direction.
Indeed, \cite{zhu2025abstopk} note that their hard-thresholding principle could also be applied to JumpReLU,
introducing a threshold on both positive and negative activations, and leave such variants to future work;
Bi-Jump-ReLU realises this direction with learnable per-latent thresholds. Mechanistically, SA-GSAE stands to
AbsTopK as gated and JumpReLU-style SAEs stand to fixed-$k$ TopK SAEs
\cite{rajamanoharan2024gated, rajamanoharan2024jumprelu, gao2024scaling}: it replaces the fixed-$k$ signed
selector with learned two-sided thresholds (input-dependent, variable per-token sparsity) and a gated,
shrinkage-free magnitude path.

Conceptually, Bi-Jump-ReLU is closer to a signed dead-zone unit than to a proximal shrinkage map.
Related two-sided thresholded units also appear outside the SAE literature, for example in symmetric
threshold-linear networks and more recent symmetric-threshold ReLU constructions
\cite{hahnloser2003thresholdlinear, han2023strelu}. Our claim is not that a signed dead zone is new in the abstract
(and, to our knowledge, no prior SAE has used a two-sided jump activation),
but that it can be combined with decoder-aligned projections, gated no-shrinkage training, and auxiliary support
supervision to remove sign splitting in modern SAEs.
Classical soft-thresholding subtracts a threshold from the coefficient magnitude once a feature is active,
and symmetric thresholding rules typically use the same thresholded quantity to determine both support and
amplitude \cite{daubechies2004ista, beck2009fista}. JumpReLU similarly couples a jump threshold to a
one-sided active magnitude \cite{rajamanoharan2024jumprelu}. By contrast, our unit uses thresholds only
to decide whether a latent fires, while a separate unpenalized path determines how strongly
it contributes once active. This distinction is what lets us combine sign-awareness with the no-shrinkage
logic of Gated SAEs.

\subsection{Evaluating SAE feature dictionaries}
Because ground-truth internal features are rarely available, SAE quality is often assessed using proxy
metrics such as reconstruction error, sparsity ($L_0$), and loss recovered
\cite{gao2024scaling}.
Recent work argues that these proxies are necessary but insufficient, and introduces evaluations that tie
feature quality to interpretability and control.
Examples include controlled environments with ground-truth features \cite{karvonen2024measuring},
task-specific supervised dictionaries for comparison \cite{makelov2024principled}, robustness-focused
evaluations \cite{li2025illusions}, and multi-metric suites that emphasize trade-offs between
interpretability, disentanglement, and practical interventions \cite{karvonen2025saebench, paulo2025evaluating}.
Our experiments are designed to report standard sparsity-fidelity curves, but also to explicitly test
whether sign-aware latents reduce redundancy for anticorrelated pairs without sacrificing interpretability.

\section{Supplementary Theory and Method Details}
\label{app:theory}

\subsection{Training-hyperparameter asymmetry between SA-GSAE and Gated SAE}
\label{app:warmup_asymmetry}
SA-GSAE uses a two-sided threshold warmup of $40{,}000$ steps on Pythia-1B and $80{,}000$ steps on SmolLM3-3B
($17.8\%$ of the $225{,}000$/$450{,}000$-step training runs; thresholds unfreeze at step $40{,}001$/$80{,}001$ and
train for the remaining $82\%$), whereas the Gated baseline uses a $2{,}000$-step warmup. A post-submission
warmup-schedule ablation (\cref{app:collapse_diagnosis}) shows that on the one unstable cell the warmup
\emph{length} is not the cause of the \texttt{resid-mid/SmolLM3-3B} collapse: full-width runs destabilize at
whatever step threshold learning begins, for warmups of $0$, $40$k, and $80$k steps alike.

\subsection{Computational Complexity}

\textbf{Inference.} By caching the projection $t(x) = D^\top (x - b_{\text{dec}})$, the forward pass requires only one
matrix multiplication for the encoder, identical in structure to the baseline. The subsequent gating and Bi-Jump-ReLU
activations operate elementwise over the reduced sign-aware width $H_{\pm}=H(1-p/2)$, just as in the standard Gated SAE. Since $H_{\pm} \le H$, the total
inference FLOPs scale down by the factor $\left(1 - \frac{p}{2}\right)$ relative to a baseline with width $H$.

\textbf{Training.} The Gated SAE objective requires an additional decoder pass for the auxiliary loss. In a conventional
implementation this roughly adds one extra forward/backward pass through the decoder per optimization step. Relative to
a baseline SAE with the same target width, this increases decoder-side compute by a factor on the order of $1.5\times$ (the
exact factor depends on how encoder and decoder costs balance), and this factor is further reduced by the width
multiplier $\left(1 - \frac{p}{2}\right)$ that we gain from exploiting anticorrelated pairs. In other words, training
is modestly more expensive per update than a non-gated SAE with the same width, but yields a permanently smaller and
faster dictionary at inference time.

\subsection{Parameter-Savings Derivation}
\label{app:param_savings_derivation}

This subsection records the full derivation of the parameter-savings condition stated in
\S\ref{sec:parameter_efficiency}. Approximating the total parameter counts as
\begin{align}
    P_{\text{base}} &\approx H (d_{\text{in}} + c), \\
    P_{\text{sign}} &\approx H_{\pm} (d_{\text{in}} + c + 3) = H \left(1 - \frac{p}{2}\right) (d_{\text{in}} + c + 3),
\end{align}
we obtain the condition for net parameter savings:
\begin{align}
    P_{\text{sign}} < P_{\text{base}}
    &\;\Longrightarrow\; \left(1 - \frac{p}{2}\right) (d_{\text{in}} + c + 3) < d_{\text{in}} + c \\
    &\;\Longrightarrow\; 3 - \frac{p}{2}(d_{\text{in}} + c + 3) < 0 \\
    &\;\Longrightarrow\; p > \frac{6}{d_{\text{in}} + c + 3}.
\end{align}

\subsection{Why the Magnitude-Path ReLU Does Not Reintroduce Classical Shrinkage}

A natural concern is that SA-GSAE contains two support-affecting nonlinearities: the two-sided gate on $\pi_i(x)$ and the
ReLU in the magnitude path. Together they define effective positive and negative support thresholds on the shared
projection $t_i(x)$; the explicit formulas appear below in the discussion of effective thresholds. The key point is what
happens once a branch is active: on either sign, $a_i(x)$ is affine in $t_i(x)$, and the sparsity penalty acts only on
the gate, not on the active magnitude. So the extra ReLU can change which tokens activate, but it does not reintroduce
the subtractive bias term characteristic of soft-thresholding. This does not guarantee perfect calibration, because
support selection can still be imperfect, but it clarifies the sense in which SA-GSAE preserves the no-shrinkage logic
of gated SAEs.

\subsection{Method Details}

\paragraph{Implementation note.}
In our implementation, we stop gradients from the gate sparsity term and the auxiliary reconstruction from flowing into
the decoder dictionary. Concretely, the gate logits use $\pi_i(x)=\alpha_i \,\mathrm{stopgrad}(t_i(x))+\beta_i$, and the
auxiliary decoder path uses $D^{\text{sg}}$ and $b_{\text{dec}}^{\text{sg}}$ (a stop-gradient view of the current decoder
and bias). Following standard SAE practice, decoder columns are renormalized during training; the shared projection
$t_i(x)$ and the auxiliary path therefore both use the current normalized decoder. For numerical stability we
parameterize $\alpha_i=\exp(\log\alpha_i)$ and $g_i^\pm=\exp(r_i^\pm)$ and clamp the exponent inputs to a fixed range
(e.g., $[-20,20]$), and we enforce $\delta_i^\pm\ge 0$ via a ReLU/softplus parameterization (we use ReLU in code). This
preserves the intended separation: the auxiliary objective trains gate parameters without shaping the decoder geometry.

\textbf{Training dynamics:} The auxiliary reconstruction provides gradients only to the gate-related parameters
$(\alpha_i, \beta_i, \delta_i^+, \delta_i^-)$ via $\pi_i(x)$, while $D^{\text{sg}}$ (and $b_{\text{dec}}^{\text{sg}}$)
receive no gradients through $\mathcal{L}_{\text{aux}}$. The main reconstruction term updates both the decoder $D$ and
the magnitude parameters $(r_i^+, r_i^-, b_{\text{mag},i})$ via the signed activations $a_i(x)$. This cleanly separates
detection (learned through the auxiliary path and gate penalty) from magnitude estimation (learned through the main
reconstruction), mirroring the no-shrinkage argument for Gated SAEs in the non-signed case.

\paragraph{Support/magnitude separation and effective thresholds.}
Because $\alpha_i=\exp(\log \alpha_i)>0$, the positive branch can contribute only when both the gate and the magnitude path are active, i.e.,
\[
    t_i(x) > \max\!\left\{\frac{\delta_i^+ - \beta_i}{\alpha_i},\; -\frac{b_{\text{mag},i}}{g_i^+}\right\}.
\]
Analogously, the negative branch can contribute only when
\[
    t_i(x) < \min\!\left\{\frac{-\delta_i^- - \beta_i}{\alpha_i},\; \frac{b_{\text{mag},i}}{g_i^-}\right\}.
\]
Thus the gate thresholds and the magnitude-path ReLU jointly determine support. What they do not do is
subtract from the active magnitude: once a branch is on, $a_i(x)$ is affine in the shared projection $t_i(x)$ with slope
$\pm g_i^\pm$, and neither $\delta_i^\pm$ nor the sparsity penalty appears in that slope. In this sense the extra ReLU
can change which tokens activate, but it does not reintroduce the classical shrinkage effect of soft-thresholding,
where the active coefficient itself is biased toward zero. Our calibration analyses in Protocol A and on real LLM
activations are designed to test exactly this point.

\paragraph{On learnable thresholds and degenerate solutions.} A trivial failure mode for two-sided hinge penalties is
to ``solve'' the sparsity term by expanding the dead zone. In SA-GSAE this is not viable in isolation: when $s_i(x)=0$
we set $a_i(x)=0$, so the magnitude path cannot contribute unless the gate crosses a threshold. Increasing $\delta_i^\pm$
therefore tends to reduce reconstruction capacity unless compensated by shifting/scaling $\pi_i(x)$ so that truly
useful latents still cross their thresholds (in which case the hinge penalty depends on the threshold margins
$\pi_i(x)-\delta_i^+$ and $-\pi_i(x)-\delta_i^-$ rather than the absolute sizes of $\delta_i^\pm$). Accordingly, when
analyzing trained models we focus on threshold margins and effective thresholds rather than raw $\delta_i^\pm$ values.

\paragraph{Cost of packing unrelated features.}
Sign-sharing is not free for unrelated features. Consider two equally weighted, mutually exclusive unit feature
directions $u, v$ with cosine $c = u^\top v \le 0$. The best single unit direction $d$ minimizes the average
directional residual $1 - \tfrac12\big[(u^\top d)^2 + (v^\top d)^2\big] \ge 1 - \lambda_{\max}\!\big(\tfrac{uu^\top + vv^\top}{2}\big) = \tfrac{1+c}{2}$,
attained at the leading eigenvector. Exact antipodes ($c = -1$) incur zero residual, while orthogonal unrelated
features ($c = 0$) sacrifice half of their directional energy. The architecture therefore \emph{prefers} to pair
geometrically opposed directions, but under finite width and sparsity pressure it may still accept the residual cost
of packing less-opposed features; the blinded audit of \cref{app:rebuttal_studies} measures how often this happens
in practice.

\paragraph{Threshold tying is an exact reparameterization.}
Because the gate has a free bias, separate thresholds add no representational capacity: any pair
$(\delta_i^+, \delta_i^-)$ is exactly equivalent to a single symmetric threshold
$\theta_i = (\delta_i^+ + \delta_i^-)/2$ with shifted gate bias
$\beta_i' = \beta_i - (\delta_i^+ - \delta_i^-)/2$. Writing $\pi_i' = \alpha_i t_i + \beta_i'$, we have
$\pi_i - \delta_i^+ = \pi_i' - \theta_i$ and $-\pi_i - \delta_i^- = -\pi_i' - \theta_i$, so both gate conditions and
both hinge-penalty terms are preserved identically (our training objective uses no asymmetric threshold
regularizer). Threshold asymmetry can therefore only affect optimization, not the represented function class, and
is not claimed as a contribution.

\paragraph{Symmetric-magnitude and tied variants.}
An ablation of our architecture ties the per-polarity magnitude scaling parameters ($r_i^+=r_i^-$), while retaining
signed gating and asymmetric thresholds. In the parameter-counting model above, this reduces the per-latent scalar
overhead by one, i.e., the per-latent count becomes $d_{\text{in}} + c + 2$ instead of $d_{\text{in}} + c + 3$;
additionally tying the thresholds (exact reparameterization above) gives a fully tied unit with overhead
$d_{\text{in}} + c + 1$ and break-even $p > 2/(d_{\text{in}} + c + 1)$.
Empirically, on Protocol A with the default generator settings (LogNormal positives vs.\ Exponential negatives, $p_+=0.7$),
tying magnitudes leaves reconstruction essentially unchanged and yields near-ideal split-regime calibration. At our main
operating point ($\lambda=10^{-3}$; 16 seeds), the full model is marginally better calibrated on the positive regime,
while the symmetric-magnitude variant is marginally better on the negative regime; both remain close to the ideal $\gamma_\pm=1$.
On real LLM activations the fully tied unit matches (marginally exceeds) the asymmetric reference
(\cref{app:rebuttal_studies}).
This indicates that the dominant gain comes from enabling signed activations along a shared decoder direction, with
independent per-polarity parameterization acting as a small second-order refinement at best.

\section{Experimental Protocols}
\label{app:protocols}

We evaluate sign-awareness on three fronts:
\begin{enumerate}
    \item asymmetric signed-axis calibration,
    \item directional anomaly detection under a strict latent budget,
    \item consolidation of anticorrelated pairs in toy geometry.
\end{enumerate}
Unless otherwise stated, we report not only a single operating point, but also sparsity-fidelity trade-offs by sweeping
the sparsity strength (e.g., $\lambda$) and plotting reconstruction MSE versus average $L_0$ (active latents per example).

\subsection{Protocol A: The "Polarity Dial" (Asymmetry \& Calibration)}
\label{sec:protocol_a}

Protocol A is a controlled synthetic setting where each ground-truth feature is a signed scalar coefficient along a
random axis $u_j \in \mathbb{R}^d$. It tests whether an SAE can represent and calibrate both positive and negative
evidence along the same axis without duplicating decoder atoms. For non-negative SAEs, this requires paired atoms
$(+u_j,-u_j)$; a sign-aware latent can represent both directions with one decoder column and a signed activation.

\paragraph{Data generation.}
We construct a synthetic dataset $X \subset \mathbb{R}^d$ consisting of $k$ axes aligned to random directions
$u_j \in \mathbb{R}^d$. Each axis is active with probability $\rho$. If active, its sign $s_j \in \{-1, +1\}$ is sampled from
a Bernoulli distribution with parameter $p_+$, and the magnitude follows different distributions conditioned on the sign:
\begin{align}
    m_j \mid (s_j=+1) &\sim \text{LogNormal}(\mu=0, \sigma=0.5), \\
    m_j \mid (s_j=-1) &\sim \text{Exponential}(\lambda=1.5).
\end{align}
We deliberately use different positive and negative magnitude laws to break sign symmetry. The LogNormal positive branch
creates broader, heavier-tailed magnitudes, while the Exponential negative branch keeps the opposite sign more concentrated
near zero. The aim is to test whether one latent can share a decoder direction even when the two signs have different
scale statistics, not just different frequencies.
The input is
\begin{equation}
    x = \sum_j s_j m_j u_j + \epsilon,
\end{equation}
with small isotropic noise $\epsilon$.

\paragraph{Baselines and ablations.}
We compare the following model families:
\begin{itemize}
    \item \textbf{Sign-Aware Gated SAE (ours):} Signed latents with two-tailed gating (a learnable dead zone) and the gated
    objective to avoid shrinkage.
    \item \textbf{Sign-Aware, symmetric magnitude (ablation):} The same model, but with tied per-polarity magnitude scales
    ($r_i^+=r_i^-$). This tests whether independent per-polarity scaling is necessary in practice, and provides a more
    parameter-efficient variant.
    \item \textbf{Standard Gated SAE (non-negative, width $H$):} A conventional gated SAE with ReLU latents.
    \item \textbf{Standard Gated SAE (non-negative, width $2H$):} The same gated SAE trained at $2\times$ width. This
    controls for the representational requirement that non-negative models need paired atoms $(+u_j,-u_j)$ to represent a
    signed axis.
    \item \textbf{Baseline ReLU SAE (width $H$):} A standard SAE with ReLU latents.
    \item \textbf{Baseline ReLU SAE (width $2H$):} The same ReLU SAE trained at $2\times$ width.
    \item \textbf{Signed Soft-Threshold SAE (signed-latent baseline):} A signed-latent baseline using a soft-threshold
    nonlinearity (a standard choice in signed sparse coding). This baseline can represent negative coefficients without paired
    atoms, but exhibits the classical $L_1$ shrinkage trade-off between sparsity and coefficient bias (and can become dense
    at low $\lambda$), separating ``signed latents'' from the gated no-shrinkage objective.
\end{itemize}

\paragraph{Axis-aware alignment and pairing.}
To compare learned representations to ground-truth axes $u_j$, we evaluate at the axis level rather than treating
$u_j$ and $-u_j$ as distinct targets. We normalize decoder columns to unit norm for alignment, then compute signed cosine
similarities
\[
    s_{ij} = \cos(D_{:,i}, u_j),
\]
and define an axis similarity score $|s_{ij}|$. For sign-aware (signed-latent) models, we perform a one-to-one matching
between latents $i$ and axes $j$ via the Hungarian algorithm on the absolute similarities $|s_{ij}|$, and we record the sign
of $s_{ij}$ for diagnostics.
For non-negative baselines, we additionally allow two latents to represent one axis: for each axis $j$, we select the
best positively aligned latent $i^+(j)=\arg\max_i s_{ij}$ and the best negatively aligned latent $i^-(j)=\arg\min_i s_{ij}$
(subject to an alignment threshold). This credits baselines that learn paired atoms $(+u_j,-u_j)$, which is the expected
mechanism by which non-negative models represent signed structure.

\paragraph{Metrics.}
We focus on \textit{Split-Scale Calibration} at the axis level. For each ground-truth axis $j$, let
\begin{equation}
    c_j(x) = s_j m_j
\end{equation}
denote the ground-truth scalar coefficient along $u_j$ in the generative model. We define the model's scalar coefficient
estimate $\hat{c}_j(x)$ by projecting the relevant reconstructed contribution(s) onto $u_j$.
For sign-aware (signed-latent) models with a single matched latent $i(j)$, we use
\begin{equation}
    \hat{c}_j(x) = u_j^\top \big( a_{i(j)}(x)\, D_{:,i(j)} \big).
\end{equation}
For non-negative baselines using paired atoms $(i^+(j), i^-(j))$, we use an axis-aggregated estimate
\begin{equation}
    \hat{c}_j(x) = u_j^\top \Big( a_{i^+(j)}(x)\, D_{:,i^+(j)} + a_{i^-(j)}(x)\, D_{:,i^-(j)} \Big),
\end{equation}
which reduces to $\hat{c}_j(x)\approx a_{i^+(j)}(x)-a_{i^-(j)}(x)$ when $D_{:,i^+(j)}\approx u_j$ and
$D_{:,i^-(j)}\approx -u_j$.
We then compute separate least-squares rescaling factors for the positive and negative halves of the data:
\begin{align}
    \gamma_+ &= \arg\min_{\alpha} \E\big[(\alpha \hat{c}_j(x) - c_j(x))^2 \,\big|\, c_j(x) > 0\big], \\
    \gamma_- &= \arg\min_{\alpha} \E\big[(\alpha \hat{c}_j(x) - c_j(x))^2 \,\big|\, c_j(x) < 0\big].
\end{align}
\begin{itemize}
    \item \textbf{Ideal outcome:} $\gamma_+ \approx 1$ and $\gamma_- \approx 1$.
    \item \textbf{Primary hypothesis:} The Sign-Aware Gated SAE achieves $\gamma_+ \approx 1$ and $\gamma_- \approx 1$ using
    one latent per axis at width $H$. Non-negative models can approach this only if they allocate paired atoms; the $2H$
    baselines test whether remaining gaps at width $H$ are primarily representational (insufficient atoms) rather than
    optimization-related.
    \item \textbf{Secondary question:} Whether independent per-polarity magnitude scaling ($r_i^+ \neq r_i^-$) yields
    additional calibration gains beyond sign-awareness, tested via the symmetric-magnitude ablation.
    \item \textbf{Robustness diagnostics:} Because $\gamma_\pm$ can become ill-conditioned when $\hat{c}_j(x)$ collapses
    toward zero on a regime, we also report per-regime coefficient-space MSE, Pearson correlation between $\hat{c}_j(x)$ and
    $c_j(x)$, and explained variance ($R^2$), each computed separately on $\{c_j(x)>0\}$ and $\{c_j(x)<0\}$.
\end{itemize}

\subsection{Protocol B: Directional Deviation Detection (Resource Constrained)}
\label{sec:protocol_b}

In control systems and operations monitoring, deviations often carry directional semantics (e.g., ``pressure too high'' vs.\ ``pressure too low'').
Protocol B benchmarks the Sign-Aware Gated SAE as a direction-aware anomaly detector under a strict latent budget constraint,
and clarifies when such detection is feasible or fundamentally limited.

\paragraph{Setup.}
We simulate a high-dimensional sensor array $x \in \mathbb{R}^N$ with $N=1024$ channels. We train models under a strict
latent budget constraint $H < N$ and evaluate detection at a fixed false positive rate (FPR).

\paragraph{Normal distributions.}
We consider two normal regimes:
\begin{itemize}
    \item \textbf{Isotropic normal:} $x \sim \mathcal{N}(0, I)$.
    \item \textbf{Mildly heteroskedastic normal:} $x_j \sim \mathcal{N}(0, \sigma_j^2)$ with distinct $\sigma_j$ values
    close to 1. This breaks rotational symmetry while keeping channels approximately comparable, and is intended to make
    axis semantics learnable without strongly reweighting the reconstruction objective.
\end{itemize}

\paragraph{Anomalies.}
We evaluate two anomaly families:
\begin{itemize}
    \item \textbf{Coordinate (channel) anomalies:} choose a channel $j$ and shift $x_j \mapsto x_j \pm \Delta$.
    \item \textbf{Feature-direction anomalies:} choose a direction $v_k \in \mathbb{R}^N$ and shift $x \mapsto x \pm \Delta_k v_k$.
\end{itemize}
For feature-direction anomalies, directions $v_k$ are derived from a reference Sign-Aware model trained on the same normal
data. We select the top-$M$ directions by the empirical standard deviation of the reference model's detection signal (gate
preactivation) on a clean normal split, which prioritizes directions that produce a strong gate signal. To reduce width-dependent
effects, we optionally scale the input perturbation per direction using a fixed-$\Delta \pi$ rule:
\begin{equation}
    \Delta_k = \frac{\Delta}{\alpha_k + \epsilon},
\end{equation}
where $\alpha_k$ is the reference model's per-latent gate scale for the selected latent and $\epsilon$ is a small stabilizer.
This targets an approximately constant first-order shift in the reference latent's gate preactivation across selected directions.

\paragraph{Directional scoring and calibration.}
For the Sign-Aware model, let $\pi_i(x)$ denote the gate preactivation of latent $i$. We optionally normalize the
per-latent signal on validation normals to improve robustness under heteroskedastic or heavy-tailed normals:
\begin{equation}
    \tilde{\pi}_i(x) = \frac{\pi_i(x) - \mu_i}{\max(\sigma_i, \varepsilon)},
\end{equation}
where $(\mu_i,\sigma_i)$ are fitted on validation normals and $\varepsilon$ is a small floor. We then define threshold margins
for the positive and negative tails:
\begin{equation}
    z_i^+(x) = \text{ReLU}\big(\tilde{\pi}_i(x) - \tau_i^+\big), \qquad
    z_i^-(x) = \text{ReLU}\big(-\tilde{\pi}_i(x) - \tau_i^-\big),
\end{equation}
where $\tau_i^+$ and $\tau_i^-$ are per-latent thresholds (e.g., the 99th percentile of $\tilde{\pi}_i(x)$ and $-\tilde{\pi}_i(x)$
on validation normals). We pool these threshold margins across latents to form directional anomaly scores:
\begin{equation}
    \text{Score}_{\text{high}}(x) = \operatorname{Pool}_i\big[z_i^+(x)\big], \qquad
    \text{Score}_{\text{low}}(x)  = \operatorname{Pool}_i\big[z_i^-(x)\big],
\end{equation}
where $\operatorname{Pool}$ is a configurable aggregation (sum, max, or top-$k$ sum). We choose score thresholds
$\theta_{\text{high}}$ and $\theta_{\text{low}}$ to achieve a target FPR (1\%) on validation normals, and report recall
on held-out anomalies.

\paragraph{Baselines and leakage prevention.}
For baselines with non-negative activations, we use one-sided evidence signals and restrict directional scoring to avoid
``two-sided leakage'' from a single latent. Direction assignment uses decoder geometry (e.g., the sign of the dominant
decoder component) and scoring pools threshold margins within the assigned direction group.

\paragraph{Metrics.}
We report:
\begin{itemize}
    \item \textbf{Directional Recall at Fixed FPR:} recall for high and low anomalies at 1\% FPR.
    \item \textbf{Coverage:} the fraction of targets for which both-direction recall exceeds a preset threshold (default 0.8). Targets are channels for coordinate anomalies and directions for feature-direction anomalies.
    \item \textbf{Alignment diagnostics:} decoder channel-purity and the number of unique channels claimed by latents (argmax channel occupancy), used to detect random-rotation behavior and quantify axis alignment.
\end{itemize}

\paragraph{Protocol variants and expectations.}
Under isotropic normals, coordinate anomalies serve as a non-identifiability control. Under mild heteroskedasticity,
they test whether channel semantics emerge under strict compression. The main positive setting is feature-direction
anomalies with strength control and localized pooling.

\subsection{Protocol C: Geometric Validation on Toy Models}
\label{sec:protocol_c}

We use a geometry-first toy-model validation suite that makes the anticorrelated pair consolidation behavior visually
explicit in a controlled low-dimensional setting. Following \cite{elhage2022toy}, we generate $k$ antipodal feature pairs
in $\R^2$ and train:
\begin{itemize}
    \item a non-negative baseline at width $M=2k$;
    \item a Sign-Aware SAE at width $M=k$ (with an optional overcomplete Sign-Aware variant at $M=2k$).
\end{itemize}

Protocol C includes:
\begin{itemize}
    \item \textbf{Unit-circle geometry:} plot decoder directions against the ground-truth antipodal directions.
    \item \textbf{Bi-jump histogram:} plot the activation histogram for a representative sign-aware latent to visualize signed pair membership.
    \item \textbf{Superposition sweep:} vary the number of active pairs per sample $n_{\mathrm{active}}$ and measure signed recoverability as well as a thresholded Pair Consolidation Rate.
    \item \textbf{Robustness sweep:} vary within-pair correlation $\rho \in [-1,0]$ at fixed $n_{\mathrm{active}}=1$.
\end{itemize}

\section{Results for Protocols A-C}
\label{app:protocol_results}

This appendix reports controlled results for asymmetric signed-axis calibration (Protocol~A), two-tailed directional
detection under strict compression (Protocol~B), and antipodal-pair consolidation in toy models (Protocol~C).
Experiments for Protocols~A-C closed in approximately 60 compute hours of AWS g4dn.xlarge instances (NVIDIA T4 GPU).

\subsection{Protocol A: Asymmetry and Split-Scale Calibration}
\label{sec:results_protocol_a}

\paragraph{Setup.}
We evaluate Protocol A on the synthetic signed-axis dataset described in \S\ref{sec:protocol_a} with
$d=512$ input dimensions and $k=128$ ground-truth axes. Each axis is active with probability $\rho=0.05$ and, when active,
takes a positive sign with probability $p_+=0.7$ and a negative sign otherwise. Conditional on sign, magnitudes are drawn from
a LogNormal$(\mu=0,\sigma=0.5)$ distribution for positives and an Exponential$(\lambda=1.5)$ distribution for negatives, with
additive isotropic noise $\epsilon \sim \mathcal{N}(0, 0.1^2 I)$.
We generate 200k training samples, 20k validation samples, and 20k test samples. All models are trained for 50 epochs with
Adam \cite{kingma2014adam} (learning rate $10^{-4}$, $\beta=(0.9,0.999)$) and batch size 1024. We sweep the sparsity coefficient $\lambda$ over
64 log-spaced values in $[10^{-5},10^{-2}]$; the table below reports a representative sparse operating point at $\lambda=10^{-3}$.
We report mean $\pm$ standard deviation over 16 random seeds (0-15), except for the Signed Soft-Threshold baseline, which
uses the corrected implementation and is aggregated over 8 seeds (0-7) (marked $\dagger$ in \cref{tab:protocol_a_results}).

\paragraph{Evaluation protocol.}
We evaluate calibration at the axis level using the alignment and aggregation procedure in \S\ref{sec:protocol_a}.
Concretely, we match axes using absolute cosine similarity between decoder columns and ground-truth directions, with a minimum
alignment threshold of $\tau=0.9$. For non-negative baselines, we use the paired-atom aggregation $\hat{c}_j(x)$ when both a
positively aligned and negatively aligned latent are available for a given axis. This gives non-negative models credit for the
intended representation $(+u_j,-u_j)$ and ensures the comparison is not biased against baselines by the evaluation metric.
Across the sparsity sweep, we find that axis matching under the strict threshold $\tau=0.9$ can degrade when sparsity pressure
is too weak (consistent with rotational non-identifiability); we therefore focus our main conclusions on the sparse regime where
axes are reliably matched.

\paragraph{Metrics.}
We report reconstruction MSE on the test set, the average number of non-zero latents per example ($L_0$), and split-regime
calibration slopes $\gamma_+$ and $\gamma_-$ together with their mean absolute errors. Because $\gamma_\pm$ can become unstable
when a model collapses $\hat{c}_j(x)$ toward zero on one regime, we additionally compute per-regime coefficient-space MSE,
Pearson correlation, and $R^2$; these diagnostics support the same conclusions and are included in the released artifacts.

\begin{table}[t]
    \centering
    \caption{Protocol A results at $\lambda=10^{-3}$ (mean $\pm$ std). $H$ denotes the SAE width. $L_0$ denotes the average number
    of non-zero latents per example. $\bar{\gamma}_\pm$ is the mean split-regime calibration slope across axes, and
    $\mathbb{E}[|\gamma_\pm-1|]$ is the mean absolute slope error (lower is better). Non-negative baselines are evaluated with
    axis-level aggregation that credits paired atoms $(+u_j,-u_j)$ when present (\S\ref{sec:protocol_a}). All rows use 16 seeds
    (0-15) except Signed Soft-Threshold ($\dagger$), which uses the corrected implementation and is aggregated over 8 seeds (0-7).}
    \label{tab:protocol_a_results}
    \small
    \begin{tabular}{lcc}
        \toprule
        Model & MSE$\downarrow$ & $L_0\downarrow$ \\
        \midrule
        ReLU SAE ($H=128$)
            & 0.018035 $\pm$ 0.001097 & 6.60 $\pm$ 0.33 \\
        ReLU SAE ($H=256$)
            & 0.017257 $\pm$ 0.000587 & 7.08 $\pm$ 0.64 \\
        Gated SAE ($H=128$)
            & 0.012981 $\pm$ 0.000047 & 9.36 $\pm$ 0.35 \\
        Gated SAE ($H=256$)
            & \textbf{0.009956 $\pm$ 0.000008} & 9.22 $\pm$ 0.11 \\
        Signed Soft-Threshold SAE$^{\dagger}$ ($H=128$)
            & 0.011087 $\pm$ 0.000006 & 8.84 $\pm$ 0.04 \\
        Sign-Aware Gated SAE ($H=128$)
            & 0.009963 $\pm$ 0.000007 & 9.20 $\pm$ 0.07 \\
        Sign-Aware Gated SAE, symmetric magnitude ($H=128$)
            & 0.009964 $\pm$ 0.000007 & 9.20 $\pm$ 0.09 \\
        \bottomrule
    \end{tabular}

    \vspace{0.5em}

    \begin{tabular}{lcc}
        \toprule
        Model & $\bar{\gamma}_+$ & $\bar{\gamma}_-$ \\
        \midrule
        ReLU SAE ($H=128$) & 0.773 $\pm$ 0.103 & -1.554 $\pm$ 1.174 \\
        ReLU SAE ($H=256$) & 0.889 $\pm$ 0.055 & -2.087 $\pm$ 1.563 \\
        Gated SAE ($H=128$) & 0.588 $\pm$ 0.063 & -0.051 $\pm$ 0.011 \\
        Gated SAE ($H=256$) & 1.017 $\pm$ 0.018 & 1.037 $\pm$ 0.018 \\
        Signed Soft-Threshold SAE$^{\dagger}$ ($H=128$) & 1.243 $\pm$ 0.001 & 1.256 $\pm$ 0.002 \\
        Sign-Aware Gated SAE ($H=128$) & 1.039 $\pm$ 0.006 & 1.055 $\pm$ 0.005 \\
        Sign-Aware Gated SAE, symmetric magnitude ($H=128$) & 1.042 $\pm$ 0.005 & 1.041 $\pm$ 0.005 \\
        \bottomrule
    \end{tabular}

    \vspace{0.5em}

    \begin{tabular}{lcc}
        \toprule
        Model & $\mathbb{E}[|\gamma_+ - 1|]\downarrow$ & $\mathbb{E}[|\gamma_- - 1|]\downarrow$ \\
        \midrule
        ReLU SAE ($H=128$) & 0.504 $\pm$ 0.064 & 2.558 $\pm$ 1.175 \\
        ReLU SAE ($H=256$) & 0.426 $\pm$ 0.037 & 3.089 $\pm$ 1.562 \\
        Gated SAE ($H=128$) & 0.440 $\pm$ 0.061 & 1.051 $\pm$ 0.011 \\
        Gated SAE ($H=256$) & 0.050 $\pm$ 0.017 & 0.070 $\pm$ 0.017 \\
        Signed Soft-Threshold SAE$^{\dagger}$ ($H=128$) & 0.243 $\pm$ 0.001 & 0.256 $\pm$ 0.002 \\
        Sign-Aware Gated SAE ($H=128$) & \textbf{0.039 $\pm$ 0.006} & 0.055 $\pm$ 0.005 \\
        Sign-Aware Gated SAE, symmetric magnitude ($H=128$) & 0.042 $\pm$ 0.005 & \textbf{0.041 $\pm$ 0.005} \\
        \bottomrule
    \end{tabular}
\end{table}

\begin{figure}[t]
    \centering
    \includegraphics[width=0.95\linewidth]{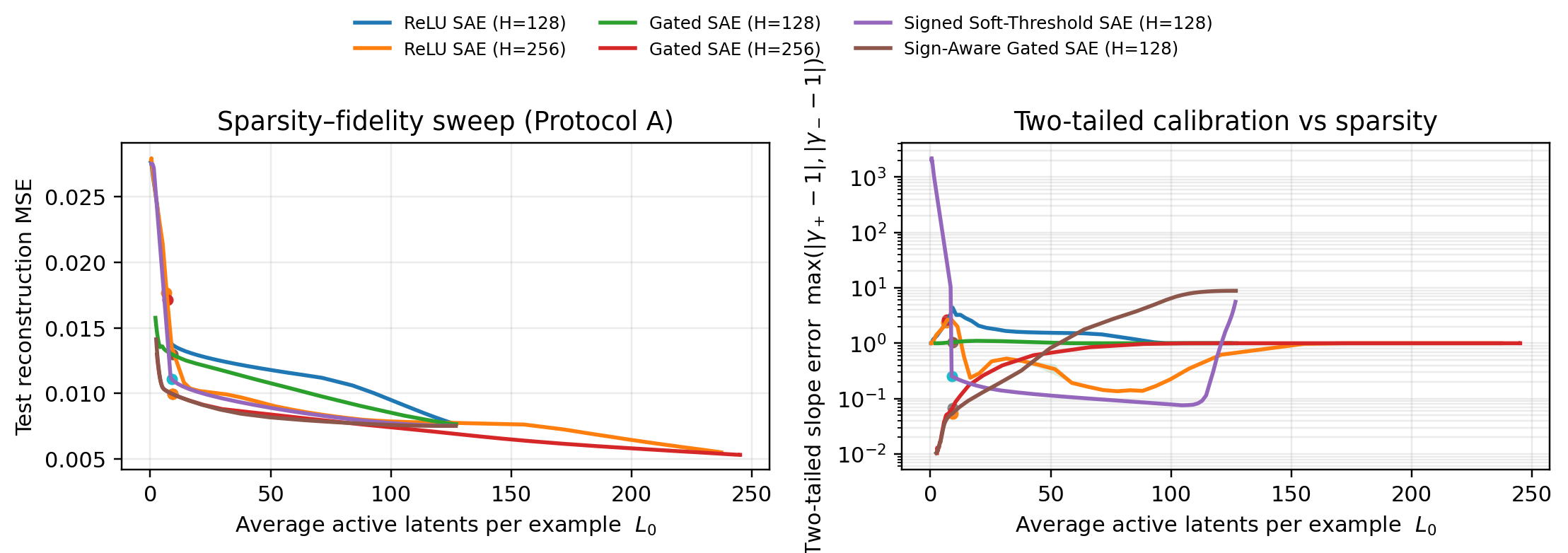}
    \caption{Protocol A (Polarity Dial) across the full sparsity sweep. Left: reconstruction MSE vs. average $L_0$. Right: two-tailed split-regime calibration error $\max(|\gamma_+ - 1|,|\gamma_- - 1|)$ (log-scale y-axis) vs. $L_0$. Markers highlight the operating point $\lambda=10^{-3}$ used in Table~\ref{tab:protocol_a_results}. Shaded bands show $\pm$SE over seeds.}
    \label{fig:protocol_a_sweep}
\end{figure}

Figure~\ref{fig:protocol_a_sweep} shows the full sparsity sweep, making explicit that the sign-aware model matches the $2\times$ width
non-negative gated baseline across a broad $L_0$ range while maintaining low two-tailed calibration error.

\paragraph{The gap comes from needing two non-negative latents per signed axis, and sign-awareness closes it.}
At width $H=128$, non-negative models cannot allocate paired atoms $(+u_j,-u_j)$ for all axes, and the deficit shows up as a
systematic failure to calibrate one side of the distribution. The standard Gated SAE achieves moderate positive-side calibration
($\bar{\gamma}_+ \approx 0.59$) but collapses the negative side ($\bar{\gamma}_- \approx -0.05$), yielding a large negative-regime
slope error ($\mathbb{E}[|\gamma_- - 1|]\approx 1.05$). The ReLU SAE baselines perform substantially worse, reflecting both
the same non-negativity constraint and shrinkage under the standard $L_1$ objective.

When we double the width of the non-negative Gated SAE to $H=256$, it can represent each signed axis with two latents and
recovers near-ideal calibration on both regimes ($\bar{\gamma}_+\approx 1.02$, $\bar{\gamma}_-\approx 1.04$).
At width $H=128$, the Sign-Aware Gated SAE reaches comparable calibration quality using one latent per axis,
while also achieving slightly lower mean absolute slope errors at this operating point.
Its reconstruction error (0.009963) is within 0.1\% of the $H=256$ non-negative gated baseline (0.009956), despite using half
the dictionary width. This resolves the anticorrelation efficiency gap in the intended sense.

\paragraph{Signed activations alone are not sufficient without the gated objective.}
The Signed Soft-Threshold SAE can represent negative coefficients without paired atoms and, with the corrected implementation,
can be tuned to match the sparsity of the gated models (e.g.\ $L_0\approx 8.8$ at $\lambda=10^{-3}$; \cref{tab:protocol_a_results},
$\dagger$). However, it exhibits the classical $L_1$ shrinkage bias: at this sparse operating point its split-regime slopes
deviate systematically from the ideal value $1$ ($\bar{\gamma}_+\approx 1.24$, $\bar{\gamma}_-\approx 1.26$), yielding substantially
larger mean absolute slope error than the gated no-shrinkage models. Coefficient-space diagnostics show high correlation but degraded
scale accuracy, and reconstruction remains worse than the sign-aware and $2\times$ width gated baselines at comparable sparsity.
Across the $\lambda$ sweep, soft-threshold exhibits a pronounced sparsity-bias trade-off: calibration improves in dense regimes,
but degrades as $\lambda$ increases; moreover, we observe a sharp instability/collapse for slightly larger $\lambda$ as many axes
become nearly inactive. This supports the motivation for combining sign-awareness with the gated no-shrinkage training objective
rather than adopting signed sparse coding activations in an otherwise standard SAE.

\paragraph{Independent $r^+,r^-$ scaling is not a dominant factor at these settings.}
Tying the per-polarity magnitude scales ($r_i^+=r_i^-$) leaves reconstruction essentially unchanged and maintains near-ideal
split-regime calibration. Under the default (asymmetric) Protocol A generator, we observe a small but consistent trade-off:
the full model is marginally better calibrated on the positive regime (lower $\mathbb{E}[|\gamma_+ - 1|]$), while the
symmetric-magnitude variant is marginally better on the negative regime (lower $\mathbb{E}[|\gamma_- - 1|]$). Overall, these
differences are minor compared to the gap between sign-aware and non-sign-aware baselines, suggesting that independent
per-polarity scaling is a second-order refinement rather than the primary driver of the Protocol A gains. Accordingly, the
symmetric-magnitude variant is a reasonable default when prioritizing parameter minimality, while retaining the full model as
a strictly more expressive option for settings with more extreme sign-conditioned asymmetry.

\subsection{Protocol B: Directional Deviation Detection}
\label{sec:results_protocol_b}

\paragraph{Coordinate anomalies on isotropic normals are undetectable under strict compression.}
On isotropic normals $x \sim \mathcal{N}(0, I)$ with single-channel shifts, all evaluated models achieve directional
recall close to the calibrated FPR (roughly 1\% to 2\% recall at 1\% FPR) and 0 coverage across widths $H \in \{32,64,128,256,512\}$.
Decoder alignment diagnostics are consistent with a random-rotation regime (low channel-purity and argmax-channel
occupancy matching random assignment). This matches the expected non-identifiability: when the normal distribution is
rotationally symmetric, no coordinate system is preferred, so single-channel shifts do not consistently create large
per-latent tail events in the learned basis.

\paragraph{Symmetry-breaking improves alignment but does not rescue coordinate monitoring.}
We next break rotational symmetry using heteroskedastic normals with distinct channel variances, and enable per-latent
z-score normalization and max pooling. While alignment diagnostics shift in the expected direction (higher decoder
channel-purity and non-random argmax-channel occupancy), directional recall remains near the calibrated FPR and coverage
remains 0 across all widths. Empirically, even the best channels detect only a small fraction of anomalies (well below
the coverage threshold). This suggests that for independent channels, strict dimensionality reduction $H < N$ can make
per-channel monitoring infeasible without additional shared structure (e.g., correlations or low-rank factors) that
allows compression without losing channel-level anomaly information.

\paragraph{Feature-direction anomalies yield strong two-tailed detection for the Sign-Aware model.}
We then evaluate feature-direction anomalies aligned to operational axes derived from a reference Sign-Aware model.
We select $M=128$ directions by the standard deviation of the reference model's gate preactivation on validation normals
(top-$\mathrm{std}(\pi)$), and scale per-direction input perturbations using a fixed-$\Delta \pi$ rule to reduce width-dependent
effects. We calibrate thresholds at 1\% FPR and use localized pooling (max or top-$k$).

\begin{table}[t]
    \centering
    \caption{Protocol B results for feature-direction anomalies (mean over 16 seeds, target FPR 1\%). Directions: $M=128$ selected by top-$\mathrm{std}(\pi)$ from a reference Sign-Aware model; per-direction inputs scaled by a fixed-$\Delta\pi$ rule and gate signals z-scored on validation normals before thresholding/pooling. Coverage denotes the fraction of directions with both-direction recall $\ge 0.8$.}
    \label{tab:protocol_b_feature_direction}
    \small\begin{tabular}{lcccc}
        \toprule
        Width $H$ & Pooling & Sign-Aware Recall (high) & Sign-Aware Recall (low) & Coverage \\
        \midrule
        128 & max & 0.999 & 0.999 & 1.00 \\
        256 & max & 0.998 & 0.999 & 1.00 \\
        512 & max & 0.998 & 0.998 & 1.00 \\
        \midrule
        512 & top-$k$ ($k=4$) & 0.971 & 0.970 & 1.00 \\
        \bottomrule
    \end{tabular}
\end{table}

Across all widths, the Sign-Aware model achieves near-saturated directional recall and full direction coverage with
max pooling (\cref{tab:protocol_b_feature_direction}). Non-sign-aware baselines remain near chance on this
evaluation (directional recall close to the calibrated FPR and 0 coverage), consistent with their one-sided evidence and
the difficulty of representing both tails of a direction efficiently under a fixed latent budget.

\begin{figure}[t]
    \centering
    \includegraphics[width=0.98\linewidth]{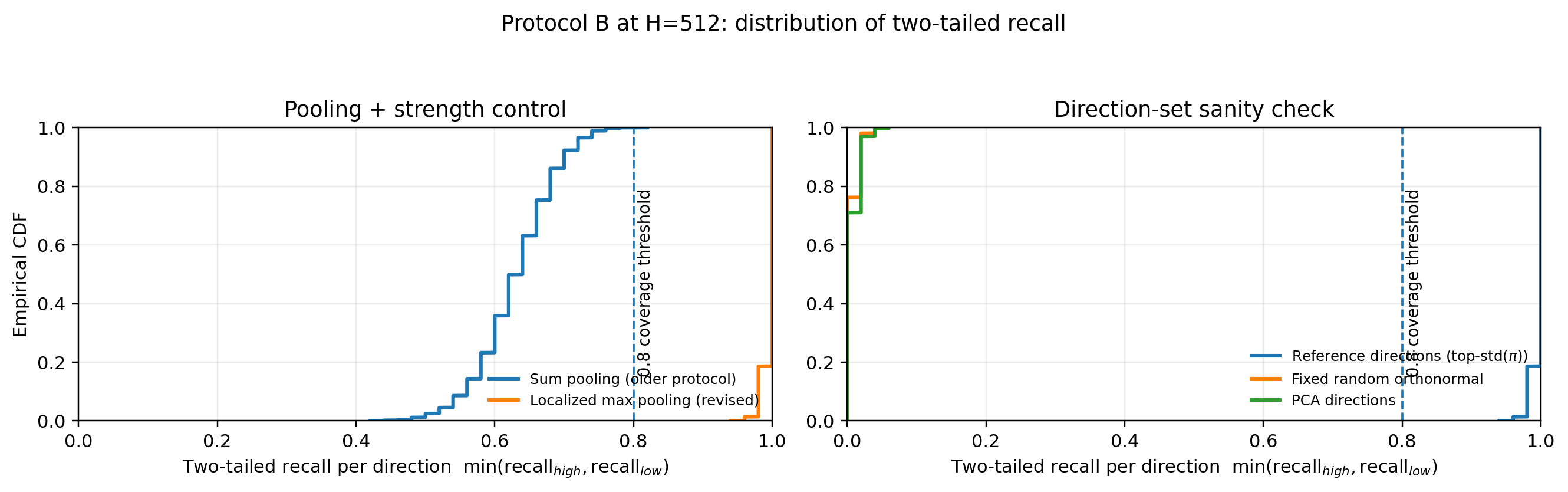}
    \caption{Protocol B (feature-direction anomalies) at $H=512$: empirical CDF of per-direction two-tailed recall, $\min(\mathrm{recall}_{\mathrm{high}}, \mathrm{recall}_{\mathrm{low}})$, across directions and seeds. Left: the revised protocol with fixed-$\Delta\pi$ scaling, z-score normalization and localized max pooling concentrates above the 0.8 coverage threshold, while the earlier sum pooling variant degrades substantially. Right: direction-set sanity check under isotropic normals: when anomaly directions are fixed random orthonormal or PCA directions (independent of the reference), recall collapses to near-chance and coverage goes to zero, reflecting rotational non-identifiability.}
    \label{fig:protocol_b_h512_min_recall_cdf}
\end{figure}

\paragraph{Unbiased direction sets are not detectable under isotropic normals.}
We repeat the feature-direction evaluation using:
\begin{itemize}
    \item a fixed random orthonormal basis;
    \item PCA directions of the normal data.
\end{itemize}
Under isotropic normals these direction sets have no privileged alignment to a learned basis; all models (including Sign-Aware)
achieve recall close to the calibrated FPR ($\approx$1-2\%) and 0 coverage across widths, even when we give non-sign-aware
baselines $2\times$ dictionary width. \cref{fig:protocol_b_h512_min_recall_cdf} (right) visualizes this non-identifiability at $H=512$.

An earlier feature-direction variant without strength control and using global sum pooling degrades sharply at larger
widths: at $H=512$, the Sign-Aware model drops to roughly $0.67$ directional recall with near-zero coverage (16 seeds),
motivating the fixed-$\Delta\pi$ scaling and localized pooling used above. \cref{fig:protocol_b_h512_min_recall_cdf}
visualizes this effect at $H=512$ by plotting the empirical CDF of per-direction two-tailed recall.

\subsection{Protocol C: Geometric Validation on Toy Models}
\label{sec:results_protocol_c}

\paragraph{Setup.}
We evaluate Protocol C described in \S\ref{sec:protocol_c}, replicating the setup of \cite{elhage2022toy} to provide
visual and geometric confirmation that the Sign-Aware SAE correctly identifies and consolidates anticorrelated subspaces.

\paragraph{Data Geometry.}
We generate $k$ pairs of features embedded in $\R^2$ (the "unit circle" visualization), with paired directions that are antipodal on the unit circle.
For the qualitative visualizations below, we use the mutually exclusive setting ($\rho=-1$) so that exactly one member of each pair can be active at a time.
We train models with dictionary width $M=k$ (Sign-Aware) and $M=2k$ (Baseline), so that in the ideal case the baseline
allocates one latent per feature, while the Sign-Aware model allocates one latent per pair.
Unless stated otherwise, we use $k=8$ for the qualitative visualizations and within-pair robustness sweep, and $k=16$ for the superposition sweep.

\paragraph{Why Sign-Aware Consolidates Antipodal Pairs.}
In the antipodal construction, each ground-truth pair corresponds to a single axis direction $u_j \in \R^2$, with two features at directions $\pm u_j$.
In the mutually exclusive setting ($\rho=-1$), an ideal Sign-Aware SAE can represent the entire pair using a single decoder column aligned with $u_j$:
positive activations correspond to the $+u_j$ feature and negative activations correspond to the $-u_j$ feature.
A non-negative SAE cannot use the sign of activations to flip direction, so representing both $\pm u_j$
requires allocating two separate decoder columns, one per feature direction.

\paragraph{Visualization A: The Unit Circle.}
We project the columns of the decoder $D$ onto the 2D data plane and plot them on the unit circle.

\begin{figure}[t]
    \centering
    \includegraphics[width=0.48\linewidth]{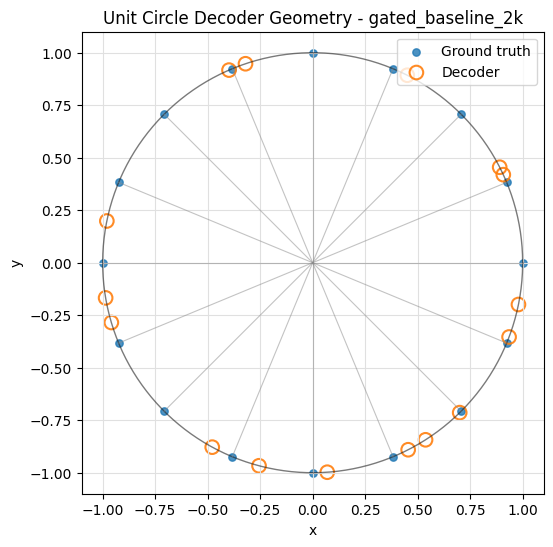}
    \hfill
    \includegraphics[width=0.48\linewidth]{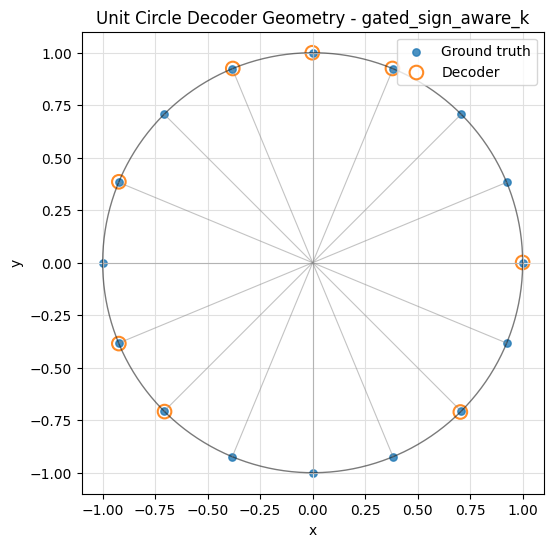}
    \caption{Protocol C qualitative geometry. Blue points mark the ground-truth feature directions on the unit circle, and orange points mark learned decoder columns projected into the same plane. In the mutually exclusive antipodal setting ($\rho=-1$), a non-negative baseline at width $M=2k$ learns one decoder direction per feature, while the Sign-Aware SAE at width $M=k$ learns one decoder direction per pair axis and uses activation sign to distinguish the two antipodal features. Exact overlap is not expected because training only recovers the generating axes approximately.}
    \label{fig:protocol_c_unit_circle}
\end{figure}

\begin{itemize}
    \item \textbf{Baseline Observation:} Two distinct vectors at $180^\circ$ separation (antipodal) for each feature pair. Each orange point should be matched to the nearest blue point on the same axis; nearby rather than exact overlap reflects ordinary optimization error.
    \item \textbf{Sign-Aware Observation:} A single vector per pair. The "negative" feature is represented by the negative activation of the same latent, rather than by an additional decoder vector.
\end{itemize}

\paragraph{Visualization B: The Bi-Jump Histogram.}
We collect activations for a single trained latent over the dataset and plot the histogram of $a_i(x)$.

\begin{figure}[t]
    \centering
    \includegraphics[width=0.65\linewidth]{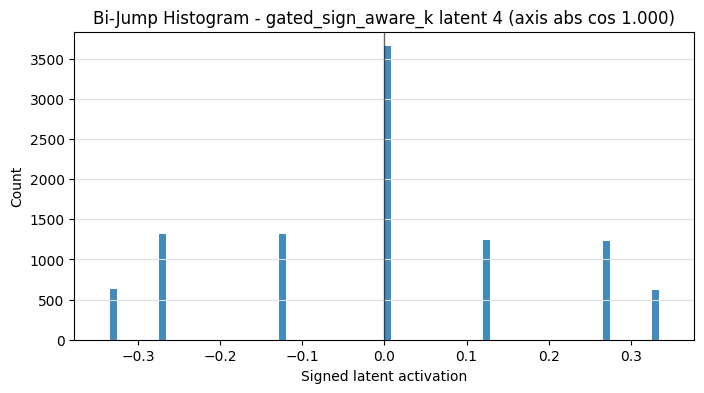}
    \caption{Bi-jump histogram for a representative Sign-Aware latent in Protocol C. The distribution exhibits two symmetric modes, consistent with a single latent representing an antipodal pair via positive vs negative activations.}
    \label{fig:protocol_c_bijump}
\end{figure}

The histogram reveals a tri-modal distribution: a large mass exactly at zero (the dead zone $[-\delta_i^-, \delta_i^+]$), a lobe of positive activations, and a lobe of negative activations. This visually confirms the functionality of the Bi-Jump-ReLU activation function defined in \cref{eq:bi_jump_relu}.

\paragraph{Quantitative Sweep: Superposition Tolerance.}
To obtain a transition-like quantitative curve, we fix the mutually exclusive antipodal construction ($\rho=-1$) and vary
the amount of superposition by changing the number of active pairs per sample (denoted $n_{\mathrm{active}}$).
As $n_{\mathrm{active}}$ increases, multiple pairs contribute to each observation, and the model must represent signed
pair membership in the presence of superposed features.

\paragraph{Quantitative Metric: Signed Recoverability.}
Let $c_j(x)$ denote the signed ground-truth coefficient for pair $j$: $c_j(x)>0$ when the $+u_j$ feature is active, $c_j(x)<0$ when
the $-u_j$ feature is active, and $c_j(x)=0$ otherwise.
For each pair axis $u_j$, we consider latents whose decoder columns satisfy $|\cos(D_{:,i},u_j)|\ge \tau$ and orient
their activations as $\tilde a_i(x)=\mathrm{sign}(D_{:,i}^\top u_j)\,a_i(x)$, where $D_{:,i}^\top u_j$ is the dot product
between the decoder column and the pair axis.
Among these aligned latents, we select a single best candidate per pair using the same sign-consistency checks as in the
consolidation metric below, and compute the Pearson correlation between $\tilde a_{i^*(j)}(x)$ and $c_j(x)$ over test
samples where the pair is active.
We report the mean of this best-candidate correlation across pairs as a continuous measure of signed recoverability.

\begin{figure}[t]
    \centering
    \includegraphics[width=0.75\linewidth]{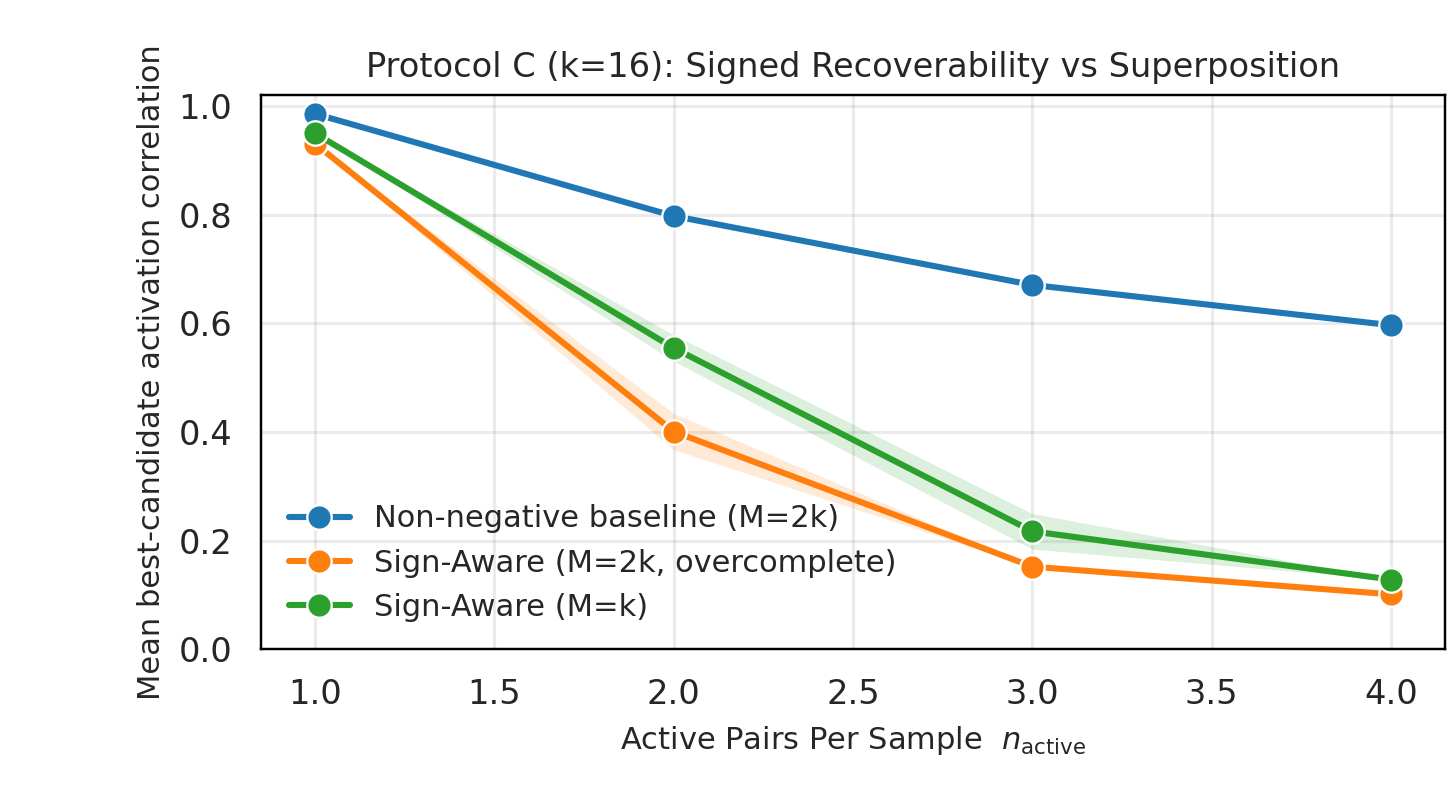}
    \caption{Protocol C quantitative curve (superposition tolerance, $k=16$): signed recoverability decreases as the number of active pairs per sample increases. We plot mean best-candidate activation correlation (mean $\pm$ standard error (SE) over 16 seeds).}
    \label{fig:protocol_c_superposition_recoverability}
\end{figure}

\paragraph{Thresholded Metric: Pair Consolidation Rate.}
We also report the \textit{Pair Consolidation Rate}, a stricter, thresholded variant of the above metric: the fraction
of ground-truth pairs represented by a single latent that is both axis-aligned and sign-consistent.

We define that a ground-truth pair $(j_1,j_2)$ is consolidated by latent $i$ if:
\begin{itemize}
    \item The decoder column $D_{:,i}$ has absolute cosine similarity above a threshold $\tau$ (e.g., $\tau = 0.9$) with the pair axis (equivalently, with both antipodal directions), and
    \item The oriented activation $\tilde a_i(x)$ is positively correlated with the signed ground-truth coefficient for the pair, and
    \item The sign of $\tilde a_i(x)$ agrees with the ground-truth sign in both regimes (when $+u_j$ is active vs when $-u_j$ is active).
\end{itemize}
The Pair Consolidation Rate is the fraction of ground-truth pairs for which there exists at least one latent satisfying these conditions.

\begin{figure}[t]
    \centering
    \includegraphics[width=0.75\linewidth]{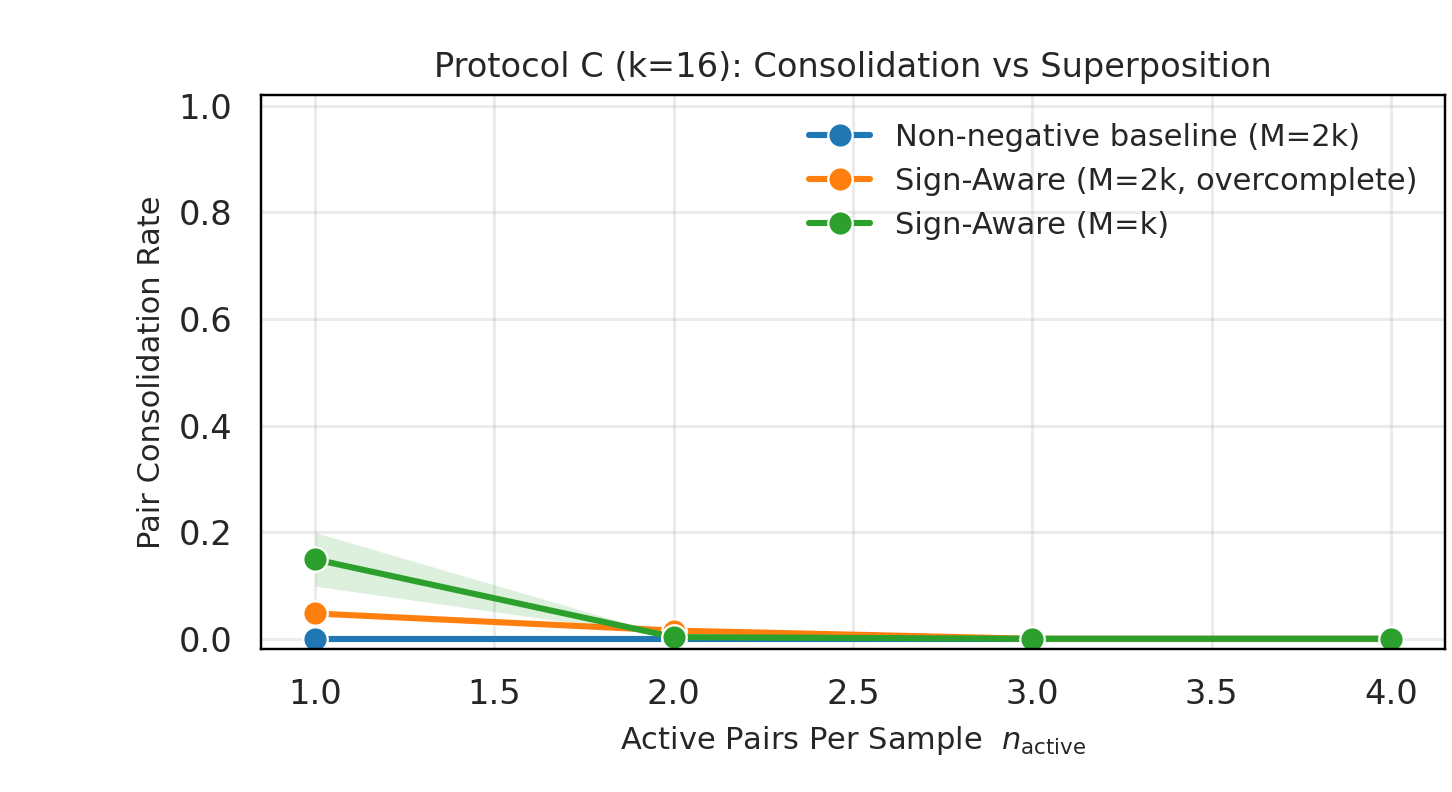}
    \caption{Thresholded Pair Consolidation Rate vs superposition in Protocol C (mean $\pm$ SE over 16 seeds).}
    \label{fig:protocol_c_superposition_consolidation}
\end{figure}

Signed recoverability shows a clear transition-like degradation with superposition (Fig.~\ref{fig:protocol_c_superposition_recoverability}). The overcomplete Sign-Aware setting ($M=2k$) degrades faster, consistent with symmetry between one-latent and two-latent solutions.

The thresholded Pair Consolidation Rate collapses rapidly as superposition increases (Fig.~\ref{fig:protocol_c_superposition_consolidation}), indicating that sign-consistent single-latent representations become unreliable under strong interference. We therefore restrict the single-latent consolidation claim to the low-superposition regime ($n_{\mathrm{active}} = 1$): at $n_{\mathrm{active}} \ge 2$, exact one-latent pair consolidation largely fails (rate $< 0.05$ for the width-$k$ model), while the graded signed-recoverability signal of Fig.~\ref{fig:protocol_c_superposition_recoverability} persists at ${\sim}0.55$ -- i.e., signed information survives, but the clean one-latent-per-pair geometry does not, and we do not claim it generalizes to realistic superposition levels.

\paragraph{Robustness Sweep: Within-Pair Correlation.}
As a robustness check, we also sweep the within-pair correlation $\rho$ from $-1$ (mutually exclusive) to $0$ (independent)
at fixed $n_{\mathrm{active}}=1$ and measure the Pair Consolidation Rate.
This sweep produces a gradual degradation rather than a sharp transition, and is best interpreted as a robustness check
rather than a phase transition.

\begin{figure}[t]
    \centering
    \includegraphics[width=0.75\linewidth]{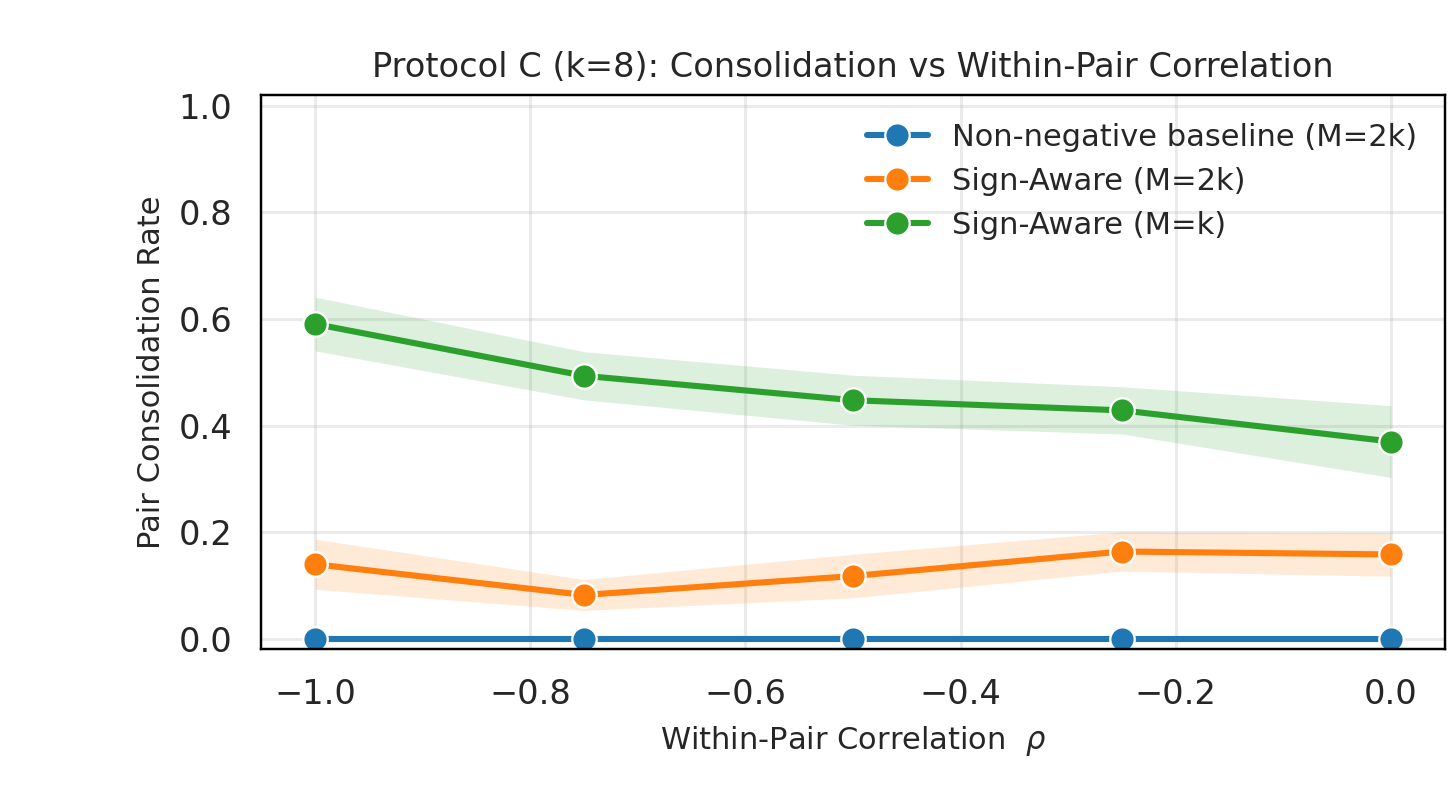}
    \caption{Robustness sweep: Pair Consolidation Rate vs within-pair correlation $\rho$ at $n_{\mathrm{active}}=1$ (mean $\pm$ SE over 16 seeds).}
    \label{fig:protocol_c_corr_sweep}
\end{figure}

\section{Baseline Implementations for the LLM Benchmark}
\label{sec:llm_baseline_defs}

For real-LLM activation experiments, the main-text comparison uses Gated SAE and AbsTopK baselines implemented in
the shared harness. We document the exact implementation details used in these runs:
\begin{itemize}
    \item \textbf{Gated SAE:} we use the shared projection $t_i(x)=D_{:,i}^\top(x-b_{\text{dec}})$, gate pre-activation $\pi_i(x)=\alpha_i t_i(x)+\beta_i$, and activation
    \[
        a_i(x)=\mathbf{1}[\pi_i(x)>0]\,\mathrm{ReLU}(g_i t_i(x)+b_{\text{mag},i}).
    \]
    The sparsity source is $\mathrm{ReLU}(\pi_i(x))$, and the baseline uses the standard frozen-decoder auxiliary reconstruction from \cite{rajamanoharan2024gated}.
    \item \textbf{AbsTopK SAE:} we keep signed pre-activations $u_i(x)$, apply deterministic per-sample top-$k$ selection on $|u_i(x)|$ using detached scores, and set $a_i(x)=m_i(x)u_i(x)$.
\end{itemize}
AbsTopK uses fixed-$k$ sparsity control and therefore does not add an explicit sparsity-coefficient term to the
reconstruction loss in these runs. Like SA-GSAE and Gated SAE, AbsTopK is trained at both widths with the same $k$
grid; we refer to the width-$16{,}384$ runs as \emph{AbsTopK (half)} in the census and calibration tables.

\paragraph{Fairness note on encoder parameterization.}
SA-GSAE and Gated SAE use the shared projection $t_i(x)=D_{:,i}^\top(x-b_{\text{dec}})$, whereas AbsTopK retains
its native encoder parameterization in the shared harness. We intentionally keep each baseline in its standard,
well-performing form rather than forcing a common encoder template, because the practical question is whether sign-aware
gating is competitive with the methods practitioners actually use. The LLM tables should therefore be read as
end-to-end recipe comparisons, not as a pure isolation of encoder tying versus untied encoders. The hybrid
AbsTopK+GatedMag ablation in Appendix~\ref{sec:ablations} partially disentangles this issue by pairing our magnitude path with
a fixed-$k$ signed selector.

\section{LLM Ablations}
\label{sec:ablations}

\paragraph{The auxiliary loss is essential.}
Removing the auxiliary loss ($\lambda_{\text{aux}} = 0$) is catastrophic: on Pythia-1B \texttt{mlp\_out}, the entire
$\lambda$ sweep of the no-aux variant clamps at $L_0 \in [1.7, 2.2]$ -- the model fires effectively two latents per
token regardless of $\lambda$ -- dead-fraction rises to $98\%$, and LR drops to $0.27$. Without auxiliary supervision,
gate parameters receive gradients only through the main reconstruction and sparsity terms, which is insufficient to
learn effective two-sided detection. This confirms that the auxiliary frozen-decoder path is not merely a training
convenience but a structural requirement for sign-aware gating.

\paragraph{Symmetric magnitude is a viable simplification.}
Tying the per-polarity magnitude scales ($r_i^+ = r_i^-$) yields $R^2 = 0.742 \pm 0.000$ and dead-fraction $0.9\% \pm 0.0\%$
at matched $L_0 = 64$, versus $R^2 = 0.741$ and dead-fraction $0.8\%$ for the full SA-GSAE model -- effect sizes of
$|\Delta R^2| = 0.0015$ and $|\Delta\text{MSE}| = 0.0002$, both practically negligible; the LR delta is within one
standard error, and the dead-fraction delta ($\approx 0.06$ pp) is $\approx 2$ paired SE from zero but absolutely
negligible. This weakens the Protocol-A suggestion that independent per-polarity scaling
contributes on real LLM activations: on Pythia-1B \texttt{mlp\_out}, it does not. The elements SA-GSAE cannot train
without at this operating point are the two-sided gating mechanism and the auxiliary reconstruction path; independent per-polarity
magnitude asymmetry is a small second-order refinement at best.

\paragraph{Dead-zone initialization exhibits a U-shape.}
We sweep the initial dead-zone half-width $\delta_0 \in \{10^{-3}, 10^{-2}, 0.5, 1.0\}$, keeping all other hyperparameters
fixed. Small initializations degrade reconstruction and inflate dead fraction: $\delta_0 = 10^{-3}$ attains $R^2 = 0.715$
and dead-fraction $27\%$ at matched $L_0 = 64$, versus the reference $R^2 = 0.741$ and $0.8\%$ dead. Large initializations
are neutral to mildly harmful: $\delta_0 = 1.0$ attains $R^2 = 0.744$ (marginally higher than the reference) with a
slightly elevated dead-fraction of $7.8\%$. We recommend $\delta_0 \in [0.1, 1.0]$ as a default; the common practice of
initializing thresholds near zero actively harms capacity utilization.

\paragraph{A reset-offsets variant is indistinguishable on stable cells -- and protective on the unstable one.}
Enabling threshold resets on dead-latent resampling (\texttt{reset\_offsets\_true}) yields $R^2 = 0.741 \pm 0.000$ and
dead-fraction $0.8\% \pm 0.0\%$ at matched $L_0 = 64$ on Pythia-1B \texttt{mlp\_out}, indistinguishable from the
reference. On the one unstable cell (\texttt{resid-mid/SmolLM3-3B}), however, this setting separates every stable from
every collapsing run in our training-dynamics audit (\cref{app:collapse_diagnosis}); we therefore recommend enabling
it by default.

\paragraph{Hybrid AbsTopK + gated magnitude is dominated by SA-GSAE.}
To disentangle the contribution of sign-aware gating from the gated magnitude path, we evaluate a hybrid baseline that
uses AbsTopK-style absolute selection (deterministic top-$k$ on $|t_i(x)|$) but retains the gated magnitude computation
from SA-GSAE. At matched $L_0 = 64$, this hybrid attains $R^2 = 0.724 \pm 0.000$ and dead-fraction $41\% \pm 0.4\%$,
both strictly worse than full SA-GSAE ($R^2 = 0.741$, dead-fraction $0.8\%$) and than AbsTopK alone on dead-fraction
($0.0\%$). Fixed-$k$ absolute selection therefore does not combine favourably with the gated magnitude path -- it
neither matches AbsTopK's dead-fraction floor nor matches SA-GSAE's reconstruction quality -- which clarifies that
learned two-sided thresholds with auxiliary supervision drive both capacity utilization and reconstruction quality
jointly, rather than being separable into a selection and a magnitude component.

\begin{table}[!t]
    \centering
    \caption{Ablation study on Pythia-1B \texttt{mlp\_out} at matched $L_0 = 64$ (mean $\pm$ SE over 3 seeds). All
    variants use width $H = 32{,}768$. $\delta_0$ denotes the initial dead-zone half-width. Entries whose SE displays
    as $(0.000)$ or $(0.0000)$ have SE below $5\!\cdot\!10^{-4}$ (LR, $R^2$, Dead) or $5\!\cdot\!10^{-5}$ (MSE),
    respectively; e.g., the reference $R^2$ SE is $5\!\cdot\!10^{-5}$. ``no aux'' did not reach
    $L_0 = 64$: its sweep clamps at $L_0 \in [1.7, 2.2]$; we report its metrics at the maximum attained $L_0$ rather
    than at $L_0 = 64$ and flag the row accordingly.}
    \label{tab:ablations}
    \small
    \setlength{\tabcolsep}{4pt}
    \begin{tabular}{lcccc}
        \toprule
        Variant & LR $\uparrow$ & $R^2$ $\uparrow$ & MSE $\downarrow$ & Dead $\downarrow$ \\
        \midrule
        SA-GSAE, full (reference)        & 0.785 (0.002) & 0.741 (0.000) & 0.0375 (0.0000) & 0.008 (0.001) \\
        SA-GSAE, sym.\ mag.\ ($r^+ = r^-$) & 0.784 (0.002) & \textbf{0.742} (0.000) & \textbf{0.0373} (0.0000) & 0.009 (0.000) \\
        SA-GSAE, reset offsets           & 0.781 (0.000) & 0.741 (0.000) & 0.0375 (0.0000) & 0.008 (0.000) \\
        SA-GSAE, $\delta_0 = 1.0$         & 0.788 (0.001) & 0.744 (0.000) & \textbf{0.0371} (0.0000) & 0.078 (0.000) \\
        SA-GSAE, $\delta_0 = 0.5$         & \textbf{0.789} (0.002) & 0.743 (0.000) & \textbf{0.0371} (0.0000) & 0.044 (0.001) \\
        SA-GSAE, $\delta_0 = 10^{-2}$     & 0.777 (0.003) & 0.733 (0.001) & 0.0387 (0.0001) & 0.018 (0.002) \\
        SA-GSAE, $\delta_0 = 10^{-3}$     & 0.757 (0.003) & 0.715 (0.001) & 0.0413 (0.0001) & 0.269 (0.003) \\
        Hybrid AbsTopK + gated mag       & 0.773 (0.001) & 0.724 (0.000) & 0.0400 (0.0000) & 0.414 (0.004) \\
        SA-GSAE, no aux${}^{\dagger}$    & 0.278${}^{\dagger}$ & 0.353${}^{\dagger}$ & 0.094${}^{\dagger}$ & 0.978${}^{\dagger}$ \\
        \midrule
        Gated SAE, full (baseline)       & 0.761 (0.003) & 0.711 (0.001) & 0.0419 (0.0001) & 0.737 (0.004) \\
        AbsTopK, full (baseline)         & 0.746 (0.002) & 0.666 (0.000) & 0.0484 (0.0000) & \textbf{0.000} (0.000) \\
        \bottomrule
        \multicolumn{5}{l}{\footnotesize ${}^{\dagger}$ reported at the max attained $L_0 \approx 2$, not at $L_0 = 64$; the sweep never reaches $L_0 = 64$.}
    \end{tabular}
\end{table}

\FloatBarrier

\section{Full-Width SA-GSAE Results}
\label{app:full_width_results}

For completeness, at matched $L_0 \approx 64$ on all six hookpoint $\times$ backbone cells, full-width SA-GSAE
($H = 32{,}768$) is well-behaved on five of six cells: it attains the best $R^2$ on
\texttt{mlp\_out-mid/Pythia-1B}, \texttt{mlp\_out-mid/SmolLM3-3B}, and \texttt{resid-mid/Pythia-1B}, and is within
$0.016$ of the best $R^2$ on the remaining two non-collapsing cells.

The sixth cell, \texttt{resid-mid/SmolLM3-3B}, exhibits a \emph{reproducible reconstruction collapse}: at the matched
operating point of $L_0 = 64$, full-width SA-GSAE attains MSE $= 1.15 \pm 0.22$ and $R^2 = -0.006 \pm 0.189$, whereas
every other variant on the same cell remains at MSE $\leq 0.01$ and $R^2 \geq 0.99$. \cref{tab:full_width_resid_smollm_per_seed}
shows the per-seed $\lambda$ sweep for this cell. All three seeds start well-behaved at $\lambda \leq 2.4 \cdot 10^{-4}$
($L_0 \approx 270$ to $760$, MSE $\approx 0.003$, $R^2 \geq 0.995$), then enter a regime where reconstruction fidelity
degrades sharply: MSE inflates to $0.2$--$6.0$ at interior $\lambda$ values, $R^2$ drops to as low as $-4.3$, and
$3$ of $24$ sweep points fall below LR $= 0.5$, one on each of seeds $0$, $1$, and $2$, at
$\lambda \in \{2.44 \cdot 10^{-3}, 3.89 \cdot 10^{-3}\}$ (the two largest sparsity coefficients). The
worst-case sweep point is seed $2$ at $\lambda = 3.9 \cdot 10^{-3}$: $\text{loss\_reconstructed} = 10.25$ is only
$1.51$ nats below $\text{loss\_ablated} = 11.76$, so the reconstruction recovers only $\text{LR} = 0.175$ of the
ablation gap. The reconstruction never strictly worsens relative to ablation, but it comes within
$\sim 1.5$ nats of doing so.

The half-width SA-GSAE on the same cell is monotone on $3/3$ seeds, $R^2 \geq 0.985$ at every sparsity point, and
MSE never exceeds $0.018$. This contrast should be read together with the training-dynamics diagnosis of
\cref{app:collapse_diagnosis}: the reported full- and half-width configurations differ not only in width but also in
threshold initialization and in the dead-latent threshold-reset setting (\cref{tab:sa_config}). The audit of all
preserved runs on this cell locates the trigger at the threshold-unfreeze step, rules out the warmup schedule and the
auxiliary coefficient, shows that half-width pilots without threshold resets destabilized as well, and identifies
small threshold initialization with dead-latent threshold resets -- the reported half-width configuration -- as the
robust setting. The halved-width operating point therefore remains motivated primarily by the capacity argument
(sign-aware latents carry double the information, so $H_\pm = H$ suffices), with the stability contrast on this cell
attributable to the accompanying configuration rather than width alone.

\begin{table}[!h]
    \centering
    \caption{Per-seed $\lambda$ sweep of full-width SA-GSAE on \texttt{resid-mid/SmolLM3-3B}. MSE values inside the
    collapse regime ($\text{MSE} > 0.1$) are bolded; the bolded cells are the same rows in which $R^2$ drops below
    $0.85$ and the LM-loss gap closes substantially. The baseline losses are $\text{loss\_original} = 3.153$ and
    $\text{loss\_ablated} = 11.762$ (identical across variants); $\text{loss\_reconstructed}$ never strictly exceeds
    $\text{loss\_ablated}$ but comes within $1.51$ nats of it at the worst-case sweep point (seed $2$,
    $\lambda = 3.9 \cdot 10^{-3}$).}
    \label{tab:full_width_resid_smollm_per_seed}
    \scriptsize
    \setlength{\tabcolsep}{3pt}
    \begin{tabular}{cccccc}
        \toprule
        Seed & $\lambda$ & $L_0$ & MSE & $R^2$ & $\text{loss\_reconstructed}$ \\
        \midrule
        \multirow{8}{*}{0}
          & $1.48 \cdot 10^{-4}$ & $761.06$ & $0.0032$ & $0.997$ & $3.143$ \\
          & $2.36 \cdot 10^{-4}$ & $273.87$ & $0.0052$ & $0.995$ & $3.188$ \\
          & $3.77 \cdot 10^{-4}$ & $113.02$ & $\mathbf{0.234}$ & $0.795$ & $3.585$ \\
          & $6.01 \cdot 10^{-4}$ & $72.31$ & $\mathbf{1.738}$ & $-0.520$ & $4.376$ \\
          & $9.59 \cdot 10^{-4}$ & $48.32$ & $0.072$ & $0.937$ & $3.716$ \\
          & $1.53 \cdot 10^{-3}$ & $35.77$ & $\mathbf{2.509}$ & $-1.20$ & $6.634$ \\
          & $2.44 \cdot 10^{-3}$ & $26.64$ & $\mathbf{1.641}$ & $-0.436$ & $5.894$ \\
          & $3.89 \cdot 10^{-3}$ & $31.56$ & $\mathbf{0.733}$ & $0.359$ & $7.637$ \\
        \midrule
        \multirow{8}{*}{1}
          & $1.48 \cdot 10^{-4}$ & $751.39$ & $0.0032$ & $0.997$ & $3.170$ \\
          & $2.36 \cdot 10^{-4}$ & $271.91$ & $0.0057$ & $0.995$ & $3.162$ \\
          & $3.77 \cdot 10^{-4}$ & $108.86$ & $\mathbf{1.236}$ & $-0.081$ & $4.034$ \\
          & $6.01 \cdot 10^{-4}$ & $75.10$ & $\mathbf{0.191}$ & $0.833$ & $3.875$ \\
          & $9.59 \cdot 10^{-4}$ & $48.92$ & $\mathbf{1.669}$ & $-0.460$ & $4.730$ \\
          & $1.53 \cdot 10^{-3}$ & $40.51$ & $\mathbf{0.625}$ & $0.454$ & $4.659$ \\
          & $2.44 \cdot 10^{-3}$ & $28.37$ & $\mathbf{6.015}$ & $-4.26$ & $8.182$ \\
          & $3.89 \cdot 10^{-3}$ & $27.24$ & $\mathbf{0.212}$ & $0.814$ & $6.964$ \\
        \midrule
        \multirow{8}{*}{2}
          & $1.48 \cdot 10^{-4}$ & $755.40$ & $0.0032$ & $0.997$ & $3.129$ \\
          & $2.36 \cdot 10^{-4}$ & $276.24$ & $0.0052$ & $0.995$ & $3.186$ \\
          & $3.77 \cdot 10^{-4}$ & $112.51$ & $\mathbf{0.183}$ & $0.840$ & $3.660$ \\
          & $6.01 \cdot 10^{-4}$ & $67.95$ & $\mathbf{1.359}$ & $-0.189$ & $5.566$ \\
          & $9.59 \cdot 10^{-4}$ & $49.66$ & $\mathbf{1.967}$ & $-0.721$ & $6.527$ \\
          & $1.53 \cdot 10^{-3}$ & $33.43$ & $\mathbf{0.220}$ & $0.807$ & $4.252$ \\
          & $2.44 \cdot 10^{-3}$ & $26.95$ & $\mathbf{0.632}$ & $0.447$ & $6.692$ \\
          & $3.89 \cdot 10^{-3}$ & $29.81$ & $\mathbf{3.660}$ & $-2.20$ & $10.251$ \\
        \bottomrule
    \end{tabular}
\end{table}

\FloatBarrier
\section{Collapse Diagnosis on \texttt{resid-mid/SmolLM3-3B}}
\label{app:collapse_diagnosis}

This appendix diagnoses the full-width reconstruction collapse of \cref{app:full_width_results} from the preserved
per-step training histories of \emph{every} run on this cell -- reported runs and superseded pilots alike -- plus 12
prospective full-length (450k-step) confirmation runs. We first disclose the per-cell SA-GSAE training configuration
(\cref{tab:sa_config}), which the submitted paper omitted: threshold initializations $\delta_0$ were tuned per cell,
and the reported half-width run on this cell -- uniquely among all reported runs -- enables
\texttt{reset\_activation\_offsets\_on\_resample} (re-initializing the thresholds of resampled dead latents).

\begin{table}[!h]
    \centering
    \caption{Per-cell SA-GSAE training configuration of the reported runs. $\delta_0$: initial dead-zone half-width
    ($10^{-3}$ denotes the repository default); reset: \texttt{reset\_activation\_offsets\_on\_resample}. All reported
    runs use 3 seeds, the standard threshold warmup ($40$k steps on Pythia-1B, $80$k on SmolLM3-3B), and the training
    lengths of the benchmark setup ($225$k / $450$k steps). Superseded pilots with other settings are preserved in the
    released artifact and excluded from all published numbers.}
    \label{tab:sa_config}
    \small
    \setlength{\tabcolsep}{5pt}
    \begin{tabular}{llcc}
        \toprule
        Backbone & Hookpoint & $\delta_0$ (full / half) & reset (full / half) \\
        \midrule
        \multirow{3}{*}{Pythia-1B}
        & \texttt{mlp\_out} & $0.1$ / $0.1$   & off / off \\
        & \texttt{attn}     & $0.5$ / $0.5$   & off / off \\
        & \texttt{resid}    & $1.0$ / $1.0$   & off / off \\
        \midrule
        \multirow{3}{*}{SmolLM3-3B}
        & \texttt{mlp\_out} & $10^{-3}$ / $10^{-3}$ & off / off \\
        & \texttt{attn}     & $0.15$ / $0.15$ & off / off \\
        & \texttt{resid}    & $0.5$ / $10^{-3}$ & off / \textbf{on} \\
        \bottomrule
    \end{tabular}
\end{table}

\paragraph{Method.}
For each subrun we classify the recorded history with a fixed rule: \emph{collapsed} if final $R^2 < 0.5$ or the
reconstruction loss exceeds $10\times$ its settle-window baseline on more than $10\%$ of post-settle steps;
\emph{degraded} if it exceeds $3\times$ on more than $25\%$; \emph{stable} otherwise. The baseline is the median loss
over the settle window (ending at $\max(\text{warmup},\, \text{steps}/10)$), and onsets are the first sustained
$3\times$ crossing after that window, reported relative to the threshold-unfreeze step.

\begin{table}[!h]
    \centering
    \caption{Collapse incidence by configuration on \texttt{resid-mid/SmolLM3-3B} (all preserved runs plus prospective
    confirmations; ``recorded'' is the preserved history length -- some superseded pilots were truncated when replaced,
    and their classifications cover only the recorded window). Onset is the median first sustained $3\times$ excursion
    relative to threshold unfreeze. The prospective block (bottom) contains new full-length 450k-step runs
    ($\delta_0 = 0.5$, reset off, seed 0, three sparsity levels each). With no warmup, thresholds train from step 1 and
    the classifier's detection window starts at step 45k; with 40k warmup the detection window also starts at 45k, so
    those onsets are upper-bounded observations.}
    \label{tab:collapse_incidence}
    \scriptsize
    \setlength{\tabcolsep}{3.5pt}
    \begin{tabular}{lccccccc}
        \toprule
        Run & Width & $\delta_0$ & reset & Warmup & Recorded & Collapsed/Degraded/Stable & Onset$-$unfreeze \\
        \midrule
        full v1 (pilot)            & 32{,}768 & 0.3       & off & 80k & 450k       & 21/0/3   & $+28$ \\
        full v2 (pilot)            & 32{,}768 & 0.1       & off & 80k & 450k       & 6/0/10 (of 16)  & $+4$ \\
        full v3 (pilot)            & 32{,}768 & 1.0       & off & 80k & 225k       & 8/0/0    & $+145$k (window end) \\
        full (reported)            & 32{,}768 & 0.5       & off & 80k & 450k       & 17/0/7   & $+8$ \\
        full (pilot, resets)       & 32{,}768 & $10^{-3}$ & on  & 40k & 225k       & 0/2/6    & degraded only, $+185$k \\
        full (pilot, resets)       & 32{,}768 & 0.5       & on  & 80k & 200k       & 0/2/6    & degraded only, $+120$k \\
        \midrule
        half v1 (pilot)            & 16{,}384 & $10^{-3}$ & off & 80k & 450k/250k  & 8/4/4 (of 16) & $+40$ to $+130$k (spread) \\
        half v2 (pilot)            & 16{,}384 & 1.0       & off & 80k & 225k       & 8/0/0    & $+145$k (window end) \\
        half v3 (pilot, resets)    & 16{,}384 & $10^{-3}$ & on  & 40k & 225k       & 0/0/8    & none \\
        half v4 (pilot, resets)    & 16{,}384 & $10^{-3}$ & on  & 40k & 450k       & 0/0/24   & none (min $R^2$ $0.981$) \\
        half (reported, resets)    & 16{,}384 & $10^{-3}$ & on  & 80k & 450k       & 0/0/24   & none (min $R^2$ $0.985$) \\
        Gated SAE (reported)       & 32{,}768 & --        & --  & 2k  & 450k       & 0/0/24   & none (min $R^2$ $0.971$) \\
        \midrule
        prosp.\ wu0k, aux 1.0      & 32{,}768 & 0.5 & off & 0   & 450k & 3/3 collapsed & unstable from detection window \\
        prosp.\ wu40k, aux 1.0     & 32{,}768 & 0.5 & off & 40k & 450k & 3/3 collapsed & $+5.1$k--$5.5$k (detection floor) \\
        prosp.\ wu80k, aux 1.0     & 32{,}768 & 0.5 & off & 80k & 450k & 3/3 collapsed & $+0/+0/+4$ (min $R^2$ $-12.0$) \\
        prosp.\ wu80k, aux 0.3     & 32{,}768 & 0.5 & off & 80k & 450k & 3/3 collapsed & $+0/+4/+4$ (min $R^2$ $0.44$) \\
        \bottomrule
    \end{tabular}
\end{table}

\paragraph{Findings.}
(i)~\emph{Trigger: threshold learning.} Wherever the classifier can resolve it, the sustained excursion begins at the
threshold-unfreeze step (reported full-width run: median $+8$ steps; prospective 80k-warmup runs: $+0$ to $+4$ steps).
(ii)~\emph{Not the warmup schedule, not the auxiliary coefficient.} Prospective runs destabilize for warmups of $0$,
$40$k, and $80$k steps alike, and for auxiliary coefficients $0.3$ and $1.0$ (12/12 collapsed).
(iii)~\emph{Not width alone.} Half-width pilots without threshold resets destabilized as well (8/16 at
$\delta_0 = 10^{-3}$; 8/8 at $\delta_0 = 1.0$ in the recorded window); among no-reset runs, incidence grows with
$\delta_0$.
(iv)~\emph{Configuration that separates stable from unstable.} Every run with dead-latent threshold resets enabled --
both widths, $\delta_0 \in \{10^{-3}, 0.5\}$ -- is collapse-free in its recorded window, including 48/48 full-length
half-width subruns; every no-reset run destabilizes in part or in full. (v)~\emph{Site-specificity.} The other five
cells are stable under the same architecture with no-reset settings.

\paragraph{Interpretation and recommendation.}
The collapse is a threshold-learning-triggered optimization instability specific to this highest-dynamic-range
hookpoint, entering through the dead-latent path and suppressed by periodically re-initializing the thresholds of
resampled dead latents. Because the reported full- and half-width runs differ in this reset setting
(\cref{tab:sa_config}), the stability contrast between them in \cref{app:full_width_results} should not be attributed
to width alone. We recommend small threshold initialization with dead-latent threshold resets (with the tied-threshold
default of \cref{app:theory}) as the robust configuration; the half-width operating point remains recommended on the
independent capacity grounds of \S\ref{sec:half_width}. A fully crossed width $\times$ $\delta_0$ $\times$ reset
factorial at three seeds is deferred to future work.

\FloatBarrier
\section{Additional Validation Studies}
\label{app:rebuttal_studies}

We ran four additional studies that directly test the semantic and causal content
of two-sided latents and the simplification of the unit. All artifacts (packets, raw judgments, per-latent records,
run configs) are released with the code.

\paragraph{Blinded semantic-coherence audit.}
We sampled 25 both-sign SA-GSAE latents per cell (150 total; half-width checkpoints at matched $L_0 \approx 64$;
strata: polarity-imbalance $\times$ firing-rate $\times$ calibration-quality tertiles; sampling seed fixed before
viewing any contexts). Each latent was rendered as two label-free context sets (12 contexts per side: 8 top + 4
random-active; $\pm 31$-token windows; activating token highlighted; magnitudes stripped) and judged by two
independent frontier-LLM judges (independent order randomization and per-judge sign flips) in two phases: describe
each side, then classify the relation (coherent opposition / same-domain non-opposition / unrelated packing /
indeterminate). Controls: 30 random cross-latent pairings, 30 AbsTopK signed latents, 30 near-antipodal Gated-SAE
latent \emph{pairs} (max-weight cosine matching), 12 same-side split-half ceiling items, and 10 calibration items
with known expected labels. Validity: 4-way inter-judge $\kappa = 0.50$; both judges labeled all four clean antipodal
calibration items correctly; the same-side ceiling was judged related (not packing) in $75\%$ of items. Results:
judged coherent opposition for SA-GSAE is $1.5\%$ (95\% CI $[0, 4.7]$; 2 of 135 consensus items), not above random
pairings ($0\%$); unrelated packing dominates at $64\%$ (CI $[50, 79]$) and is statistically indistinguishable from
random pairings ($65\%$), AbsTopK signed latents ($59\%$), and Gated antipodal pairs ($63\%$); only $36\%$ of
both-sign latents had both sides even partially interpretable to both judges. Pre-registered hypotheses that SA-GSAE
exceeds random controls are not supported (permutation $p = 0.70/0.55$). The rare exceptions are genuine axes (e.g.,
\texttt{resid-mid/Pythia-1B} latent 14393: concrete/countable vs.\ abstract plural nouns, described independently by
both judges). Conclusion: at 1B--3B scale and $L_0 \approx 64$, two-sided usage does not generally correspond to
judge-nameable semantic opposition -- \emph{for any tested method, including the two-latent decompositions of
non-negative SAEs} -- and we scope the paper's claims to capacity, dead-latent efficiency, and the causal
characterization below. In hindsight this is what the training incentive predicts: contrasts such as ``pressure too
high'' vs.\ ``pressure too low'' are a conceptual motivation, but the objective gives the model no incentive to pack
strictly opposing concepts into one latent based on human understandability; the two sides of a latent are paired
statistically, by whichever features are most anticorrelated in the data under the reconstruction--sparsity
trade-off (cf.\ the packing-cost analysis in \cref{app:theory}), and such pairs only occasionally coincide with a
nameable semantic axis.

\paragraph{Sign-conditioned causal interventions.}
For the same population (150 sampled / 146 evaluated SA-GSAE latents; 30 AbsTopK; 30 Gated pairs) we applied
feature-isolated edits $x' = x + (a^* - a_i(x))\, D_{:,i}$ at the hookpoint, with 7 sign-conditional dose quantiles
($q_{5}^-$ to $q_{95}^+$) on 12 held-out active contexts per latent, and measured the Sign-Consistency Score
$\mathrm{SCS}_i = -\mathrm{corr}$ over the vocabulary of the next-token log-prob shifts at the extreme positive vs.\
negative doses. SA-GSAE: mean SCS $0.97/0.92/0.96$ on Pythia \texttt{mlp\_out}/\texttt{attn}/\texttt{resid} and
$0.80/0.52/0.80$ on SmolLM3-3B (pooled $0.83$), with monotone dose response (mean Kendall $\tau$ $0.75$--$0.99$).
Norm-matched random-direction $\pm$ edits give SCS $-0.35$ to $-0.49$ (both directions merely degrade, in a
correlated way); driving the two latents of a near-antipodal Gated pair in opposition gives pair-SCS $-0.38$ to
$-0.71$ -- the pair does not implement one axis, while a single signed latent does. AbsTopK signed latents show
comparable SCS ($0.83$--$0.99$), consistent with signed support being the operative ingredient. Off-target LM-loss
deltas at the dosed conditions are small (median $|\Delta| \approx 3\!\cdot\!10^{-4}$, p90 $\approx 1.2\!\cdot\!10^{-3}$).

\paragraph{Boundary and jump diagnostics.}
The activation is monotone piecewise-linear in the shared projection $t_i(x)$: along any continuous
interpolation of the input, a latent crosses each branch boundary at most once and passes through the dead zone
when moving between polarities; at branch entry a discontinuous jump of up to
$J_i^+ = \max(0,\, g_i^+ t_i^+ + b_{\text{mag},i})$ is possible (with $t_i^+$ the effective positive support
threshold of \cref{app:theory}, and symmetrically on the negative side). The following measurements quantify how
often trained models operate near these boundaries.
On held-out data (half-width models, matched $L_0$, all six cells): the mean fraction of tokens within
$0.05\,\mathrm{sd}(\pi_i)$ of a gate boundary is $2.3$--$6.2 \times 10^{-4}$ per latent (median latent
$\le 6 \times 10^{-5}$ at the $0.01$ threshold); median dead-zone width is $6.7$--$11.7\,\mathrm{sd}(\pi_i)$;
median branch-entry jumps are $0.19$--$0.66\times$ the latent's typical nonzero magnitude ($0$--$4\%$ of latents
enter continuously); and the gate-open-but-magnitude-zero region covers $\le 0.1\%$ of gate-open events. Latents
live deep inside one regime or the dead zone; the two-stage design does not create a wasted regime.

\paragraph{Simplification ablations (tied units and auxiliary sweep).}
At half width on \texttt{mlp\_out-mid/Pythia-1B} with the exact paper protocol (225k steps, 8-$\lambda$ sweep,
matched-$L_0$ anchor at 64): fully tied unit (single threshold, tied gains) $R^2 = 0.7193 \pm 0.0005$ vs.\ asymmetric
reference re-run $0.7169 \pm 0.0004$ (3 seeds each; the artifact's stored paper run gives $0.7170 \pm 0.0003$,
validating pipeline equivalence); single-seed tied-thresholds $0.7165$ and tied-gains $0.7193$; dead fractions all
$0.001$--$0.002$. The fully tied symmetric unit matches (marginally exceeds) the asymmetric one and is adopted as the
default. Auxiliary coefficient: $\lambda_{\text{aux}} = 0$ collapses (attainable $L_0 \le 1.3$, $R^2 = 0.31$,
$98.7\%$ dead), while $\lambda_{\text{aux}} \in \{0.3, 1.0, 3.0\}$ give the same $R^2$-vs-$L_0$ frontier within
$0.001$ $R^2$ at equal $L_0$ over the overlapping ranges; the coefficient shifts the $\lambda \to L_0$ mapping
(stronger aux admits denser operating points), so comparisons are made at matched $L_0$. The auxiliary path is a
necessary mechanism with flat sensitivity over a $10\times$ coefficient range.

\section{Full-Distribution Calibration and Bipolar Census}
\label{app:gamma_bipolar_full}

This appendix supplements the \S Results block on calibration symmetry and bipolar-latent
usage (\cref{tab:gamma_symmetry,tab:bipolar_census_full}) with per-cell CDFs of $\gamma_\pm$
and an extended bipolar census over two polarity-imbalance thresholds.

\begin{figure}[H]
    \centering
    \IfFileExists{llm_benchmarking_plots/gamma_cdfs_grid.png}{
        \includegraphics[width=0.98\textwidth]{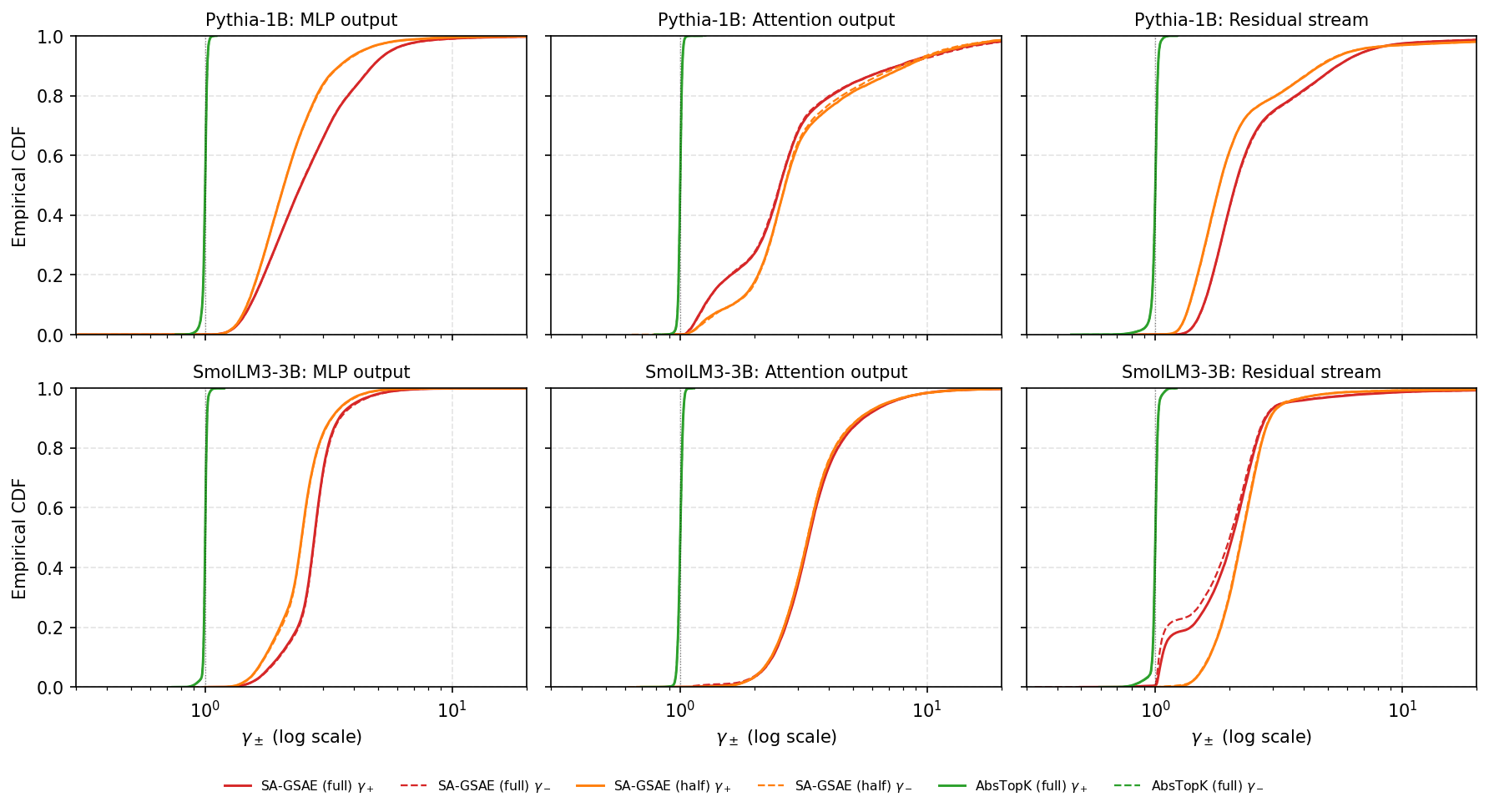}
    }{
        \fbox{\parbox[c][0.25\textwidth][c]{0.80\textwidth}{\centering Missing: gamma CDFs grid}}
    }
    \caption{Empirical CDFs of per-latent calibration slopes $\gamma_+$ (solid) and $\gamma_-$ (dashed), pooled across
    all valid latents and 3 seeds at matched $L_0 \approx 64$. Top row: Pythia-1B at MLP output / attention output /
    residual stream. Bottom row: SmolLM3-3B at the same three hookpoints. Curves are shown for SA-GSAE (full), SA-GSAE
    (half), and AbsTopK (full). The dotted vertical line at $\gamma = 1$ marks the ideal. Two structural observations
    hold on every cell: (i) solid and dashed curves for each variant are essentially superimposed, i.e.\ the $\gamma_+$
    and $\gamma_-$ distributions are symmetric; (ii) SA-GSAE's distribution is concentrated near $\gamma \in [1.5, 5]$
    while AbsTopK's collapses onto $\gamma \approx 1$, which follows mechanically from AbsTopK's lack of a learnable dead
    zone and is not a calibration-training advantage (\cref{app:theory}).}
    \label{fig:gamma_cdfs}
\end{figure}

\begin{table}[!h]
    \centering
    \caption{Full bipolar-latent census at matched $L_0 = 64$ (mean $\pm$ SE over 3 seeds) over both polarity-imbalance
    thresholds. ``Bipolar $<\theta$'' is the fraction of alive latents with $I_i < \theta$. $\hat{p}$ is the
    both-sign calibration coverage (\S\ref{sec:parameter_efficiency}); it measures two-sided capacity usage, not
    semantic bipolarity. Entries whose SE displays as $(0.000)$ have SE below $5\!\cdot\!10^{-4}$. Ranking between SA-GSAE and
    AbsTopK is stable across $\theta \in \{0.3, 0.5\}$.}
    \label{tab:bipolar_census_full}
    \scriptsize
    \setlength{\tabcolsep}{3pt}
    \begin{tabular}{llccccc}
        \toprule
        Cell & Variant & $L_0$ & Alive frac & Bipolar $<0.3$ & Bipolar $<0.5$ & $\hat{p}$ \\
        \midrule
        \multirow{4}{*}{Pythia-1B / \texttt{mlp\_out}}
          & SA-GSAE (full) & 64.6 & 0.992 (0.001) & 0.053 (0.001) & 0.097 (0.001) & 0.300 (0.001) \\
          & SA-GSAE (half) & 66.4 & 0.998 (0.000) & 0.062 (0.001) & 0.090 (0.001) & 0.283 (0.002) \\
          & AbsTopK (full) & 64.0 & 1.000 (0.000) & 0.940 (0.001) & 0.994 (0.000) & 1.000 (0.000) \\
          & AbsTopK (half) & 64.0 & 1.000 (0.000) & 0.945 (0.001) & 0.992 (0.000) & 1.000 (0.000) \\
        \midrule
        \multirow{4}{*}{SmolLM3-3B / \texttt{mlp\_out}}
          & SA-GSAE (full) & 66.0 & 0.994 (0.000) & 0.287 (0.003) & 0.429 (0.003) & 0.707 (0.005) \\
          & SA-GSAE (half) & 65.9 & 0.999 (0.000) & 0.279 (0.005) & 0.387 (0.004) & 0.767 (0.004) \\
          & AbsTopK (full) & 64.0 & 1.000 (0.000) & 0.932 (0.002) & 0.997 (0.000) & 1.000 (0.000) \\
          & AbsTopK (half) & 64.0 & 1.000 (0.000) & 0.960 (0.000) & 0.996 (0.000) & 1.000 (0.000) \\
        \midrule
        \multirow{4}{*}{Pythia-1B / \texttt{attn}}
          & SA-GSAE (full) & 64.6 & 0.226 (0.003) & 0.065 (0.005) & 0.100 (0.008) & 0.077 (0.003) \\
          & SA-GSAE (half) & 64.6 & 0.429 (0.002) & 0.051 (0.003) & 0.080 (0.003) & 0.141 (0.001) \\
          & AbsTopK (full) & 64.0 & 0.498 (0.001) & 0.649 (0.002) & 0.862 (0.002) & 0.603 (0.003) \\
          & AbsTopK (half) & 64.0 & 0.843 (0.005) & 0.709 (0.003) & 0.911 (0.002) & 0.902 (0.006) \\
        \midrule
        \multirow{4}{*}{SmolLM3-3B / \texttt{attn}}
          & SA-GSAE (full) & 65.4 & 0.370 (0.019) & 0.115 (0.007) & 0.188 (0.011) & 0.162 (0.000) \\
          & SA-GSAE (half) & 65.4 & 0.679 (0.021) & 0.124 (0.003) & 0.203 (0.004) & 0.322 (0.005) \\
          & AbsTopK (full) & 64.0 & 0.968 (0.001) & 0.639 (0.002) & 0.884 (0.001) & 0.981 (0.001) \\
          & AbsTopK (half) & 64.0 & 1.000 (0.000) & 0.724 (0.002) & 0.938 (0.001) & 1.000 (0.000) \\
        \midrule
        \multirow{4}{*}{Pythia-1B / \texttt{resid}}
          & SA-GSAE (full) & 65.4 & 0.995 (0.000) & 0.089 (0.002) & 0.162 (0.002) & 0.439 (0.003) \\
          & SA-GSAE (half) & 68.0 & 1.000 (0.000) & 0.099 (0.003) & 0.175 (0.004) & 0.481 (0.005) \\
          & AbsTopK (full) & 64.0 & 1.000 (0.000) & 0.865 (0.003) & 0.981 (0.001) & 1.000 (0.000) \\
          & AbsTopK (half) & 64.0 & 1.000 (0.000) & 0.878 (0.003) & 0.973 (0.001) & 1.000 (0.000) \\
        \midrule
        \multirow{4}{*}{SmolLM3-3B / \texttt{resid}}
          & SA-GSAE (full) & 64.9 & 0.596 (0.007) & 0.007 (0.001) & 0.013 (0.001) & 0.039 (0.001) \\
          & SA-GSAE (half) & 64.1 & 0.845 (0.016) & 0.043 (0.005) & 0.052 (0.004) & 0.082 (0.003) \\
          & AbsTopK (full) & 64.0 & 1.000 (0.000) & 0.859 (0.001) & 0.985 (0.000) & 1.000 (0.000) \\
          & AbsTopK (half) & 64.0 & 1.000 (0.000) & 0.876 (0.001) & 0.980 (0.000) & 1.000 (0.000) \\
        \bottomrule
    \end{tabular}
\end{table}

\FloatBarrier
\subsection{Sweep-based mean-frontier dominance summary (SA-half vs.\ Gated-full)}
\label{app:pareto_sweep_summary}
This subsection replaces the retired matched-$L_0=64$ tables as the reproducibility anchor for the frontier
comparison between half-width SA-GSAE ($H=16{,}384$) and full-width Gated SAE ($2H=32{,}768$). For each of the
six hookpoint $\times$ backbone cells we compute, on a $2000$-point geometric grid in $\log L_0$ over the
overlap of the two variants' swept $L_0$ means, (i) the fraction of the grid on which SA-half's aggregate mean
curve dominates Gated-full's on both metrics (SA-half $R^2 \ge$ Gated-full $R^2$ and SA-half dead-fraction
$\le$ Gated-full dead-fraction; a seed-averaged, not statistically certified, comparison),
(ii) the median and geometric-mean dead-fraction reduction ratio (Gated-full / SA-half; the SA-half denominator is
floored at $0.01$, since unfloored ratios become unstable as it approaches zero -- absolute differences in
\cref{tab:half_width} are primary),
(iii) the peak floored ratio and the $L_0$ at which it occurs, and (iv) the signed extremum of
$\Delta R^2 = R^2_{\text{SA-half}} - R^2_{\text{Gated-full}}$. All statistics are linear-interpolated in
$\log L_0$ between adjacent $\lambda$-sweep aggregate points.

\begin{table}[!h]
    \centering
    \caption{Sweep-based mean-frontier summary. ``Dom.\ frac.''\ is the fraction of the overlap $L_0$ grid on which
    SA-half's mean curve dominates Gated-full's on both metrics (strict $\epsilon=0$). All ratios use a disclosed
    denominator floor of $0.01$ (SA-half dead fractions below $0.01$ are clipped to $0.01$; unfloored ratios reach
    $10^4$ at points where the SA-half dead fraction is ${\sim}10^{-5}$ and are reported nowhere else in the paper).
    Peak ratio $(L_0^{\text{peak}})$ is the
    argmax of the floored ratio on the overlap grid. On attention cells and
    \texttt{resid-mid/SmolLM3-3B} (marked $^\dagger$), SA-half also dominates a larger region under a
    noise-tolerant variant $\epsilon_{R^2} = 10^{-3}$ (at or below the per-sweep-point $R^2$ SE): the tolerant
    dominance regions are $L_0 \in [13.7,\,25.3]$ ($31\%$) on Py/\texttt{attn},
    $L_0 \in [4.8,\,43.3]$ ($69\%$) on SmolLM3/\texttt{attn}, and
    $L_0 \in [36.9,\,111.9]$ ($54\%$) on SmolLM3/\texttt{resid}. Signed $\Delta R^2$ extrema are reported in the
    main-body frontier paragraph.}
    \label{tab:pareto_sweep_summary}
    \scriptsize
    \setlength{\tabcolsep}{4pt}
    \begin{tabular}{llccc}
        \toprule
        Backbone & Hookpoint & Overlap $L_0$ & Dom.\ frac. & Median / Geomean / Peak ratio ($L_0^{\text{peak}}$) \\
        \midrule
        \multirow{3}{*}{Pythia-1B}
        & \texttt{mlp\_out} & $[15.3,\,144.4]$ & $100\%$             & $73\times\,/\,56\times\,/\,76\times$ ($144.4$) \\
        & \texttt{attn}     & $[13.7,\,97.8]$  & $26\%\,^\dagger$    & $1.5\times\,/\,1.5\times\,/\,1.6\times$ ($67.7$) \\
        & \texttt{resid}    & $[17.9,\,129.3]$ & $100\%$             & $54\times\,/\,41\times\,/\,63\times$ ($129.3$) \\
        \midrule
        \multirow{3}{*}{SmolLM3-3B}
        & \texttt{mlp\_out} & $[5.4,\,127.4]$  & $100\%$             & $60\times\,/\,28\times\,/\,84\times$ ($86.1$) \\
        & \texttt{attn}     & $[4.8,\,116.0]$  & $66\%\,^\dagger$    & $2.2\times\,/\,2.1\times\,/\,3.1\times$ ($50.3$) \\
        & \texttt{resid}    & $[21.1,\,111.9]$ & $10\%\,^\dagger$    & $2.0\times\,/\,3.9\times\,/\,71\times$ ($111.9$) \\
        \bottomrule
    \end{tabular}
\end{table}

\FloatBarrier
\subsection{Per-cell $\gamma_\pm$ distribution summary}
\label{app:gamma_symmetry_table}
\begin{table}[!h]
    \centering
    \caption{$\gamma_+$ and $\gamma_-$ distribution summaries at matched $L_0 = 64$ (mean over 3 seeds). Medians agree
    within $\le 0.05$ and $[p_{10}, p_{90}]$ intervals overlap almost exactly on every cell, supporting the
    symmetric-magnitude default $r_i^+ = r_i^-$. AbsTopK's tight spike at $\gamma \approx 1$ is a mechanical consequence
    of its architecture (no dead zone; see App.~\ref{app:theory}), not a calibration-training win.}
    \label{tab:gamma_symmetry}
    \scriptsize
    \setlength{\tabcolsep}{3pt}
    \begin{tabular}{llcccccc}
        \toprule
        & & \multicolumn{3}{c}{$\gamma_+$} & \multicolumn{3}{c}{$\gamma_-$} \\
        \cmidrule(lr){3-5} \cmidrule(lr){6-8}
        Cell & Variant & median & $p_{10}$ & $p_{90}$ & median & $p_{10}$ & $p_{90}$ \\
        \midrule
        Pythia-1B / \texttt{mlp\_out}  & SA-GSAE (full) & 2.42 & 1.52 & 4.68 & 2.41 & 1.52 & 4.67 \\
                                       & SA-GSAE (half) & 2.20 & 1.52 & 3.78 & 2.21 & 1.52 & 3.80 \\
                                       & AbsTopK (full) & 1.00 & 0.97 & 1.02 & 1.00 & 0.97 & 1.02 \\
        SmolLM3-3B / \texttt{mlp\_out} & SA-GSAE (full) & 2.63 & 1.89 & 3.26 & 2.64 & 1.90 & 3.29 \\
                                       & SA-GSAE (half) & 2.38 & 1.68 & 3.10 & 2.38 & 1.69 & 3.13 \\
                                       & AbsTopK (full) & 1.00 & 0.98 & 1.02 & 1.00 & 0.98 & 1.02 \\
        Pythia-1B / \texttt{attn}      & SA-GSAE (full) & 2.53 & 1.26 & 7.63 & 2.52 & 1.27 & 7.66 \\
                                       & SA-GSAE (half) & 2.64 & 1.62 & 8.17 & 2.64 & 1.61 & 7.85 \\
                                       & AbsTopK (full) & 1.00 & 0.98 & 1.02 & 1.00 & 0.98 & 1.02 \\
        SmolLM3-3B / \texttt{attn}     & SA-GSAE (full) & 3.23 & 2.31 & 5.32 & 3.23 & 2.30 & 5.30 \\
                                       & SA-GSAE (half) & 3.21 & 2.28 & 5.15 & 3.19 & 2.28 & 5.14 \\
                                       & AbsTopK (full) & 1.00 & 0.97 & 1.03 & 1.00 & 0.97 & 1.03 \\
        Pythia-1B / \texttt{resid}     & SA-GSAE (full) & 2.05 & 1.53 & 5.34 & 2.05 & 1.53 & 5.36 \\
                                       & SA-GSAE (half) & 1.90 & 1.41 & 4.68 & 1.90 & 1.41 & 4.64 \\
                                       & AbsTopK (full) & 1.00 & 0.96 & 1.03 & 1.00 & 0.96 & 1.03 \\
        SmolLM3-3B / \texttt{resid}    & SA-GSAE (full) & 2.02 & 1.06 & 2.73 & 1.98 & 1.04 & 2.70 \\
                                       & SA-GSAE (half) & 2.22 & 1.65 & 2.87 & 2.20 & 1.64 & 2.86 \\
                                       & AbsTopK (full) & 1.00 & 0.97 & 1.03 & 1.00 & 0.97 & 1.03 \\
        \bottomrule
    \end{tabular}
\end{table}

\FloatBarrier
\subsection{Qualitative latent semantics}
\label{app:qualitative_latents}
\paragraph{Scope caveat.}
The examples in this appendix are illustrative, selected, and not blinded; they should not be read as evidence that
two-sided latents generally encode semantically opposed concepts. A blinded, controlled audit
(\cref{app:rebuttal_studies}) finds judge-nameable semantic opposition to be rare for SA-GSAE and for all tested
baselines at this scale, while sign-conditioned interventions show that the two sides of a latent are nonetheless
causally opposed. We keep the examples below as qualitative illustrations of what individual latents respond to.

\paragraph{Qualitative latent semantics.}
We inspect the top-$64$ latents by absolute activation magnitude from SA-GSAE full-width checkpoints, one per
hookpoint $\times$ backbone (seed $0$; $6$ files in total from the \texttt{examples} stage). Classifying each latent
as \emph{bipolar} when its weaker-sign peak is within a factor of three of its stronger-sign peak,
\emph{positive-only} or \emph{negative-only} when one sign peak is at least ten times the other, and
\emph{sign-dominant} otherwise, we find that bipolar structure concentrates at attention hookpoints:
$12/64$ ($19\%$) of top-$64$ latents on \texttt{attn-mid/Pythia-1B}, $3/64$ ($5\%$) on \texttt{attn-mid/SmolLM3-3B},
and $5/64$ ($8\%$) on \texttt{resid-mid/Pythia-1B}; MLP-output hookpoints and \texttt{resid-mid/SmolLM3-3B} exhibit
$0$--$1$ bipolar latents per file under this criterion and are dominated by sign-monopolar features. This matches
the paper's theoretical picture (sign-aware latents are most useful where bidirectional semantic structure is
actually present) and also sharpens it (bidirectional structure is hookpoint-specific).
Representative latents are collected in \cref{tab:llm_latent_examples}: for each backbone we show one bipolar, one
positive-only and one negative-only latent, covering all three mid-depth hookpoints. The Pythia \texttt{attn}
bipolar latent $22641$ is the paper's motivating case -- its positive side fires inside the ``Parent Revolution''
education-astroturf sentence while its negative side fires on Murdoch-owned-media and product-pitch boilerplate,
carrying anticorrelated real-world evidence along one decoder direction. The SmolLM3 \texttt{mlp\_out} bipolar
latent $26596$ makes the same point at a structural level: its positive side fires on expository sentence endings
while its negative side fires on mid-turn dialogue continuations, showing that sign-sharing also captures contrasts

that are syntactic rather than topical. The human-interpretable descriptions in \cref{tab:llm_latent_examples}
(``education-reform astroturf'', ``Murdoch-owned media'', etc.) were produced by the Amazon Kiro assistant from the
top-$N$ positive and negative activation contexts of each selected latent.

\begin{table*}[t]
    \centering
    \caption{Selected qualitative latents from SA-GSAE full-width checkpoints, seed $0$. For each backbone we show one
    bipolar, one positive-only and one negative-only latent, covering all three mid-depth hookpoints. Each
    positive/negative cell gives a human-interpretable description followed by one representative activation in
    italics, with the activating token rendered in \textbf{bold}; empty cells correspond to the inactive sign of a
    one-sided latent.}
    \label{tab:llm_latent_examples}
    \scriptsize
    \setlength{\tabcolsep}{3pt}
    \begin{tabular}{@{}llcL{0.31\textwidth}L{0.31\textwidth}@{}}
        \toprule
        Model / hookpoint & Category & Latent & Positive & Negative \\
        \midrule
        Pythia-1B / \texttt{attn} & bipolar & 22641
            & Education-reform astroturf sentences (``Parent Revolution'', Excellence-in-Education Foundation, ALEC):
              \newline \emph{released this month. trickster group Parent \textbf{Revolution,} the spark for the current ``W}
            & Murdoch-owned-media and corporate-pitch boilerplate:
              \newline \emph{Wireless Generation is owned by Rupert \textbf{Murdoch}, who also owns Fox News and the Wall Street Journal.} \\
        \midrule
        Pythia-1B / \texttt{mlp\_out} & positive-only & 18628
            & Cross-article boundary: fires immediately after an \texttt{<|endoftext|>} token as a new article begins:
              \newline \emph{reach out to us @Algorithmia.\textbf{<|endoftext|>}As more activists call attention}
            &  \\
        \midrule
        Pythia-1B / \texttt{attn} & negative-only & 14958
            &
            & Progressive / social-justice critique of corporate education reform (MLK citations, anti-privatization discourse):
              \newline \emph{quotes Martin Luther King's \textbf{Letter} from a Birmingham Jail \ldots progressives with a strong belief in social justice} \\
        \midrule
        SmolLM3-3B / \texttt{mlp\_out} & bipolar & 26596
            & End-of-clause punctuation in expository/narrative prose (completed sentence boundaries):
              \newline \emph{Global Energy Mining and Minerals \textbf{Limited,} a Hungarian company, and}
            & Mid-turn continuations in dialogue or interior monologue:
              \newline \emph{but Hugh refuses to reveal \textbf{what} he knows about the situation} \\
        \midrule
        SmolLM3-3B / \texttt{mlp\_out} & positive-only & 20224
            & Transliteration-table entries for proper nouns across Asian languages (Mandarin / Japanese / Korean / Vietnamese):
              \newline \emph{Mandarin [Xi\=u] \textbf{Transliteration} of his Japanese name Russian}
            &  \\
        \midrule
        SmolLM3-3B / \texttt{attn} & negative-only & 19176
            &
            & Video-game dialogue turns (Pok\'emon-franchise NPC lines and interior-monologue asides embedded in web text):
              \newline \emph{``There might be more in there!'' ``Could you be Gym \textbf{Leader} Burgh?'' ``Is that\ldots so?''} \\
        \bottomrule
    \end{tabular}
\end{table*}
\clearpage
\section{Licenses}
\label{app:asset_licenses}

Licenses checked on 21.03.2026.

\begin{table}[H]
    \centering
    \caption{Hosted LLM benchmark assets, licenses, and URLs.}
    \label{tab:asset_licenses}
    \small
    \setlength{\tabcolsep}{4pt}
    \begin{tabular}{L{0.13\linewidth}L{0.50\linewidth}L{0.15\linewidth}}
        \toprule
        Type & Asset & License \\
        \midrule
        Model & \texttt{EleutherAI/pythia-1b} & Apache 2.0 \\
        \multicolumn{3}{L{0.78\linewidth}}{URL: \url{https://huggingface.co/EleutherAI/pythia-1b}} \\
        \midrule
        Model & \texttt{HuggingFaceTB/SmolLM3-3B} & Apache 2.0 \\
        \multicolumn{3}{L{0.78\linewidth}}{URL: \url{https://huggingface.co/HuggingFaceTB/SmolLM3-3B}} \\
        \midrule
        Dataset & \texttt{OpenWebText} & CC0 \\
        \multicolumn{3}{L{0.78\linewidth}}{URL: \url{https://skylion007.github.io/OpenWebTextCorpus}} \\
        \bottomrule
    \end{tabular}
\end{table}

\clearpage

\end{document}